%% file: paper.tex
\documentclass[11pt,a4paper]{article}

\usepackage[utf8]{inputenc}
\usepackage[T1]{fontenc}
\usepackage{lmodern}
\usepackage[margin=2.5cm]{geometry}
\usepackage{microtype}
\DisableLigatures[-]{family=tt*}
\usepackage{booktabs}
\usepackage{array}
\usepackage{enumitem}
\usepackage{xcolor}
\usepackage{amsmath,amssymb}
\usepackage{tikz}
\usetikzlibrary{positioning,arrows.meta,fit,calc,shapes.geometric,shapes.misc,backgrounds,patterns}
\usepackage{pgfplots}
\usepgfplotslibrary{groupplots}
\pgfplotsset{compat=1.17}

\definecolor{okBlue}{HTML}{0072B2}
\definecolor{okOrange}{HTML}{E69F00}
\definecolor{okGreen}{HTML}{009E73}
\definecolor{okVermillion}{HTML}{D55E00}
\definecolor{okSky}{HTML}{56B4E9}
\definecolor{okPurple}{HTML}{CC79A7}
\definecolor{okGrey}{HTML}{6E6E6E}
\tikzset{
  fig/.style={font=\footnotesize, node distance=6mm and 7mm, >={Stealth[length=2.2mm]}},
  proc/.style={draw=black!75, rounded corners=2pt, fill=#1, align=center, inner sep=3pt,
    minimum height=9mm, text width=22mm, line width=0.5pt},
  proc/.default=white,
  data/.style={draw=black!75, fill=white, align=center, inner sep=3pt, minimum height=8mm,
    text width=22mm, line width=0.5pt, path picture={\draw[line width=2pt,black!60]
      (path picture bounding box.north west) -- (path picture bounding box.south west);}},
  verif/.style={draw=black!75, double, double distance=1pt, fill=okGreen!12, align=center,
    inner sep=3pt, minimum height=9mm, text width=22mm},
  decide/.style={diamond, aspect=2, draw=black!75, fill=okOrange!18, align=center, inner sep=1pt,
    font=\footnotesize},
  bound/.style={draw=black!55, dashed, rounded corners=4pt, inner sep=3mm},
  boundlabel/.style={font=\footnotesize\bfseries, text=black!70, anchor=south west, inner sep=1pt, yshift=1.5pt},
  flow/.style={->, line width=0.7pt, black!80},
  ctrl/.style={->, line width=0.7pt, dashed, black!60},
  refuse/.style={->, line width=0.8pt, densely dotted, okVermillion},
  reflabel/.style={font=\small\itshape, text=okVermillion!85!black, align=left},
  lab/.style={font=\footnotesize, fill=white, inner sep=1.2pt, align=center},
  tag/.style={font=\footnotesize\scshape, text=black!65},
}
\pgfplotsset{
  paper/.style={tick label style={font=\footnotesize}, label style={font=\footnotesize},
    title style={font=\footnotesize\bfseries}, legend style={font=\footnotesize, draw=black!40},
    grid=major, grid style={black!12}, axis line style={black!70}},
  lineage r0/.style={okBlue, thick, mark=*, mark size=1.8pt},
  lineage r1/.style={okOrange, thick, dashed, mark=square*, mark size=1.7pt},
  lineage r2/.style={okGreen, thick, dotted, mark=triangle*, mark size=2.3pt},
  lineage r3/.style={okPurple, thick, dash dot, mark=diamond*, mark size=2.3pt},
}
\usepackage[numbers,sort&compress]{natbib}
\usepackage{caption}
\usepackage[colorlinks=true,linkcolor=blue!50!black,citecolor=blue!50!black,urlcolor=blue!50!black]{hyperref}
\usepackage[capitalise,nameinlink]{cleveref}

\setlist{nosep,leftmargin=1.4em}
\input{generated/macros}

\newcommand{\code}[1]{\texttt{\detokenize{#1}}}
\newcommand{\cl}[1]{\textsuperscript{\scriptsize\textcolor{gray}{[#1]}}}

\title{\vspace{-14mm}The Frontier Must Grow:\\[0.2em] \large Self-proposed, interpreter-labelled training roughly doubles a 4B model's accuracy on a fresh exam as its verified frontier grows}
\author{Martin Schröder\\
Swedish Embedded AB, \href{mailto:info@swedishembedded.com}{info@swedishembedded.com}}
\date{November 2026 \textperiodcentered{} \textsc{\mPaperStatus} version}

\begin{document}
\maketitle
\vspace{-6mm}
\input{sections/00-hero}

\input{sections/00-abstract}
\input{sections/01-introduction}
\input{sections/02b-method}
\input{sections/03b-results}
\input{sections/04c-earlier}
\input{sections/05b-elsewhere}
\input{sections/02-related}
\input{sections/10-limitations}
\input{sections/11-reproducibility}

\bibliographystyle{plainnat}
\bibliography{references}

\appendix
\crefalias{section}{appendix}
\crefalias{subsection}{subappendix}
\input{sections/app-series}
\input{sections/03a-loop}
\input{sections/03-architecture}
\input{sections/04-training-release}
\input{sections/05-substrate}
\input{sections/06-requirements}
\input{sections/04b-coding-summary}
\input{sections/07-coding-protocol}
\input{sections/08-pending}
\input{sections/08c-generations}
\input{sections/08b-documentation}
\input{sections/09-replication}
\input{sections/app-related-system}
\input{sections/app-limits}
\input{sections/12-deviations}
\input{sections/13-glossary}

\end{document}

%% file: generated/macros.tex
\newcommand{\mDesignNOneFiveWins}{12}

\newcommand{\mDesignNTwoFivePowerEightZero}{0.89}

\newcommand{\mDesignNTwoFiveWins}{18}

\newcommand{\mEFourCliBfOneSixOnestepHostRssMib}{6970}

\newcommand{\mEFourCliNfFourOnestepHostRssMib}{6924}

\newcommand{\mEFourCliNfFourFourZerostepLossFirst}{2.7806}
\newcommand{\mEFourCliNfFourFourZerostepLossLast}{0.0002}

\newcommand{\mEFourCommit}{258eaba9}

\newcommand{\mEFourProbeBfOneSixOneSixThreeEightFourPeakMib}{77879}

\newcommand{\mEFourProbeBfOneSixOneSixThreeEightFourTokensPerS}{19}

\newcommand{\mEFourProbeBfOneSixEightOneNineTwoPeakMib}{67011}

\newcommand{\mEFourProbeNfFourEightOneNineTwoPeakMib}{35947}

\newcommand{\mEFourProbeNfFourEightOneNineTwoTokensPerS}{20}

\newcommand{\mLedgerChecks}{494}
\newcommand{\mLedgerClaims}{381}
\newcommand{\mLedgerStructure}{240}

\newcommand{\mRsiFourDevRZeroGOneTrainedRecords}{300}

\newcommand{\mRsiFourDevRZeroGTwoTrainedRecords}{900}

\newcommand{\mRsiFourDevRZeroGThreeTrainedRecords}{2100}

\newcommand{\mThrHoursPerMillionTokens}{0.45}
\newcommand{\mThrNfFourEightOneNineTwoPeakMib}{44725}
\newcommand{\mThrNfFourEightOneNineTwoStepS}{13.18}
\newcommand{\mThrNfFourEightOneNineTwoTokensPerS}{622}

\newcommand{\mPaperStatus}{final}

\newcommand{\alphalevel}{0.05}

%% file: sections/00-hero.tex

\begin{center}
\begin{minipage}{\textwidth}
\centering
{\large\bfseries Learning from the interpreter's answers on a self-proposed frontier that grows each
round: Qwen3-4B goes from 8.3\% to 16.7\% on a fresh exam.\par}
\vspace{2mm}
\begin{tikzpicture}[font=\normalsize, >={Stealth[length=2.6mm]}]
\tikzset{hstep/.style={draw=black!75, rounded corners=3pt, fill=#1, align=center, text width=40mm,
    minimum height=11mm, inner sep=2.5pt, line width=0.6pt, font=\small},
  hstep/.default=okSky!25,
  hnum/.style={circle, fill=black!80, text=white, font=\footnotesize\bfseries, inner sep=1.2pt,
    minimum size=4.6mm}}
\node[hstep] (p) at (0,0) {\textbf{Model proposes}\\Lua programs\\from the manual};
\node[hstep=okGreen!18] (l) at (0,-15mm) {\textbf{Interpreter labels}\\each program};
\node[hstep] (f) at (0,-30mm) {\textbf{Keep the frontier}:\\wrong in $\ge$1 of 4 tries};
\node[hstep=okPurple!22] (t) at (0,-45mm) {\textbf{Fresh adapter from}\\\textbf{the base} on the pool\\(300 $\to$ 900 $\to$ 2100)};
\node[hstep=okOrange!25] (e) at (0,-61mm) {\textbf{Separate examiner}\\grades a held-out exam};
\foreach \a/\b in {p/l, l/f, f/t, t/e} \draw[->, line width=0.9pt, black!75] (\a) -- (\b);
\draw[->, line width=0.9pt, dashed, black!55] (t.east) -- ++(5mm,0) |- node[pos=0.25, right, font=\footnotesize, align=left] {next\\round} (p.east);
\foreach \n/\k in {p/1, l/2, f/3, t/4, e/5} \node[hnum, anchor=east] at ([xshift=-1.5mm]\n.west) {\k};
\begin{axis}[paper, at={(60mm,-67mm)}, anchor=south west, width=87mm, height=80mm,
  xmin=-150, xmax=2700, ymin=6, ymax=19, xtick={0,300,900,2100}, xticklabels={0,300,900,2100}, clip=false,
  xlabel={interpreter-labelled training records}, ylabel={fresh exam answered (\% of 2340)},
  tick label style={font=\small}, label style={font=\small}]
\addplot[okBlue, line width=1.6pt, mark=*, mark size=2.4pt] table[col sep=comma,x=records,y=mean] {generated/scaling_s4.csv};
\addplot[okVermillion, line width=1.3pt, densely dashed, mark=square*, mark size=2.2pt] table[col sep=comma,x=records,y=mean] {generated/scaling_m.csv};
\addplot[okBlue, only marks, mark=*, mark size=1.2pt, opacity=0.55, forget plot] table[col sep=comma,x=records,y=l0] {generated/scaling_s4.csv};
\addplot[okBlue, only marks, mark=*, mark size=1.2pt, opacity=0.55, forget plot] table[col sep=comma,x=records,y=l1] {generated/scaling_s4.csv};
\addplot[okBlue, only marks, mark=*, mark size=1.2pt, opacity=0.55, forget plot] table[col sep=comma,x=records,y=l2] {generated/scaling_s4.csv};
\addplot[okBlue, only marks, mark=*, mark size=1.2pt, opacity=0.55, forget plot] table[col sep=comma,x=records,y=l3] {generated/scaling_s4.csv};
\addplot[okVermillion, only marks, mark=square*, mark size=1.2pt, opacity=0.55, forget plot] table[col sep=comma,x=records,y=l0] {generated/scaling_m.csv};
\addplot[okVermillion, only marks, mark=square*, mark size=1.2pt, opacity=0.55, forget plot] table[col sep=comma,x=records,y=l1] {generated/scaling_m.csv};
\addplot[okVermillion, only marks, mark=square*, mark size=1.2pt, opacity=0.55, forget plot] table[col sep=comma,x=records,y=l2] {generated/scaling_m.csv};
\addplot[okVermillion, only marks, mark=square*, mark size=1.2pt, opacity=0.55, forget plot] table[col sep=comma,x=records,y=l3] {generated/scaling_m.csv};
\node[anchor=west, font=\small\bfseries, text=okBlue, align=left] at (axis cs:2150,16.7) {S4\\rounds};
\node[anchor=west, font=\small\bfseries, text=okVermillion, align=left] at (axis cs:2150,14.3) {one fit,\\base\\data};
\node[anchor=north, font=\footnotesize, text=black!70] at (axis cs:300,8.3) {round 1};
\end{axis}
\end{tikzpicture}

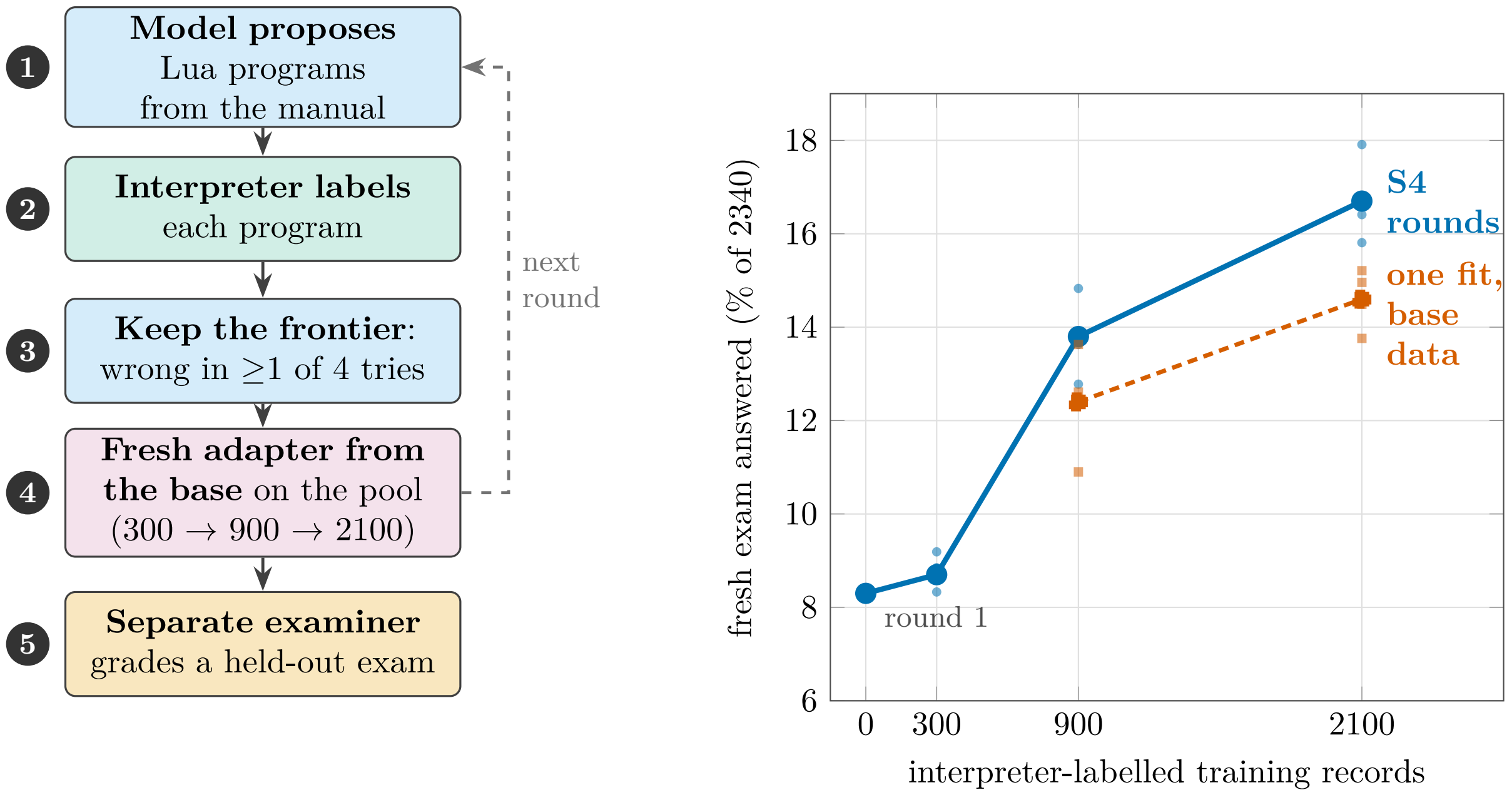
\captionof{figure}{\textbf{S4 at a glance.} Left: one round; rounds share no parameters, only the growing
pool. Right: share of a freshly drawn 2340-item exam answered (large marks: means of four lineages; small
marks: single lineages), against the
interpreter-labelled records trained on: S4's rounds (300, 900, 2100 cumulative records) and one fit from
the base on frontier data the base proposed (900, 2100). The rounds' 2.1-point lead at 2100 records is
inconclusive under the registered analysis (\cref{sec:followups,sec:matched}).}\label{fig:hero}
\end{minipage}
\end{center}
\vspace{1mm}

%% file: sections/00-abstract.tex

\begin{abstract}
A model can learn from problems it poses itself when an executable oracle supplies the answers.
Qwen3-4B proposed Lua~5.1 programs; those it predicted wrong at least once in four samples joined a pool
labelled by the interpreter, each round adding twice as many programs as the last (300, 900, 2100 in
total), and each round trained a fresh adapter from the base on the pool. On the original exam the
procedure met its registered criterion, an item-level test of each adapter against its predecessor. On a
2340-item exam drawn afterwards, accuracy rose from 8.3\% to 16.7\%; that follow-up's registered criteria
were not met because round 1 gained nothing, while rounds 2 and 3 added 5.0 and 2.9 points on average,
both intervals above zero. One fit on the same amount of base-proposed frontier data reached 14.6\%, so
the rounds' 2.1-point advantage is inconclusive. Three earlier
registered series met no per-lineage criterion, and the same loop showed no gain on repository tasks or
for a 27B agent in those configurations.
\end{abstract}

%% file: sections/01-introduction.tex

\section{Introduction}\label{sec:intro}

A model that poses its own problems can learn from them if something outside it supplies the correct
answers. Here that is an executable oracle: Qwen3-4B writes Lua~5.1 programs about the language manual,
and the interpreter says what each prints. This is environment-supervised learning: the model is trained
on the interpreter's targets for problems it gets wrong, including problems it never solves, not
reinforced on its own successes, and no stronger model or human answer takes part. Executable feedback
for self-improvement is not new;~\citep{haluptzok2022}\cl{CT24} our contribution is a registered procedure
that worked, the evidence that it did and what limits it, and the failure analysis of the series that
did not.

The procedure, S4, is pool-growing self-training (\cref{fig:hero}, detailed in \cref{fig:cycle}). In
each round the current champion proposes programs and predicts each four times; those it gets wrong at
least once join a pool labelled by the interpreter, and a fresh adapter is trained from the base on the
whole pool; each round adds twice as many programs as the last (300, 900 and 2100 in total).\cl{RS25,RS26} The rounds share no
parameters; the state that persists is the dataset.

\begin{center}
\fbox{\begin{minipage}{0.94\linewidth}\small
\textbf{Outcome of S4 (four lineages, three rounds; \cref{sec:rsi,sec:followups,sec:matched}).}\\[2pt]
\emph{Fresh exam} (2340 items, 78 templates, drawn after S4 ran, not selected by the base): 8.3\%
$\to$ 8.7\% $\to$ 13.8\% $\to$ 16.7\% answered; rounds 2 and 3 add 5.0 and 2.9 points on average,
both intervals above zero; round 1 adds nothing, so the registered criteria are not met.\cl{FU04,FU12}\\
\emph{Matched data} (one fit on 2100 base-proposed frontier programs): 14.6\%; the rounds' advantage of
2.1 points is inconclusive under the registered analysis.\cl{FU07}\\
\emph{Original exam} (889 items the base missed at selection; adapter level): the registered criterion
is met in 4 of 4 lineages; generation 3 holds in 1 or 2 of 4 under the clustered intervals.\cl{RS29,RS38}
\end{minipage}}
\end{center}

\paragraph{What it shows.} Accuracy grows with the verified frontier: on an exam drawn after the
fact, three rounds roughly doubled the share the base answered, and base-proposed frontier data scaled
the same way from 900 to 2100 records (\cref{fig:scaling}).\cl{FU04,FU07} Whether the rounds help beyond
the data they accumulate is not shown. We separate three senses of recursion: S4 has a recursive
\emph{structure}, since each round's data is proposed and selected by the previous round's model; a
\emph{benefit} of that recursion over matched data is not demonstrated; and an improving \emph{rate} of
improvement is not demonstrated (\cref{sec:matched}). The growing frontier is what the runs that improved
share; a fixed-size but better-selected frontier, more optimisation on fixed
data and other replay schedules were not tested. The first round's gain on the original exam was largely
the exam's selection on the base's failures (\cref{sec:followups}). The earlier series show what S4 needed
(\cref{sec:earlier}); the same loop on repository tasks and on a 27B agent showed no gain, in those
configurations (\cref{sec:elsewhere}).

\paragraph{Contributions.} (i)~A registered multi-round, environment-supervised training procedure on a
self-proposed frontier, examined on a fresh exam and against matched data. (ii)~A learner/examiner
separation specified fully enough to re-implement, with a reference implementation that reproduces the
published decisions. (iii)~A registry of every test of the four series and why most did not meet their
criterion. (iv)~A claims ledger tying every statement to pinned code or a raw record (\cref{sec:repro}).

%% file: sections/02b-method.tex

\section{Method}\label{sec:method}\label{sec:rsi-design}

\paragraph{Terms.} A \emph{series} is a registered experiment of four \emph{lineages}, each a chain
of three \emph{generations} from the base that differs from its siblings only in its seed; the
\emph{champion} is the model a generation's material comes from; a \emph{step} compares a generation with
the one before. The registrations call the series RSI1 to RSI4 (``recursive self-improvement''); we call
them S1 to S4, because what they test is multi-round self-training on self-proposed, oracle-labelled
material; whether an improved model is a better improver is what the matched-data control tested
(\cref{sec:matched}).\cl{RS01}

\paragraph{The learner.} The task family is predicting what a Lua~5.1 program prints. The learner is
the loop of \cref{sec:documentation} reduced to this one kind: the policy, Qwen3-4B with thinking off,
writes programs about sections of the Lua~5.1 manual, the pinned Lua~5.1.5 interpreter runs each
program and its output becomes the reference, and the policy is trained on its own programs with the
interpreter's output as the target. From S2 on every model call and every training run goes through
Brain's API: one \code{brain serve} per lineage serves the proposals, the predictions and the dev and
retention exams, and trains each candidate as a Brain training job.\cl{RS09,RS10} A lineage is a chain
of generations from the base that differs from its siblings only in its seed; each series runs four
lineages of three generations. Inside the loop a candidate is accepted by a dev gate on 400 units
(100 generator programs and 300 programs the base proposed, held out from training) with a retention
set of 300 units; neither holds an exam item.\cl{RS09} The generator programs come from a separate seeded generator of six levels
(a sum, a product, a short string transformation, a loop, a three-item table sort, an arithmetic
expression) that shares no code or templates with the exam's generator, although two of its levels call
functions for which the exam has templates (\code{string.upper}, \code{string.rep}, the length
operator, \code{table.sort}).\cl{RS51} The fresh exam of \cref{sec:followups} comes from the examiner's
own generator, and none of its items matched a program any S4 adapter was trained on or handed over (0
excluded at every step).\cl{FU15}

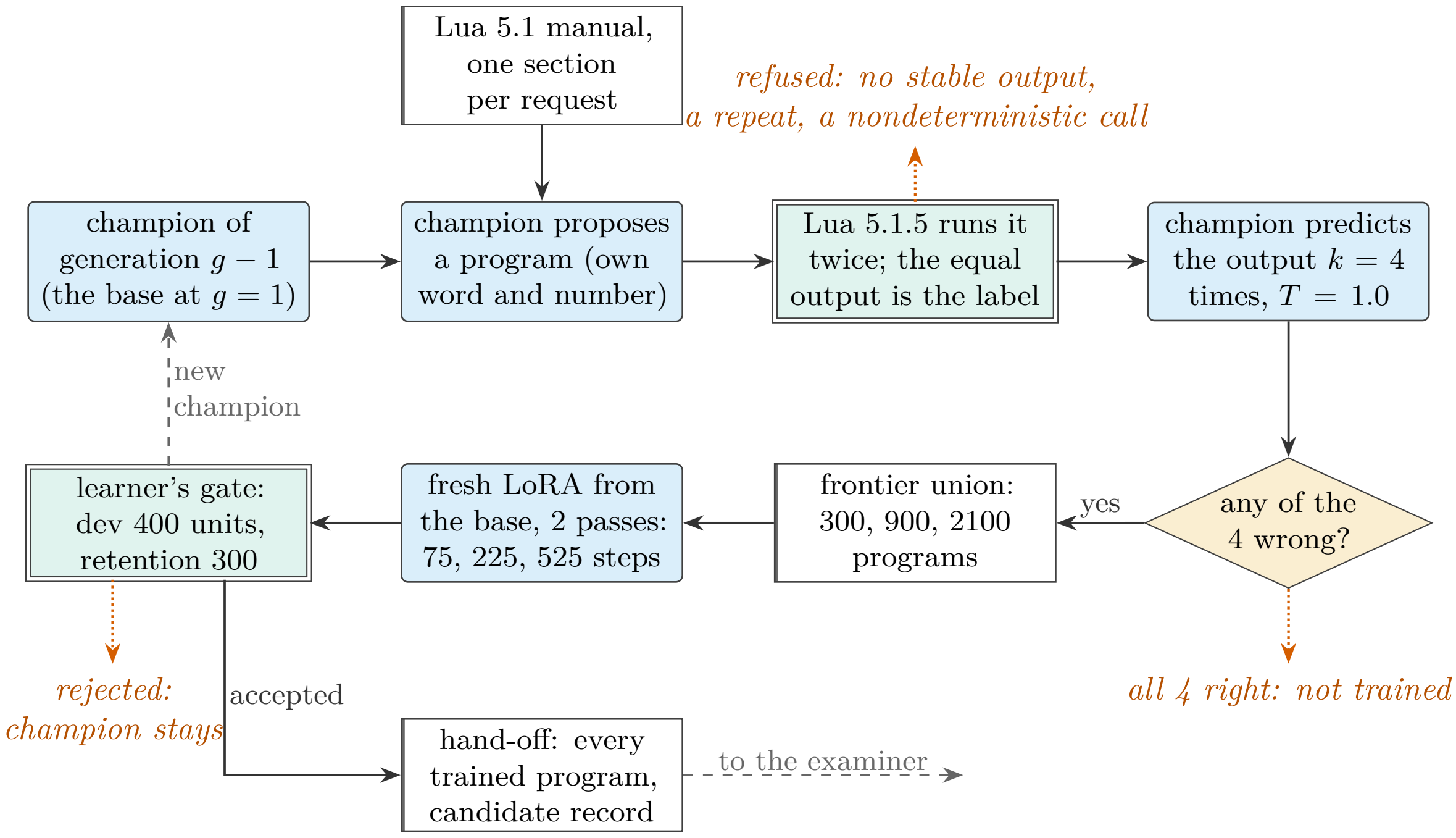
\begin{figure}[tbp]
\centering
\begin{tikzpicture}[fig]
\tikzset{cell/.style={text width=28mm}}
\def\dx{40mm}
\node[data, cell] (manual) at (\dx,0) {Lua 5.1 manual,\\one section per request};
\node[proc=okSky!22, cell] (champ) at (0,-21mm) {champion of\\generation $g-1$\\(the base at $g=1$)};
\node[proc=okSky!22, cell] (prop) at (\dx,-21mm) {champion proposes\\a program (own\\word and number)};
\node[verif, cell] (lua) at (2*\dx,-21mm) {Lua 5.1.5 runs it\\twice; the equal\\output is the label};
\node[proc=okSky!22, cell] (pred) at (3*\dx,-21mm) {champion predicts\\the output $k=4$\\times, $T=1.0$};
\node[decide, aspect=2.2, inner sep=0.5pt, text width=15mm] (front) at (3*\dx,-49mm) {any of the\\4 wrong?};
\node[data, cell] (union) at (2*\dx,-49mm) {frontier union:\\300, 900, 2100\\programs};
\node[proc=okSky!22, cell] (lora) at (\dx,-49mm) {fresh LoRA from\\the base, 2 passes:\\75, 225, 525 steps};
\node[verif, cell] (gate) at (0,-49mm) {learner's gate:\\dev 400 units,\\retention 300};
\node[data, cell] (hand) at (\dx,-76mm) {hand-off: every\\trained program,\\candidate record};
\draw[flow] (champ) -- (prop);
\draw[flow] (manual) -- (prop);
\draw[flow] (prop) -- (lua);
\draw[flow] (lua) -- (pred);
\draw[flow] (pred) -- (front);
\draw[flow] (front) -- node[lab, above] {yes} (union);
\draw[flow] (union) -- (lora);
\draw[flow] (lora) -- (gate);
\draw[flow] ($(gate.south)+(6mm,0)$) |- node[lab, pos=0.3, right] {accepted} (hand.west);
\draw[ctrl] (gate) -- node[lab, right, align=left] {new\\champion} (champ);
\draw[refuse] ($(gate.south)+(-6mm,0)$) -- ++(0,-9mm) node[reflabel, below, align=center] {rejected:\\champion stays};
\draw[refuse] (front.south) -- ++(0,-8mm) node[reflabel, below] {all 4 right: not trained};
\draw[refuse] (lua.north) -- ++(0,6mm) node[reflabel, above, align=center] {refused: no stable output,\\a repeat, a nondeterministic call};
\draw[ctrl] (hand.east) -- node[lab, above] {to the examiner} ++(30mm,0);
\end{tikzpicture}
\caption{One generation of S4, the series that met its criterion, read clockwise from the champion. The champion proposes programs about
the manual and the interpreter labels them; a program joins the frontier when the champion predicts it
wrong in at least one of 4 sampled attempts (300, 600 and 1200 new programs at generations 1 to 3); the
candidate is a fresh LoRA trained from the base on the union of every generation's frontier; the
learner's gate decides whether it becomes the next champion, and the examiner, not the gate, decides
the registered outcome. Dotted arrows are what each stage refuses.}\label{fig:cycle}
\end{figure}

\paragraph{The exam.} The exam is 889 Lua~5.1 programs written by code from 25 standard-library
templates over randomised and edge-case arguments, the items the base never answered in four sampled
attempts under a neutral request at selection (at S4's serving tier it answers 12 of them); the reference of each is what the interpreter prints (twice,
equal).\cl{RS03} Because it was selected as what the base could not do, a count of items right on it
measures progress on that hard subset, not a model's accuracy on Lua programs in general. It was
selected once, in S1, and every series uses the same file (its SHA-256 begins
\code{60ac3ae44651} in every verdict).\cl{RS20,RS31} No model writes or answers an exam item to build
it, and no run of the learner reads it. The reuse has a cost: S1 graded this exam item by item with
its own scripts, and each later design was chosen after the aggregate results of the series before it
on the same exam, so across the series the exam is a holdout reused adaptively, a setting in which a
reused holdout can be overfitted.~\citep{holdout2015}\cl{CT22} Each registration fixed its criterion before
its run, but what the earlier series showed on the reused exam informed each later design, so the
registrations are not independent of it. S4's champions were therefore re-examined on a freshly drawn
exam (\cref{sec:followups}).\cl{FU01}

\begin{figure}[tbp]
\centering
\begin{tikzpicture}[fig]
\tikzset{cell/.style={text width=29mm}}
\node[proc=okSky!22, cell] (gen) at (0,0) {generation $g$:\\propose, label,\\train via Brain};
\node[verif, cell] (lgate) at (0,-22mm) {learner's gate:\\dev 400 units,\\retention 300};
\node[data, cell] (hand) at (37mm,-22mm) {\textbf{hand-off}:\\trained programs,\\then candidate};
\node[data, cell] (exam) at (76mm,0) {\textbf{held-out exam}:\\889 items the base\\missed at selection};
\node[verif, cell] (excl) at (76mm,-22mm) {drop each item\\equal to a listed\\program (also with\\comments removed)};
\node[proc=okOrange!20, cell] (serve) at (76mm,-46mm) {serve all models:\\one pinned build\\and KV tier,\\digests checked};
\node[proc=okOrange!20, cell] (grade) at (113mm,-46mm) {direct completion:\\greedy decoding,\\thinking off,\\last fenced block};
\node[decide, aspect=2, inner sep=0.5pt, text width=14mm] (rule) at (113mm,-22mm) {$W>L$,\\$p<0.05$?};
\node[data, cell] (verd) at (113mm,0) {\textbf{signed verdict}:\\counts, $p$, holds;\\no item, no grade};
\begin{scope}[on background layer]
\node[bound, fit=(gen)(lgate), inner sep=2.5mm] (L) {};
\node[bound, fit=(exam)(serve)(grade)(verd)(rule), inner sep=2.5mm] (E) {};
\end{scope}
\node[boundlabel] at (L.north west) {Learner};
\node[boundlabel] at (E.north west) {Examiner (separate program)};
\draw[flow] (gen) -- (lgate);
\draw[flow] (lgate) -- (hand);
\draw[flow] (hand) -- (excl);
\draw[flow] (hand.south) |- (serve.west);
\draw[flow] (exam) -- (excl);
\draw[flow] (excl) -- (serve);
\draw[flow] (serve) -- (grade);
\draw[flow] (grade) -- node[lab, left, align=right] {wins,\\losses} (rule);
\draw[flow] (rule) -- node[lab, left] {step holds?} (verd);
\draw[refuse] (exam.west) -- node[reflabel, above, align=center] {no path back: no item,\\grade or verdict} (gen.east);
\end{tikzpicture}
\caption{The learner and the examiner. The only channel is one-way: per lineage and
generation the learner writes the list of every program the candidate's chain trained on (with its
held-out dev and retention programs) and then the candidate record (base, adapter and their digests).
The examiner holds the exam, drops items that match a listed program (exactly, or after removing comments and whitespace), serves every model itself
on one pinned build, grades by the direct-completion protocol, applies the registered per-step rule
and signs aggregates only, which are published with the learner's records after the series. The
dotted path does not exist.}\label{fig:examiner}
\end{figure}
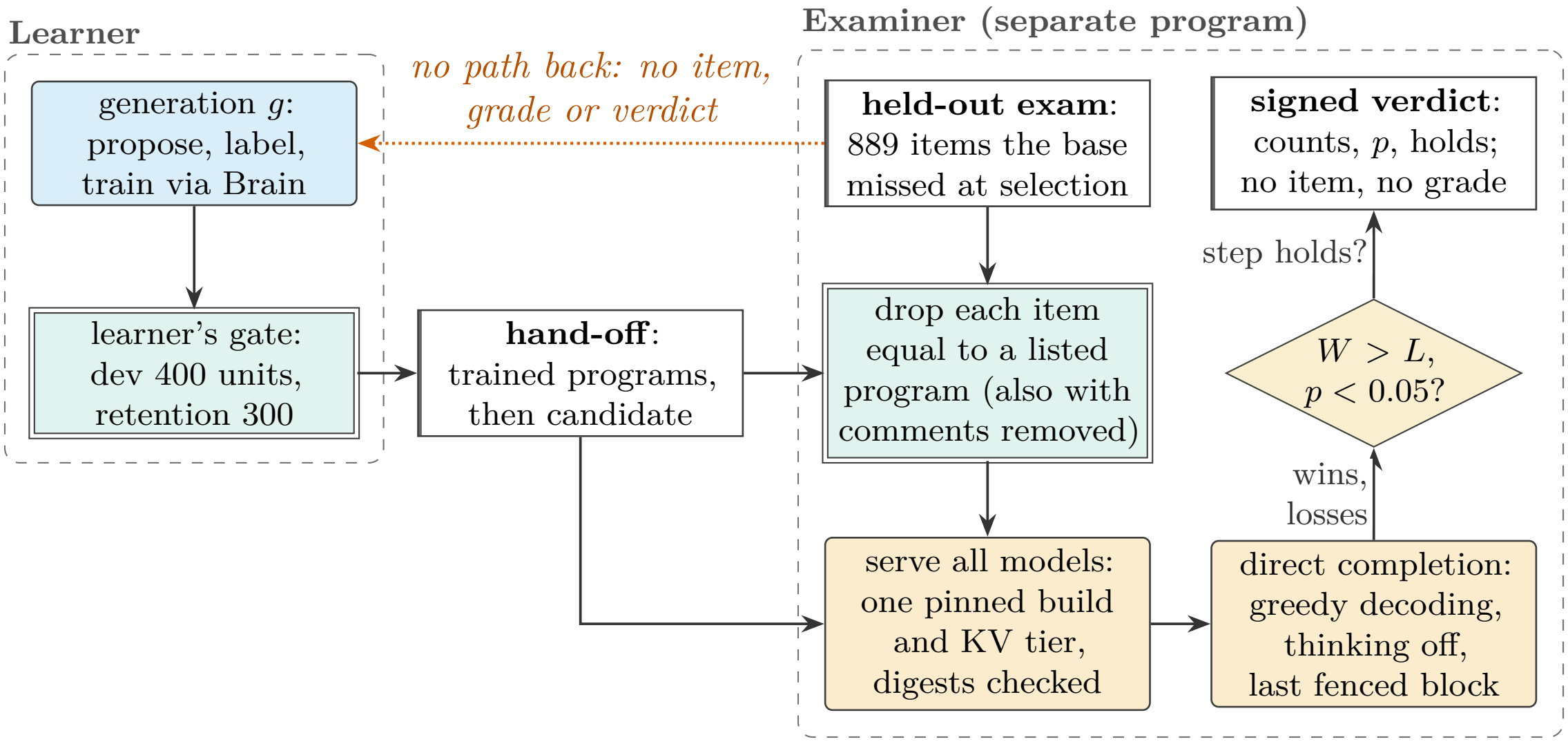

\paragraph{The examiner.} From S2 on the exam is held by an examiner, a separate program operated
beside the learner by the same operators; it is independent of the learner, not of the authors. The
learner's only contact with it is a one-way hand-off (\cref{fig:examiner}).\cl{RS04} For each lineage and generation the learner writes a directory with
the list of every program the candidate's chain trained on and, last, a candidate record (base model,
base digest, adapter path and digest, serving flags, commits). The examiner then
\begin{enumerate}
\item serves the base and every candidate itself, on one pinned Brain serving binary (SHA-256
  \code{d11e9134...}, a different Brain build from the learner's commit \code{796871647}) and, in S4, at one KV-cache tier (f32), and marks a verdict invalid when the
  served base is not the declared checkpoint (identified from the checkpoint's own files) or the
  adapter does not have the declared digest; that build's source is not in Brain's published
  history, so the reference examiner of \cref{sec:repro} serves on the learner's Brain instead;\cl{RS06,RS32}
\item excludes from each step's comparison every exam item that equals, exactly or after removing
  comments and whitespace, any program on either compared model's hand-off list (the programs its
  chain trained on and, beside them, its held-out dev and retention programs), and reports how many
  it excluded; it also reports a near-duplicate count whose similarity measure the verdicts do not
  define, and a program of the same template with other arguments is not excluded;\cl{RS05}
\item asks each item once under the direct-completion protocol: the program and its question, followed
  by the sentence ``Answer with exactly what it prints, in one fenced code block.'' (the request S1
  registered), as the only user message, no system prompt, one greedy completion with thinking off and, in S4, at most
  2048 tokens; the answer is the last fenced block of the reply, or the whole reply if it has none,
  compared with the reference token by token per line with whitespace runs collapsed and blank lines
  ignored; a reply the token limit cut is graded as written;\cl{RS32,RS36}
\item pairs the candidate with its predecessor item by item (a win: the candidate is right and the
  predecessor wrong; a loss: the reverse), computes the exact one-sided sign test over the discordant
  items, and decides the registered rule: a step holds when wins exceed losses at $p<0.05$, a lineage
  succeeds when all three of its steps hold, and the series meets its criterion when at least three of
  the four lineages succeed;\cl{RS02}
\item returns, per step, a verdict with aggregates only (items right, wins, losses, $p$, whether the
  step holds, exclusion and overlap counts, integrity flags) and, over the series, a plan verdict;
  each is signed with the examiner's HMAC-SHA256 key, so the learner, which does not hold the key,
  cannot alter or forge one; the key protects against the learner, not against the authors. Each verdict
  also carries the paired gain in the fraction of items right with a 95\% interval clustered over the
  exam's 25 templates (\cref{sec:rsi-robust}). No item text and no item-level grade leaves the
  examiner.\cl{RS04,RS31,RS38}
\end{enumerate}
The items themselves are not secret: the exam file was released with S1's records, so the
separation rests on the learner's code never opening it and on the exclusion of overlapping programs,
not on secrecy.\cl{RS03}

\paragraph{What is compared.} Every comparison is within a lineage and between consecutive
generations, on the items no compared model trained on, so a model's gain is measured against the
model it was trained from. The registered decision is per lineage. The four lineages share the exam's
items, so a sign test pooled over lineages treats correlated pairs as independent; the pooled figures
are reported as a description, not as a test. The examiner's base-against-base null replicate, run before S4, answered
12 of the 889 items on both sides with 0 discordant items and 0 failed requests: the protocol is
deterministic at the serving build and tier, so a discordant item is a difference between the
models.\cl{RS07}

\paragraph{Three levels of claim.} The item-level sign test supports a claim about two adapters: this
adapter answers more of these items than its predecessor. The template-clustered interval supports a
claim about programs of these templates. Only replication over seeds supports a claim about the
procedure, and with four lineages that is a count, not a test: 4 of 4 has a one-sided sign-test $p$ of
0.0625, and the generation-3 champions answered 219.0 items on average with a standard deviation of 20.4
between seeds.\cl{RS53} Each result below is stated at its level.

%% file: sections/03b-results.tex

\section{Results}\label{sec:rsi}

\subsection{S4: retraining from the base on a frontier that grows each round}\label{sec:rsi4}

S4 changed what S3's records pointed at (\cref{sec:earlier}), and with it the schedule
(\cref{fig:cycle}). \emph{Retraining from the base}: generation $g$'s candidate is a fresh LoRA trained
from the base on the union of every generation's frontier, not a continuation of generation $g-1$'s
adapter; only the data of generation $g$ is new, proposed and predicted by generation $g-1$'s
champion. \emph{Frontier selection}: a program enters when at least one of the champion's 4 sampled
predictions (temperature 1.0) is wrong, so the frontier also holds programs it usually gets right: one
predicted right with probability 0.9 per sample enters with probability 0.34.\cl{RS25,RS35}
\emph{Schedule}: 300, 600 and 1200 new programs per generation, two passes over the union, no separate
replay, so the generations trained on 300, 900 and 2100 records for 75, 225 and 525 steps.\cl{RS25,RS26}
These changes were made together and no ablation separates them. Before any verdict, the examiner's
null replicate (base against base) found no discordant item, the base was served at the configuration
under which it answers 12 exam items (\cref{app:series}), and the examiner's build was changed so that
it identifies the declared base by the checkpoint's content.\cl{RS07,RS27,RS06}

\begin{table}[tbp]
\centering\small
\setlength{\tabcolsep}{3.4pt}
\input{generated/rsi_table}
\caption{S3 and S4 as the examiner decided them: items right of 889 and each step's wins/losses
with the exact one-sided sign test (a step holds at $p<0.05$ with wins above losses; the criterion is
3 of 4 lineages holding at every step). Pooled rows sum the lineage rows (they equal the plan verdict's) and are descriptive, without a test, because the lineages share the exam's items.\cl{RS30} $^\dagger$~r2 is compared on the 888 items left after one overlapping item was excluded; at generation~3 it answered 242 of the 889 before the exclusion. S4 trained on \mRsiFourDevRZeroGOneTrainedRecords{}, \mRsiFourDevRZeroGTwoTrainedRecords{} and \mRsiFourDevRZeroGThreeTrainedRecords{} cumulative records at generations 1 to 3. S3's base was served under a different configuration from S4's (\cref{app:series}), so compare steps, not levels.}\label{tab:rsi}
\end{table}

\paragraph{Outcome (adapter level).} All twelve S4 steps held (\cref{tab:rsi}). From 12 items at the
base the lineages reached 192, 225, 241 and 218 at generation~3; the weakest step, r3's third, won 67
items and lost 36 ($p=0.0015$).\cl{RS28} The plan verdict reads met, 4 of 4 lineages where the rule
needed 3.\cl{RS29} One exam item equalled a program on lineage r2's hand-off list from generation~1, a base
proposal held out in its dev set, not a training program; the examiner excluded it from r2's comparisons and raised its overlap flag on r2's three
verdicts; no other overlap or flag was raised, and no request failed.\cl{RS31,RS40,RS50}

\begin{figure}[tbp]
\centering
\begin{tikzpicture}
\begin{groupplot}[paper, group style={group size=2 by 1, horizontal sep=6mm, y descriptions at=edge left},
  width=0.5\linewidth, height=7.4cm, xmin=-0.06, xmax=0.19, ymin=-14.7, ymax=0.7,
  xtick={-0.05,0,0.05,0.1,0.15}, xticklabel style={/pgf/number format/fixed, /pgf/number format/precision=2},
  ytick={0,-1,-2,-4,-5,-6,-8,-9,-10,-12,-13,-14},
  yticklabels={r0 g1,r0 g2,r0 g3,r1 g1,r1 g2,r1 g3,r2 g1,r2 g2,r2 g3,r3 g1,r3 g2,r3 g3},
  ymajorgrids=false, xlabel={paired gain (share of 889 items)},
  extra x ticks={0}, extra x tick labels={}, extra x tick style={grid style={black!80, thick}},
  error bars/error bar style={line width=0.9pt}, error bars/error mark options={rotate=90, mark size=2.5pt}]
\nextgroupplot[title={S3},
  legend style={at={(1.03,-0.2)}, anchor=north, legend columns=2, /tikz/every even column/.append style={column sep=5mm}}]
\addplot[only marks, mark=*, mark size=2.2pt, okBlue, error bars/.cd, x dir=both, x explicit]
  table[col sep=comma, x=gain, y=y, x error minus=minus, x error plus=plus,
        restrict expr to domain={\thisrow{holds}}{1:1}] {generated/rsi3_forest.csv};
\addplot[only marks, mark=o, mark size=2.4pt, thick, okVermillion, error bars/.cd, x dir=both, x explicit]
  table[col sep=comma, x=gain, y=y, x error minus=minus, x error plus=plus,
        restrict expr to domain={\thisrow{holds}}{0:0}] {generated/rsi3_forest.csv};
\legend{registered sign test holds, registered sign test does not hold}
\nextgroupplot[title={S4}]
\addplot[only marks, mark=*, mark size=2.2pt, okBlue, error bars/.cd, x dir=both, x explicit]
  table[col sep=comma, x=gain, y=y, x error minus=minus, x error plus=plus,
        restrict expr to domain={\thisrow{holds}}{1:1}] {generated/rsi4_forest.csv};
\addplot[only marks, mark=o, mark size=2.4pt, thick, okVermillion, error bars/.cd, x dir=both, x explicit]
  table[col sep=comma, x=gain, y=y, x error minus=minus, x error plus=plus,
        restrict expr to domain={\thisrow{holds}}{0:0}] {generated/rsi4_forest.csv};
\end{groupplot}
\end{tikzpicture}
\caption{Every step's paired gain with the examiner's 95\% interval clustered over the exam's 25
program templates, read from the signed verdicts. Filled points are steps whose registered item-level
sign test holds, open points steps whose test does not. Every first and second S4 step excludes zero;
of the third steps only r2's does, and r0's, r1's and r3's reach below zero, so the third generation's
gain is not robust to the clustering of items by template in three lineages.}\label{fig:rsi-forest}
\end{figure}
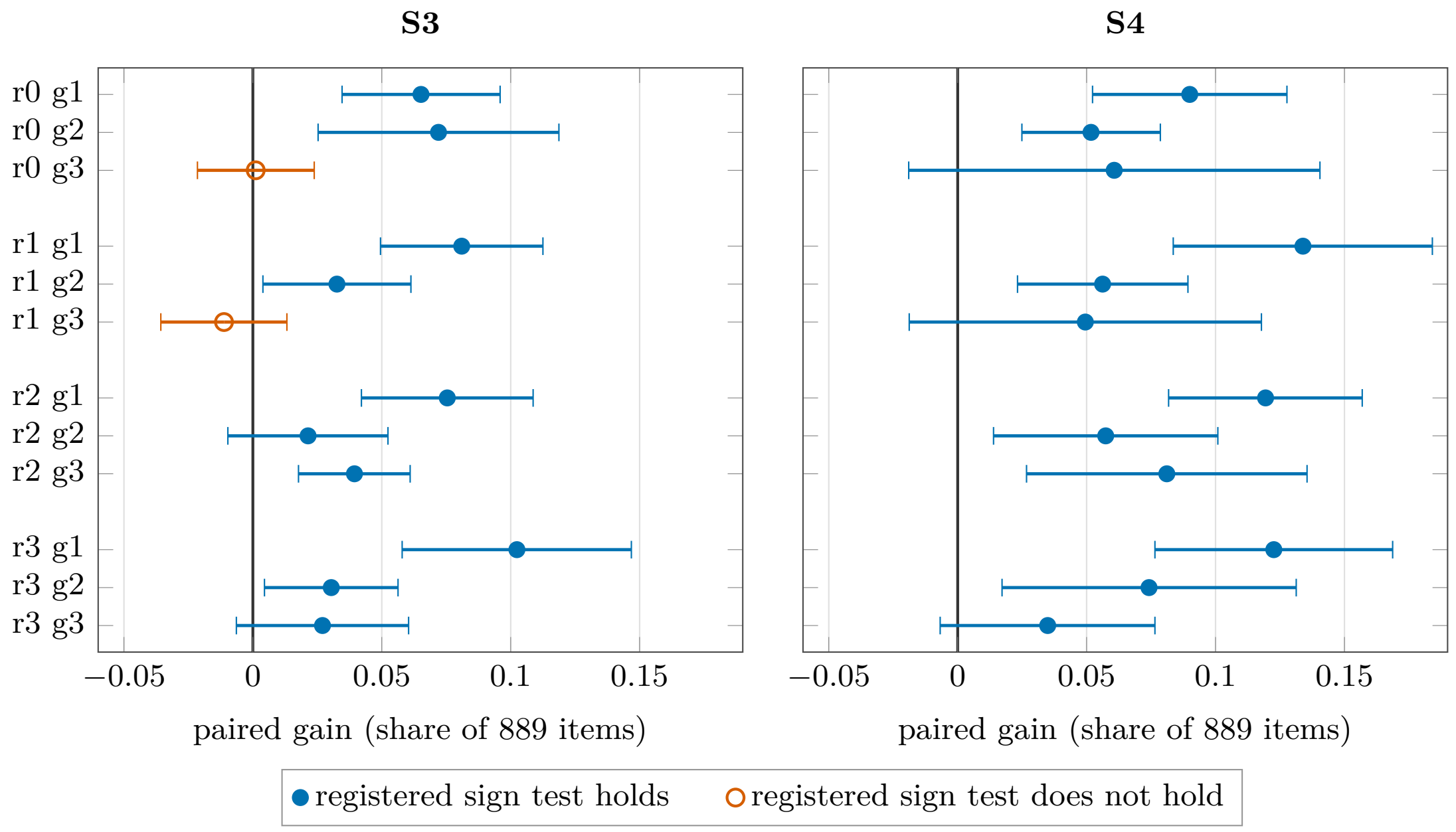

\subsection{Template level: the clustered intervals}\label{sec:rsi-robust}

The sign test counts the 889 items as independent, but they come from 25 templates. The examiner also
reports, for every step, the mean per-item paired difference $d_i\in\{-1,0,1\}$ with a cluster-robust
standard error over templates, $\sqrt{\tfrac{G}{G-1}\sum_{c}\big(\sum_{i\in c}(d_i-\bar d)\big)^2}\,/\,n$
with $G=25$, and a two-sided Student-$t$ 95\% interval with $G-1=24$ degrees of freedom
(\cref{fig:rsi-forest}). Every S4 step at generations 1 and 2 excludes zero. At generation~3 only r2's
interval does (0.0811, from 0.0267 to 0.1355); r0's, r1's and r3's include zero. The matching one-sided
95\% bounds are $-0.0054$, $-0.0071$, 0.0360 and 0.0003 for r0 to r3. At the template level, then, the
third generation improves in one lineage (two-sided) or two (one-sided) of four, and a 3-of-4 rule would
not be met; in S3 no lineage clears all three steps.\cl{RS38} The average generation-3 gain over the
lineages is 5.7 points.\cl{RS53} On the fresh exam, where the examiner computed the lineage-averaged
estimate with an interval over templates and lineages, the third round adds 0.029 (0.011 to 0.050) and
the second 0.050 (0.026 to 0.079).\cl{FU12}

\subsection{What the learner's records show}\label{sec:rsi-process}

\paragraph{Step 1 is inflated by selection.} The exam was selected from items the base got wrong, so
at step 1 the base can lose at most the 12 items it answers, and lost 2, 2, 3 and 1; steps 2 and 3 are
not so constrained.\cl{RS28} On the fresh exam, whose items were not selected by the base's answers,
step 1 holds in no lineage (\cref{sec:followups}).\cl{FU03}

\paragraph{Churn.} Items right are not cumulative mastery. At step 3 the new adapter lost 28\%, 30\%,
14\% and 19\% of the items its predecessor answered (r0 to r3) while winning 12.4\%, 14.0\%, 13.2\% and
9.5\% of those it missed, although it was trained on a superset of its predecessor's data from the same
base.\cl{RS48} Retraining the same data under another training seed shows how much of this is noise:
on the fresh exam such retrains split 87/57, 70/56, 48/56 and 35/55 items against the originals while
their mean accuracy agreed (\cref{sec:matched}).\cl{FU08}

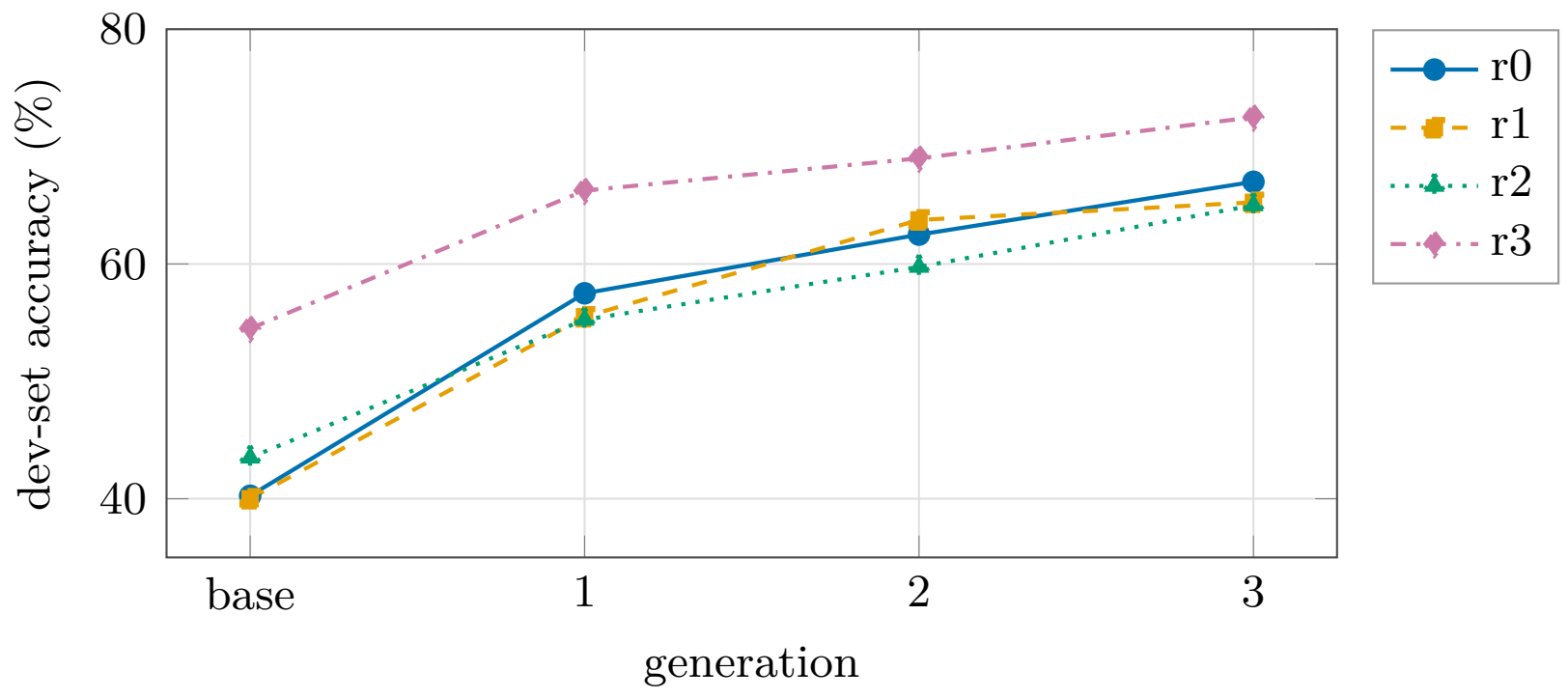
\begin{figure}[tbp]
\centering
\begin{tikzpicture}
\begin{axis}[paper, width=0.6\linewidth, height=5.2cm, xmin=-0.25, xmax=3.25, ymin=35, ymax=80,
  xtick={0,1,2,3}, xticklabels={base,1,2,3}, xlabel={generation}, ylabel={dev-set accuracy (\%)},
  legend pos=outer north east, legend cell align=left]
\addplot[lineage r0] table[col sep=comma,x=gen,y=r0] {generated/rsi4_dev.csv};
\addplot[lineage r1] table[col sep=comma,x=gen,y=r1] {generated/rsi4_dev.csv};
\addplot[lineage r2] table[col sep=comma,x=gen,y=r2] {generated/rsi4_dev.csv};
\addplot[lineage r3] table[col sep=comma,x=gen,y=r3] {generated/rsi4_dev.csv};
\legend{r0,r1,r2,r3}
\end{axis}
\end{tikzpicture}
\caption{S4 accuracy on each lineage's own 400-program dev set (100 generator programs and 300 base
proposals, not selected by failure), the absolute effect size the learner's records give.}\label{fig:dev}
\end{figure}

\paragraph{Dev accuracy and frontier yield.} On each lineage's own 400-program dev set, which was not
selected by failure, accuracy rose from 0.4025, 0.40, 0.435 and 0.545 at the base to 0.67, 0.6525, 0.65
and 0.725 at generation~3 (\cref{fig:dev}).\cl{RS46} The champions kept proposing programs they could
not predict: the share of predicted proposals on the frontier fell from 84--93\% at generation~1 to
61--71\% at generation~2 and 47--54\% at generation~3, so even the third champion missed about half of
its own proposals.\cl{RS47}

\paragraph{The dev gate filtered nothing.} As run, the learner's gate advanced every candidate with a
positive net dev gain; significance at its spent alpha (0.025, 0.0083 and 0.0042 at generations 1 to 3)
was recorded, not required. Three of S4's twelve accepted steps (r1's third, r3's second and third) and
six of S3's twelve were not significant at that alpha, and in S3 two accepted steps did not hold on the
exam. At r1's third step the gate saw 25 dev wins against 19 losses ($p=0.226$) while the exam
showed 99 against 55.\cl{RS49,RS24,RS33} TRUST replaces such a greedy gate with an online multiple-testing procedure that
controls the rate of accepted inferior updates;~\citep{trust2026}\cl{CT15} a gate of that kind exists in
the learner as an option but no series used it.\cl{RS34}

\subsection{The fresh exam}\label{sec:followups}

Both follow-ups were registered after S4's outcome and amended once before any of their verdicts
(\code{prereg/rsi4-followups.md}).\cl{RS41,RS52} \emph{F} re-examined the base and the twelve S4
champions on an exam the examiner drew afresh: 2340 programs from 78 templates, never published and not
selected by the base's answers; the base answers 194 of them (8.3\%), and the examiner's base-against-base
replicate was exact.\cl{FU01} Neither registered criterion is met: step~1 holds in no lineage, by the sign
test or by the one-sided clustered bound; steps 2 and 3 hold in every lineage by the sign test, and by
the clustered bound step~2 in four and step~3 in three.\cl{FU02,FU03} Averaged over the lineages, the
share answered was 8.3\% at the base, 8.7\% after generation~1, 13.8\% after generation~2 and 16.7\%
after generation~3, a gain of 0.084 over the base (95\% interval over templates and lineages 0.046 to
0.124); the first round's gain, 0.004 ($-0.016$ to 0.023), is indistinguishable from
zero.\cl{FU04} The first round's apparent gain on the original exam was therefore largely selection:
on F the items the base answered fell by 0.371 at generation~1 while those it missed rose by 0.038,
the signature of regression to the mean, and by generation~3 the base-wrong items had gained
0.119.\cl{FU05} The similarity audit puts the median exam item at a token-set Jaccard of 0.54 to its
nearest training program (0.67 at the 90th percentile); one item has exactly the token set of a training
program without matching it textually, so it was not excluded.\cl{FU06}

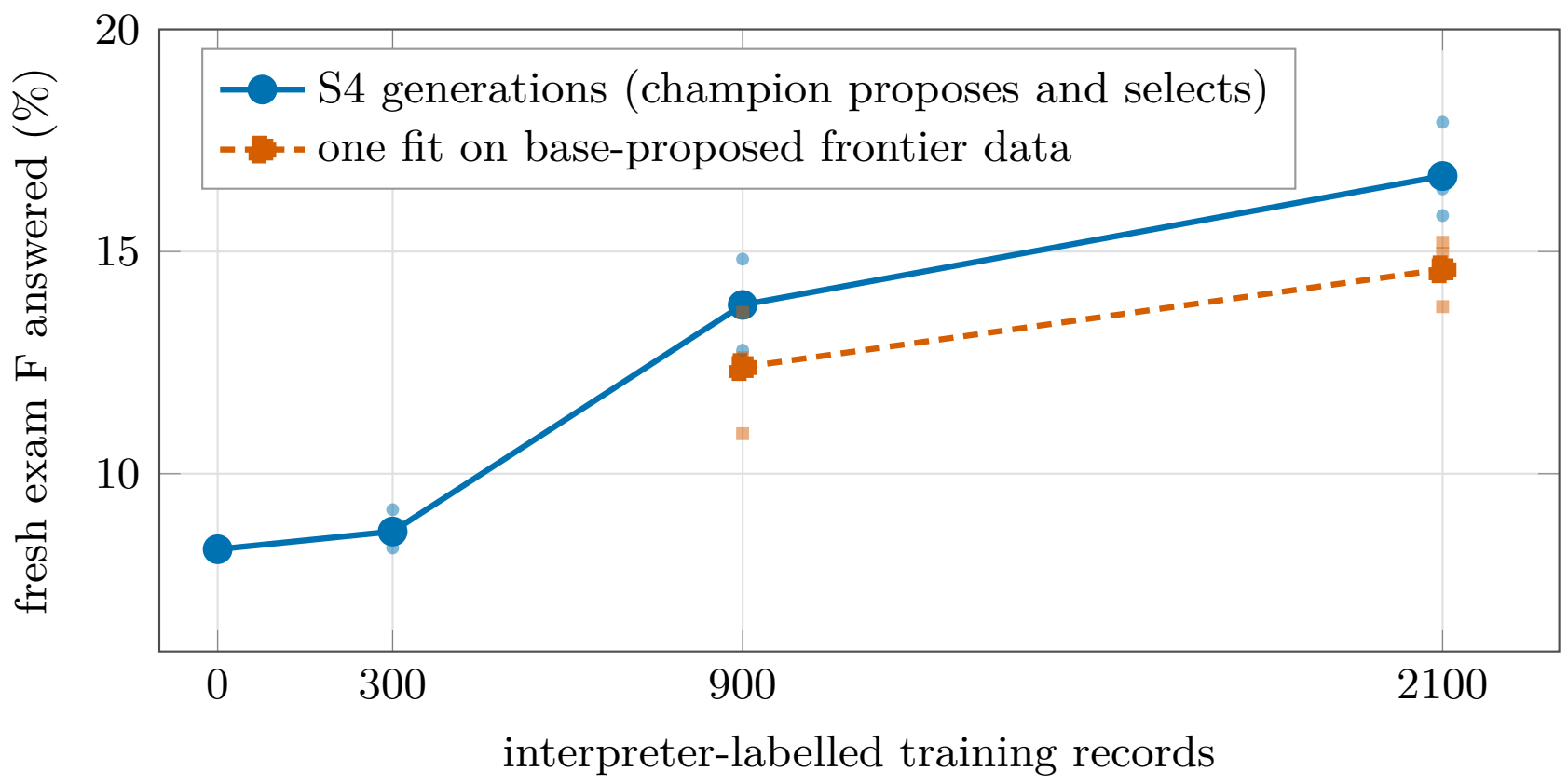
\begin{figure}[tbp]
\centering
\begin{tikzpicture}
\begin{axis}[paper, width=0.72\linewidth, height=6cm, xmin=-100, xmax=2300, ymin=6, ymax=20,
  xtick={0,300,900,2100}, xticklabels={0,300,900,2100}, xlabel={interpreter-labelled training records}, ylabel={fresh exam F answered (\%)},
  legend pos=north west, legend cell align=left]
\addplot[okBlue, very thick, mark=*, mark size=2.4pt] table[col sep=comma,x=records,y=mean] {generated/scaling_s4.csv};
\addplot[okVermillion, very thick, densely dashed, mark=square*, mark size=2.2pt] table[col sep=comma,x=records,y=mean] {generated/scaling_m.csv};
\addplot[okBlue, only marks, mark=*, mark size=1.1pt, opacity=0.5, forget plot] table[col sep=comma,x=records,y=l0] {generated/scaling_s4.csv};
\addplot[okBlue, only marks, mark=*, mark size=1.1pt, opacity=0.5, forget plot] table[col sep=comma,x=records,y=l1] {generated/scaling_s4.csv};
\addplot[okBlue, only marks, mark=*, mark size=1.1pt, opacity=0.5, forget plot] table[col sep=comma,x=records,y=l2] {generated/scaling_s4.csv};
\addplot[okBlue, only marks, mark=*, mark size=1.1pt, opacity=0.5, forget plot] table[col sep=comma,x=records,y=l3] {generated/scaling_s4.csv};
\addplot[okVermillion, only marks, mark=square*, mark size=1.1pt, opacity=0.5, forget plot] table[col sep=comma,x=records,y=l0] {generated/scaling_m.csv};
\addplot[okVermillion, only marks, mark=square*, mark size=1.1pt, opacity=0.5, forget plot] table[col sep=comma,x=records,y=l1] {generated/scaling_m.csv};
\addplot[okVermillion, only marks, mark=square*, mark size=1.1pt, opacity=0.5, forget plot] table[col sep=comma,x=records,y=l2] {generated/scaling_m.csv};
\addplot[okVermillion, only marks, mark=square*, mark size=1.1pt, opacity=0.5, forget plot] table[col sep=comma,x=records,y=l3] {generated/scaling_m.csv};
\legend{S4 generations (champion proposes and selects), one fit on base-proposed frontier data}
\end{axis}
\end{tikzpicture}
\caption{Accuracy on the fresh exam against the amount of interpreter-labelled training data. Blue: S4's
generations, each a fresh adapter from the base on 300, 900 and 2100 cumulative records; red: the
matched-data controls, one fit from the base on 900 or 2100 frontier programs the base proposed. Large
marks are group means, small marks single lineages. Accuracy grows with the verified frontier in both;
the first 300 records gain nothing on F.}\label{fig:scaling}
\end{figure}

\subsection{Matched data: is the improvement recursive?}\label{sec:matched}

\emph{M} trained one adapter from the base, in each of four new lineages, on 2100 frontier programs the
\emph{base} proposed and selected, the same amount of data as generation~3. On F it answered 14.6\%
against generation~3's 16.7\%: a difference of 0.021 whose registered interval, resampling templates and
lineages jointly, runs from $-0.002$ to 0.046, so the registered outcome is \emph{inconclusive}; an
interval over templates alone (0.002 to 0.040), which ignores seed variation, is reported only as a
secondary reading.\cl{FU07} The base-proposed data scale: 900 records gave 12.4\% and 2100 gave 14.6\%
(difference 0.022, interval 0.004 to 0.043), and generation~3 exceeds the 900-record control by 0.043
(\cref{fig:scaling}).\cl{FU07} Retraining each generation-3 training set with another seed gave 16.9\%,
equivalent to generation~3 within the registered margin (difference 0.002).\cl{FU07} That seed-null arm
also decides how to read the registered item-level M1: all four generation-3 champions beat their paired
controls (129/81, 122/94, 119/80 and 136/56), but identical-data retrains already split 87/57, 70/56,
48/56 and 35/55 against the originals, so item-level comparisons of single adapters measure seed noise
as much as treatment and are not evidence here, as the registration's amendment anticipated.\cl{FU08}
One identical-data retrain even passes the item-level step rule against its original (87/57,
$p=0.0077$), so per-lineage item sign tests, including the original exam's 4 of 4, are facts about pairs
of adapters, not evidence that the procedure gains.\cl{FU15}
M is matched in data, not in compute: S4 spent 825 training steps over three rounds against 525 for one
fit, 57\% more before the cost of proposing and predicting.\cl{FU09}

\begin{table}[tbp]
\centering\small
\input{generated/fresh_exam_table}
\caption{Every model examined on the fresh exam F (2340 items, direct-completion protocol unless noted;
S4 and control rows are means of four lineages). The last three rows are the examiner's descriptive
baselines, with no criterion; the thinking-on figure is a lower bound, because 372 replies ran out of
their 8192-token budget while thinking.}\label{tab:fresh}
\end{table}

\paragraph{What the base can do by reasoning.} S4 improves direct prediction without thinking, from 8.3\%
to 16.7\% of F (\cref{tab:fresh}). The same base with thinking on answers at least 47.3\% (1107 items,
with 372 replies cut by the 8192-token budget), Qwen3.8-27B-FP8 answers 39.3\% zero-shot without thinking,
and the base's majority over four samples answers 8.1\%.\cl{FU13} The gain is therefore in direct
prediction, not beyond what the base reaches by reasoning, which is why verified, compact reasoning is
the natural target for the next oracle-supervised step. In the direct protocol, replies cut by the
2048-token limit are rare: on the original exam the reference re-grade counts 21, 15, 1 and 14 of 889 for
the generation-1 adapters and at most 4 later.\cl{FU14}

\paragraph{What this says about recursion.} S4 has a recursive \emph{structure}: each round's training
data is proposed and selected by the model the previous round produced; resetting the weights does not
remove it, because the state that persists is the dataset. A \emph{benefit} of that recursion is not
shown: generation~3 beats data the base proposed and selected by a point estimate of 2.1 points, with a
registered interval that includes zero. An \emph{improving efficiency} of improvement is not shown
either. The comparison tests the whole adaptive acquisition, in which the champion both proposes and
selects; which of the two matters would need a proposer by selector design.

\paragraph{The frontier is a failure filter.} A program enters when at least one of four sampled
predictions is wrong, so a program predicted right with probability 0.99 per sample enters with
probability 3.9\%, one at 0.90 with 34.4\%, one at 0.50 with 93.8\%; four identical wrong answers qualify
too.\cl{FU10} The filter selects difficulty, not learnability, and keeps programs the champion never
solves, which is right here because the interpreter supplies the correct output: S4 is learning from the
environment's targets, not reinforcement of the model's own successes.

%% file: generated/rsi_table.tex
\begin{tabular}{@{}l rrrr rr rr rr l@{}}
\toprule
 & \multicolumn{4}{c}{Items right of 889} & \multicolumn{2}{c}{Gen.~1 vs base} & \multicolumn{2}{c}{Gen.~2 vs gen.~1} & \multicolumn{2}{c}{Gen.~3 vs gen.~2} & \\
\cmidrule(lr){2-5}\cmidrule(lr){6-7}\cmidrule(lr){8-9}\cmidrule(lr){10-11}
Lineage & base & g1 & g2 & g3 & W/L & $p$ & W/L & $p$ & W/L & $p$ & All hold \\
\midrule
\multicolumn{12}{@{}l}{\emph{S3}} \\
r0 & 0 & 58 & 122 & 123 & 58/0 & $3.5{\times}10^{-18}$ & 78/14 & $3.1{\times}10^{-12}$ & 29/28 & 0.5 & no \\
r1 & 0 & 72 & 101 & 91 & 72/0 & $2.1{\times}10^{-22}$ & 41/12 & $4.1{\times}10^{-5}$ & 18/28 & 0.948 & no \\
r2 & 0 & 67 & 86 & 121 & 67/0 & $6.8{\times}10^{-21}$ & 38/19 & 0.00817 & 48/13 & $3.8{\times}10^{-6}$ & yes \\
r3 & 0 & 91 & 118 & 142 & 91/0 & $4{\times}10^{-28}$ & 42/15 & 0.00023 & 47/23 & 0.00279 & yes \\
pooled & \multicolumn{4}{l}{2 of 4 lineages succeed} & 288/0 & -- & 199/60 & -- & 142/92 & -- & not met \\
\addlinespace
\multicolumn{12}{@{}l}{\emph{S4}} \\
r0 & 12 & 92 & 138 & 192 & 82/2 & $1.8{\times}10^{-22}$ & 69/23 & $8.4{\times}10^{-7}$ & 93/39 & $1.5{\times}10^{-6}$ & yes \\
r1 & 12 & 131 & 181 & 225 & 121/2 & $7.2{\times}10^{-34}$ & 79/29 & $8{\times}10^{-7}$ & 99/55 & 0.00024 & yes \\
r2 & 12 & 118$^\dagger$ & 169$^\dagger$ & 241$^\dagger$ & 109/3 & $4.5{\times}10^{-29}$ & 79/28 & $4.2{\times}10^{-7}$ & 95/23 & $6.9{\times}10^{-12}$ & yes \\
r3 & 12 & 121 & 187 & 218 & 110/1 & $4.3{\times}10^{-32}$ & 93/27 & $5.6{\times}10^{-10}$ & 67/36 & 0.00146 & yes \\
pooled & \multicolumn{4}{l}{4 of 4 lineages succeed} & 422/8 & -- & 320/107 & -- & 354/153 & -- & met \\
\bottomrule
\end{tabular}

%% file: generated/fresh_exam_table.tex
\begin{tabular}{@{}llcr@{}}
\toprule
Model & Training records & Thinking & F answered (\%) \\
\midrule
Qwen3-4B base & -- & off & 8.3 \\
S4 generation 1 & 300, own frontier & off & 8.7 \\
S4 generation 2 & 900, own frontier & off & 13.8 \\
S4 generation 3 & 2100, own frontier & off & 16.7 \\
same data, new seed & 2100, own frontier & off & 16.9 \\
M-900 (one fit) & 900, base frontier & off & 12.4 \\
M (one fit) & 2100, base frontier & off & 14.6 \\
base, majority of 4 samples & -- & off & 8.1 \\
base, thinking on (8192 tokens) & -- & on & $\ge$47.3 \\
Qwen3.8-27B-FP8, zero-shot & -- & off & 39.3 \\
\bottomrule
\end{tabular}

%% file: sections/04c-earlier.tex

\section{What did not work before S4}\label{sec:earlier}

\Cref{tab:rsi-all} lists every registered test; \cref{app:series} gives each with its numbers. S1 (one
registration with dated arms, graded by its own scripts before the examiner existed) found that gains on
the loop's own probes were mostly adaptation to the probing protocol, built the exam, and showed large
first steps on it with self-proposed programs labelled by the interpreter; third steps did not hold
within a lineage, and arms that replaced the labels with rewards on the policy's own attempts (self-imitation
of reasoning, policy gradient) did not move the exam.\cl{RS11,RS12,RS15,RS17} S2, on-policy policy
gradient, moved the exam by 2 to 4 items per lineage at generation~1 and no step held.\cl{RS19} S3 trained
a continued adapter on the champion's greedy mispredictions: all four lineages improved at generations 1
and 2, but at generation 3 one held flat and one fell, and over the four lineages the mispredicted share
of proposals fell from 2127 of 2400 to 860 of 1942.\cl{RS22,RS23} S4 therefore retrained from the base
on a growing frontier. Because that design search used one exam, S4 was re-examined on a fresh one
(\cref{sec:followups}).\cl{CT22,FU01}

\begin{table}[tbp]
\centering\small
\setlength{\tabcolsep}{3pt}
\begin{tabular}{@{}>{\raggedright\arraybackslash}p{0.13\linewidth}>{\raggedright\arraybackslash}p{0.33\linewidth}>{\raggedright\arraybackslash}p{0.21\linewidth}>{\raggedright\arraybackslash}p{0.27\linewidth}@{}}
\toprule
Test & What it trained on & Registered criterion & Outcome \\
\midrule
S1 A & own tasks, closed-book loop & gen.~3 beats gen.~2 & not met; gains were protocol adaptation\cl{RS11,RS42} \\
S1 B & A without starving proposals & gen.~2 and 3 each improve & not met\cl{RS42} \\
S1 H & own programs, interpreter labels & pooled steps, no replicate falls & met (pooled)\cl{RS12} \\
S1 HP & H in Python & as H & not met at gen.~3\cl{RS13} \\
S1 HR & HP with replay and pooled retention & as H & met (pooled), each step's gain from one replicate\cl{RS14} \\
S1 R & own verified reasoning & monotone gain on unseen families & not met\cl{RS15} \\
S1 P, P2 & policy gradient, two update scales & as R & not met\cl{RS15} \\
S1 HL & H's recipe, four lineages & 2 of 4 lineages & not evaluated (disk quota)\cl{RS16} \\
S1 HL rerun & as HL & 2 of 4 lineages & not met, 1 of 4\cl{RS17} \\
S2 & on-policy policy gradient & 3 of 4 lineages & not met, 0 of 4\cl{RS19} \\
S3 & own frontier, continued adapter & 3 of 4 lineages & not met, 2 of 4\cl{RS22} \\
S4 & own frontier, retrained from the base & 3 of 4 lineages & met, 4 of 4 (clustered: 1 or 2 of 4)\cl{RS29,RS38} \\
\bottomrule
\end{tabular}
\caption{Every registered test of the four series, in order (P and P2 are two tests). One per-lineage
criterion was met, S4's; the pooled criteria H and HR met cannot tell one improving model from gains
spread over replicates.}\label{tab:rsi-all}
\end{table}

%% file: sections/05b-elsewhere.tex

\section{The same loop elsewhere: no gain shown}\label{sec:elsewhere}

The full loop (\cref{app:loop}) was also tested where it was built for. On repository tasks, an 8B
student trained on a 27B teacher's verified trajectories was not shown better than its base by the
registered confirmatory query: it solved 23.0 of 46 tasks on average over three seeds against 20.0, 18
families won and 12 lost ($p=0.18$); a same-seed contrast showed that supervising every action, rather
than only file edits, does harm (\cref{sec:coding-summary}).\cl{QUAL1} Two promoted generations of a
27B model trained on its own agent experience did not beat their base on one measurement
(\cref{sec:generations}).\cl{G2703,G2704} An exploratory, unregistered run in which a 4B model learned
from a raw manual gained on its own probes, with no detected effect of which model wrote its tasks
(\cref{sec:documentation}).\cl{SL23} These negative results hold for those configurations: small
corpora, records truncated to the training block, harness failures, and an exam the 27B model already
solved nine times in ten; they do not show that supervised training memorises and reinforcement learning
generalises, or the reverse. The supervision finding, that training on every agent action does worse than
training on the decisive edits, is secondary but held at the registered seeds.

%% file: sections/02-related.tex

\section{Related work}\label{sec:related}

\emph{Self-training on self-proposed problems.} The closest prior design is Haluptzok et al.: a model
synthesises programming problems with solutions, an interpreter filters them for correctness, and the
model is fine-tuned on the result.~\citep{haluptzok2022}\cl{CT24} STaR and ReST-EM also restart from the
base each iteration on accumulated, filtered self-generated data, but they filter the model's
\emph{outputs} against known answers;~\citep{star2022,restem2023}\cl{CT02,CT16} here the model generates
the \emph{inputs} and the oracle supplies the outputs. Self-Instruct generates inputs too, but labels them
with the same model;~\citep{selfinstruct2022}\cl{CT01} ReST iterates grow and improve steps on filtered
samples.~\citep{rest2023} Noisy Student iterates generations on a growing pool, with a teacher's
labels.~\citep{noisystudent2019}\cl{CT26} Expert iteration amplifies the policy by search and imitates
the result;~\citep{exit2017}\cl{CT19} an interpreter is ground truth, not an amplified policy, so only the
generate--filter--retrain skeleton is shared. In self-play designs a proposer is trained to set
learnable tasks, as in asymmetric self-play and in Absolute Zero, where a code executor validates the
proposed tasks;~\citep{asymselfplay2017,absolutezero2025}\cl{CT28,CT18} our proposer is not trained to
propose, only filtered. Other self-improvement work changes a model's own context or scaffold rather than
its training data: STOP improves the improver scaffold itself,~\citep{stop2023}\cl{CT25} Learning to
Self-Evolve trains a model to improve its context at test time,~\citep{lse2026}\cl{CT31} and hyperagents
let a meta agent modify the agent;~\citep{hyperagents2026}\cl{CT32} SEAL has a model write its own
fine-tuning data and directives.~\citep{seal2025}\cl{CT33} Our matched-data control does not show that the
recursion itself helps here. On reinforcement learning with verifiable rewards, Yue et al.\ find gains at
small $k$ bounded by the base model at large $k$, and new patterns from distillation;~\citep{rlvrbound2025}\cl{CT34}
our own RL arms are too narrow to add to that question.

\emph{Selection and curricula.} Selecting what to label is active learning, here by query synthesis with
the model's own disagreement as the acquisition function; ordering by difficulty is curriculum
learning;~\citep{settles2012,curriculum2009}\cl{CT29} learned curricula over problem categories are a
recent variant.~\citep{sec2025}\cl{CT23}

\emph{Predicting execution.} Predicting what a program prints is CRUXEval's output-prediction
task;~\citep{cruxeval2024}\cl{CT17} models have been trained to predict execution since Learning to
Execute, and more recently with execution traces.~\citep{learningtoexecute2014,codeexecutor2023,semcoder2024}\cl{CT27}
CodeRL trains a generator with a critic of functional correctness from unit tests;~\citep{coderl2022}\cl{CT20}
training repeatedly on a model's own generated content can make the tails of the original distribution
disappear,~\citep{collapse2023}\cl{CT21} which our series counter only by labelling every program with the
interpreter.

\emph{Evaluation.} A greedy gate that accepts any higher validation score lets inferior updates through
across rounds; TRUST controls their rate by online multiple testing.~\citep{trust2026}\cl{CT15} Reusing a
holdout adaptively can overfit it,~\citep{holdout2015}\cl{CT22} and fresh test sets test
replication;~\citep{recht2019} evaluations need clustered standard errors and seed-level
variance.~\citep{errorbars2024,benchvariance2021,precipice2021}\cl{CT30} Related work for the system
components is in \cref{app:related-system}.

%% file: sections/10-limitations.tex

\section{Limitations}\label{sec:limitations}

\paragraph{What the result is about.} One task family (predicting what Lua~5.1 programs print), one base
model, three rounds of LoRA training on interpreter labels, and no tools and no reasoning: thinking is
off everywhere and the training records are answers only. S4 improves direct prediction without
thinking; nothing is claimed about other tasks, models, longer series or reasoning, where the base with
thinking on already answers at least 47.3\% of the fresh exam against S4's 16.7\%.\cl{FU13}

\paragraph{What the statistics support.} Four lineages are encouraging, not a reliability estimate.
``Every generation improves'' is the registrations' strict criterion, not the measure of useful learning;
the learning curves of \cref{fig:scaling} are. On the fresh exam the first round gains nothing and the
registered 3-of-4 criteria are not met; the third round holds in three lineages by the clustered bound.
The advantage of the rounds over matched base-proposed data is inconclusive, and M is matched in data,
not in compute.\cl{FU03,FU07,FU09} The exam that guided the design was reused; the fresh exam was drawn
afterwards, but its templates include those of the old exam.\cl{CT22} The overlap exclusion catches
exact and normalised matches only, and the similarity audit shows the fresh exam lies close to the
training programs (median token-set Jaccard 0.54).\cl{FU06} The examiner is a separate program run by the
same operators, and its symmetric signature protects the verdicts against the learner, not against the
authors. The learner's dev gate advanced on any net gain and filtered nothing.\cl{RS49}

\paragraph{Not continual learning.} Each round retrains from the base, a consolidation baseline rather
than a solution for an unbounded stream: an adapter loses 14 to 30\% of the items its predecessor
answered, and a retention test of what each round acquired is not implemented. The seed-null retrains
show that this churn is mostly training noise at the level of items while lineage means are
stable.\cl{RS48,FU07,FU08}

\paragraph{The rest of the paper.} The limits of the system specification and of the coding, 27B and
documentation experiments are in \cref{app:limits}.

\paragraph{Open directions.} The questions the evidence leaves are richer oracle supervision (execution
traces, compact verified reasoning), a proposer-by-selector design to separate who proposes from who
selects, compute-matched baselines, a learnability rather than a failure curriculum, and a retention
matrix over acquired capabilities.

%% file: sections/11-reproducibility.tex

\section{Reproducibility}\label{sec:repro}

\paragraph{What is needed.} Brain, Sven and Splinter, cloned from their remotes, and the evidence
dataset \code{swedishembedded/the-frontier-must-grow-evidence}. The dataset holds only what this
paper cites: the 889-item exam, the examiner's signed step and plan verdicts and the rule it applied,
the learner's round records, every candidate record with its hand-off list of programs, the twelve
S4 champion adapters, S1's analyses and per-model grades, and the records of the documentation runs
and the 27B generations, one archive per series. \code{reproduce/dataset.json} pins the dataset's
revision and the SHA-256 of every file and archive member, and \code{data/measurements.json} names
each cited record by its path in the unpacked dataset and its digest. Every commit named below is
contained in a branch of its repository's remote; \code{make verify} fails on a pin that is not.

\paragraph{Three tiers.} The targets are run in the paper's directory
(\code{make -C} with the paper's directory); downloads, builds and outputs land
under \code{target/papers}, and \code{EVIDENCE\_DIR}, \code{REPRO}, \code{BRAIN\_SRC}, \code{SVEN\_SRC},
\code{DEVICE} and \code{LOCK} move them, name the clones and the card (\code{make reproduce} lists
the tiers).

\begin{center}\small
\begin{tabular}{@{}>{\raggedright\arraybackslash}p{0.21\linewidth}>{\raggedright\arraybackslash}p{0.47\linewidth}>{\raggedright\arraybackslash}p{0.26\linewidth}@{}}
\toprule
Target & What it reproduces & Hardware and time \\
\midrule
\code{reproduce/verify} & Downloads the dataset at its pinned revision (or uses a local copy), checks
every file, archive member and adapter digest, unpacks it and runs \code{make verify} on it: the
structure checks at the pinned commits, and every measured value re-extracted from its record and
compared with \code{data/measurements.json}. & CPU; a 1.6\,GB download; under a minute on our host. \\
\code{reproduce/examine} & Builds Brain at \code{796871647}, fetches Qwen/Qwen3-4B at its pinned
revision (every file checked), serves the base and each lineage's three adapters with the paper's
flags (\code{--kv-cache f32}, \code{--max-batch 64}, adapters by digest, the base identified by the
checkpoint's content digest), re-grades all 889 items with the reference examiner and writes each
step beside the published verdict. & One GPU (the paper's runs used a GH200); time not yet measured. \\
\code{reproduce/rerun} & Builds the S4 lineage driver at Splinter \code{86f13672a} against Sven
\code{fc99a2f1} and Brain \code{796871647}, fetches the Lua~5.1 manual (and cuts the three sections
the proposer reads, checked against their digests), builds Lua~5.1.5 from its official tarball
(SHA-256 pinned), runs the four S4 lineages with the launch record's settings and seeds, then grades
the new champions as \code{reproduce/examine} does. & One GH200; about 12\,h, since S4's lineages took
2.4 to 3.4\,h each on a card of their own.\cl{RS26} \\
\bottomrule
\end{tabular}
\end{center}

\paragraph{The reference examiner.} \code{reproduce/examiner.py} implements the examination protocol
of \cref{sec:rsi-design} from that specification: the neutral request (the item's instruction and the
sentence ``Answer with exactly what it prints, in one fenced code block.''), no system prompt,
thinking off, one greedy completion of at most 2048 tokens, the last complete fenced block or else the
whole reply as the answer, a reply cut at the limit graded as written, the comparison token by token
per line with whitespace runs collapsed and blank lines ignored, the exclusion of every item equal to
a program on either compared model's hand-off list (exactly, or once comments and whitespace are
removed), the exact one-sided sign test, the paired gain with its template-clustered Student-$t$
interval ($G-1$ degrees of freedom) and the registered per-lineage plan rule. It checks the served
base against the declared base digest and each adapter file against its declared digest, and
serves one lineage at a time, because adapters of different lineages carry the same card ids. Its
unit tests run in \code{make verify}. It serves on the learner's Brain \code{796871647}: the build the
published verdicts were served on (binary SHA-256 \code{d11e9134...}) was made from a Brain source
that is not in Brain's published history. Run on that build, the reference examiner agrees with the published verdicts on
all 12 S4 step decisions, with identical items right, losses and exclusions at every step; the only
differences are the base answering 13 items instead of 12 on lineage r1's server, and r1's first step
counting 120 wins instead of 121.\cl{FU11}

\paragraph{What a reader cannot reproduce.} (1) The fresh exam F of the registered follow-ups and its
generator are not released; only its signed aggregate verdicts are published.
(2) The published verdicts were produced by a different implementation of the examiner than the
reference one, on a different Brain build, so \code{reproduce/examine} re-measures the result rather
than replaying the published grades; a difference within serving nondeterminism is expected, and
the comparison reports it. (3) The verdicts are signed with a symmetric HMAC key, so a reader cannot
verify a signature; the signature guards against the learner, not against the authors.
(4) \code{reproduce/rerun} is a statistical replication, not a bit-identical one: proposals and
frontier predictions are sampled at temperature 1.0 and GPU kernels are not deterministic across
hosts, so the reproduced lineages are new draws of the same procedure. (5) The raw logs of the
training-substrate qualification and throughput probes, a trajectory probe, and the coding-task pilots
and probes were not kept; \code{data/measurements.json} records their values, digests and commands, marked
\code{"kept": false}, and \code{make verify} reports them as not kept.

\paragraph{Verification.} \code{make verify} first runs the specifications of the pin rule, of the
record parsers and of the reference examiner, then re-runs every structure check against the commits
in \code{data/commits.json}, then \code{scripts/build\_data.py --check}, which resolves each measured
and computed claim to its value, requires the claim text to contain the value, requires every ledger
row to be cited in the text and every citation to be in the ledger, and checks that the pooled steps
computed from the examiner's lineage verdicts equal its plan verdicts; with \code{EVIDENCE=DIR} it
also re-extracts every value from the records in \code{DIR}. The ledger holds \mLedgerClaims{}
claims, \mLedgerStructure{} of them structure claims backed by \mLedgerChecks{} checks. The sibling
repositories are given by \code{--sven} and \code{--brain}. \code{make check} includes this
verification as \code{check/papers}. The bibliography entries were checked against the archive's
abstract pages on 2026-11-15.

\paragraph{Draft and final.} The paper's status is recorded in \code{status.txt}; when it reads
final, \code{build\_data.py --check} also fails while any registered-but-unmeasured or planned claim
remains, so a final version holds only structure facts, measurements, computations from measurements
and citations. This version is final: the registered follow-ups have reported, and their verdicts are in
the evidence dataset (\code{records/followups.tar.xz}).\cl{FU02,FU07}

\paragraph{Registrations.} All registrations are in the paper's \code{prereg/} directory: the
coding-task protocol in \code{coding-tasks.md}, the acquisition experiments in \code{acquisition.md},
S1 in \code{recursive-improvement.md}, and S2 to S4 each in its own file (\code{rsi4.md} for
S4) with a launch record (\code{rsi4-launch.md}) that fixes builds, binary digests, settings and
seeds before the first generation and carries the series' dated deviations; the outcome of each series
is appended to its registration.

\paragraph{Build.} From the repository root the paper is built by
\begin{center}\code{make papers/splinter-brain-sven-flywheel/pdf}\end{center}
into \code{target/papers}; every table, plot and number of the series S1--S4 and of the coding-task pilots
is generated from \code{data/} into the build directory.
The same build writes \code{target/papers/splinter-brain-sven-flywheel-arxiv.tar.gz}, the source archive
for arXiv: the main file, the sections, the generated tables, macros and plot data at the paths the
source names, the bibliography and its \code{.bbl}, with comments removed. The build compiles the
archive in an empty directory with \code{pdflatex} and \code{bibtex} and with \code{latexmk}, and
fails unless both give the main PDF's page count with no error or undefined reference; the archive is
uploaded as it is.

%% file: sections/app-series.tex

\section{The earlier series in detail}\label{app:series}

\subsection{S1: self-taught generations, judged first pooled and then per lineage}\label{sec:rsi1}

S1 is one registration with dated arms, each added before it ran; its exams were selected and
graded by the arm's own registered analysis script, without a separate examiner.\cl{RS01} Arms A and B
ran the closed-book loop of \cref{sec:documentation} for three generations and were judged on the
loop's own probes; re-graded under a neutral request, most of the gain they measured was adaptation
to the probing protocol and did not transfer.\cl{RS11} Arm H therefore built the exam described
above and measured every generation on it greedily under the neutral request. Its three replicates,
trained only on tasks their own models wrote, answered 38, 31 and 34 of the 889 items at
generation~1, 60, 63 and 49 at generation~2 and 67, 83 and 59 at generation~3 (base 0); pooled over the
replicates the steps were 103/0, 95/26 and 103/66 wins/losses ($p=0.00273$ at the third) with no
replicate falling, which met the registered pooled criterion.\cl{RS12} The same design in Python
(arm HP, 820 items, base 2) met its first two steps (49/7 and 83/8) and not the third (59/50,
$p=0.222$), where one replicate fell from 69 to 60.\cl{RS13} Arm HR added replay across generations
and a pooled retention bound and met the pooled criterion (66/4, 16/1, 25/3), but each step's gain came
from a single replicate, so it is not evidence that one lineage improved twice.\cl{RS14} Three arms
changed the update instead: self-imitation of the model's own verified reasoning (R) and policy
gradient with negative rewards at two update scales (P, P2). After three iterations they answered 21,
4 and 12 of the 889 items, and on six families none of them trained on their pass@1 went from 0.41
to 0.4479, 0.4408 to 0.4242 and 0.4419 to 0.4408: none met its criterion.\cl{RS15}

Because a pooled test cannot tell staggered single gains from one model improving repeatedly, arm HL
repeated arm H's recipe in four fresh lineages under a per-lineage criterion: a lineage succeeds when
all three of its steps hold, and the arm when two of four do. Its first run ended on the project disk's
inode quota before any lineage finished and was not evaluated.\cl{RS16} The rerun met the criterion in
1 of 4 lineages: r3 rose at every step (16/0; 14/5, $p=0.0318$; 33/12, $p=0.00123$), r2 rose twice and
fell at generation~3 (27/36), r0's later steps were not significant (35/25, $p=0.123$; 37/28,
$p=0.161$), and r1 accepted no round after generation~1. Pooled, the third step was 97/76
($p=0.0641$).\cl{RS17} S1 thus showed large first steps on items the base never answered and
third steps that did not hold within a lineage.

\subsection{S2: on-policy policy gradient}\label{sec:rsi2}

S2 asked whether replacing the update with on-policy policy gradient under powered gates makes the
later steps hold. Each generation trained on the current policy's own graded attempts at programs it
solves sometimes ($k=8$ attempts, group-standardised advantages, clip edges 0.2 and 0.28, learning
rate $6\times10^{-6}$ for 24 steps, LoRA rank 16), and the exam moved to the independent
examiner.\cl{RS10,RS18} At generation~1 the four lineages won 4, 2, 3 and 3 items against the base and
lost none ($p$ from 0.0625 to 0.25): no step held, so no lineage could succeed, and the examiner's
plan verdict reads not met.\cl{RS19} The examiner's overlap audit found that the policy's own
proposals had reproduced 1, 1 and 2 exam programs exactly in three lineages; S2 had no exclusion
and those items count as scored.\cl{RS19} Generation~2 could not be served, because every trained
candidate carried the same adapter card id; the criterion was already decided, so it was not rerun
after the fix.\cl{RS08} The recipe raised the learner's own dev set in its pilot but moved the exam by
two to four items.

\subsection{S3: training on the interpreter's answers to the policy's own programs}\label{sec:rsi3}

S3 returned to arm H's source of information, training on the interpreter's output for programs the
policy itself proposed, and replaced what had failed in S1: acceptance on a dev set powered for an
effect of 0.08, replay of earlier generations, proposals diversified per manual section, and the whole
learner behind Brain's API. A generation proposes up to 600 programs, predicts each greedily and
trains a LoRA continued from the champion (rank 16, learning rate $3\times10^{-5}$, 40 steps) on the
mispredicted programs, with the correctly predicted ones and earlier generations' records each drawn
at a share of 0.25; the examiner now excludes overlapping items.\cl{RS09,RS05}

Every lineage improved at generations~1 and~2; at generation~3 r2 and r3 improved again, r0 held
flat (29/28, $p=0.5$) and r1 fell (18/28) (\cref{tab:rsi}). Two of four lineages succeeded and the
plan verdict reads not met; pooled, the steps were 288/0, 199/60 and 142/92, and no exam item matched a
training program.\cl{RS21,RS22} The learner's records show the signal running out: over the four
lineages the champions mispredicted 2127 of 2400 proposals at generation~1, 1125 of 2228 at
generation~2 and 860 of 1942 at generation~3, so later generations trained mostly on what their
predecessor already knew, while each also continued the previous adapter.\cl{RS23} The learner's dev
gate accepted all 12 steps, and the examiner found 2 of them not to hold: r0's and r1's third steps,
the two smallest dev margins (13/8 and 7/5).\cl{RS24}

\begin{figure}[tbp]
\centering
\begin{tikzpicture}
\begin{groupplot}[paper, group style={group size=2 by 1, horizontal sep=12mm},
  width=0.48\linewidth, height=5.4cm, xmin=-0.25, xmax=3.25, ymin=0, ymax=260,
  xtick={0,1,2,3}, xticklabels={base,1,2,3}, xlabel={generation},
  ylabel={items right of 889 (exam subset)}]
\nextgroupplot[title={S3: continued adapter (not met)}]
\addplot[lineage r0] table[col sep=comma,x=gen,y=r0] {generated/rsi3_items.csv};
\addplot[lineage r1] table[col sep=comma,x=gen,y=r1] {generated/rsi3_items.csv};
\addplot[lineage r2] table[col sep=comma,x=gen,y=r2] {generated/rsi3_items.csv};
\addplot[lineage r3] table[col sep=comma,x=gen,y=r3] {generated/rsi3_items.csv};
\nextgroupplot[title={S4: retrained from the base (met)}, ylabel={},
  legend style={at={(-0.11,-0.3)}, anchor=north, legend columns=4, /tikz/every even column/.append style={column sep=4mm}}]
\addplot[lineage r0] table[col sep=comma,x=gen,y=r0] {generated/rsi4_items.csv};
\addplot[lineage r1] table[col sep=comma,x=gen,y=r1] {generated/rsi4_items.csv};
\addplot[lineage r2] table[col sep=comma,x=gen,y=r2] {generated/rsi4_items.csv};
\addplot[lineage r3] table[col sep=comma,x=gen,y=r3] {generated/rsi4_items.csv};
\legend{lineage r0, lineage r1, lineage r2, lineage r3}
\end{groupplot}
\end{tikzpicture}
\caption{Items right on the held-out exam per generation and lineage, as the examiner measured them; the
exam holds items the base did not answer when it was selected, so these are counts on that hard
subset, not an accuracy on Lua programs. In S3 two lineages stop rising at generation~3 (r0 flat, r1 down); in S4
every lineage rises at every generation. The base level differs between the panels because the base
was served at a different KV tier (\cref{tab:rsi}).}\label{fig:rsi-items}
\end{figure}
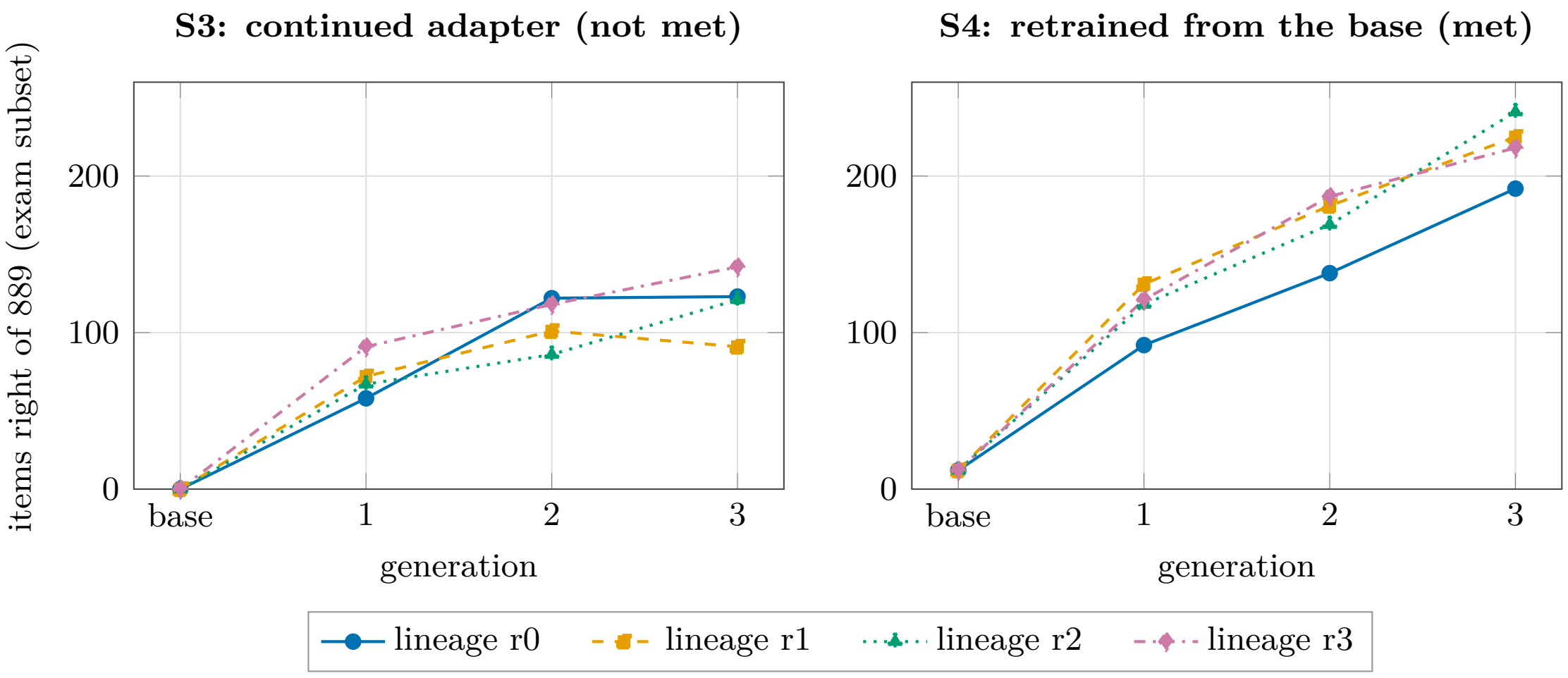

\paragraph{Serving configuration.} The examiner served S4's models with an f32 KV cache, at which the base
answers 12 exam items; S3's stored baseline was served at the build's default tier, at which it answered
none, so S3's and S4's levels are not comparable, only their steps are.\cl{RS27}

%% file: sections/03a-loop.tex

\section{The learning loop}\label{sec:loop}\label{app:loop}

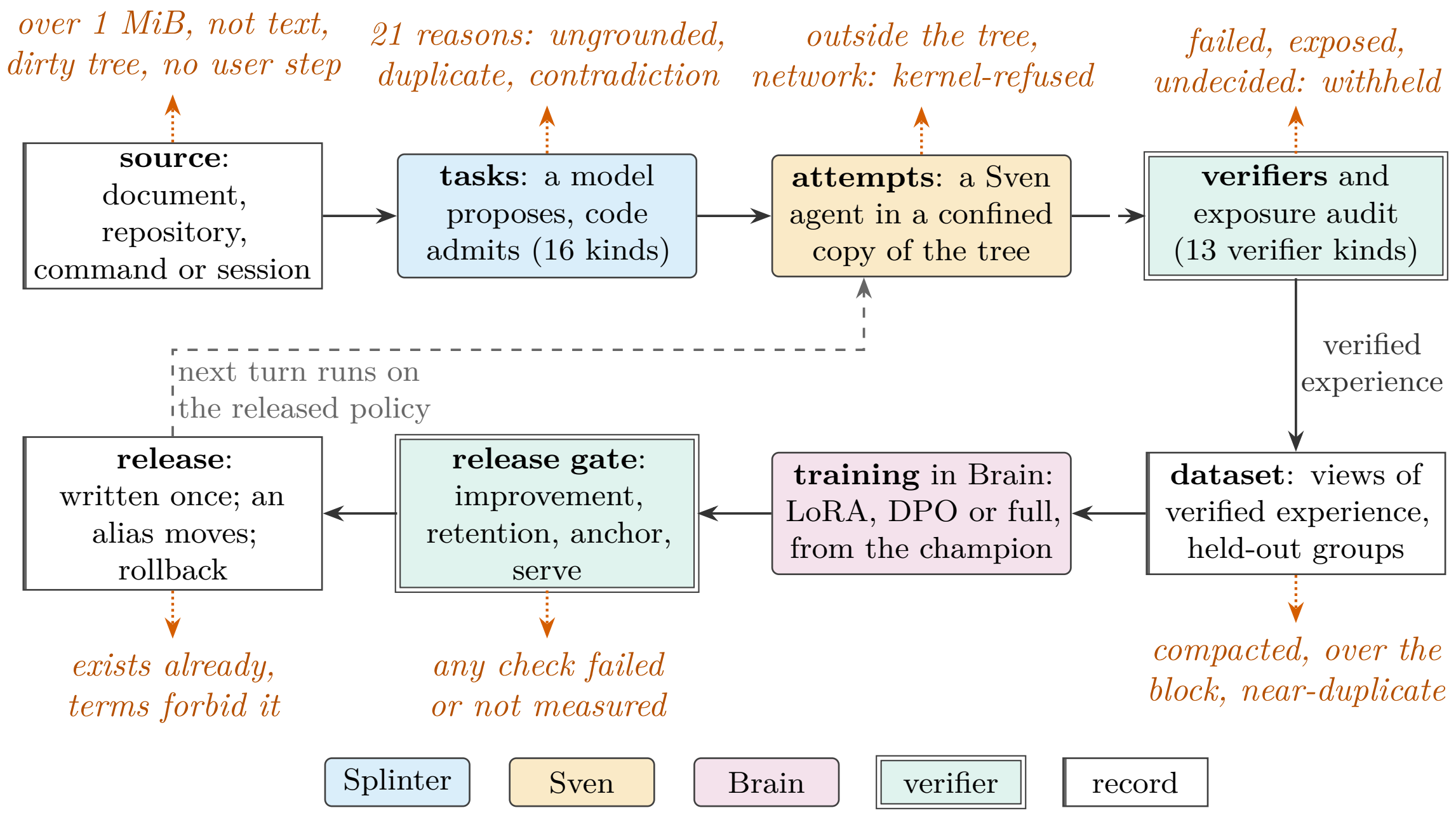
\begin{figure}[tbp]
\centering
\begin{tikzpicture}[fig]
\tikzset{cell/.style={text width=29mm}}
\def\dx{39mm}
\node[data, cell] (src) at (0,0) {\textbf{source}:\\document, repository,\\command or session};
\node[proc=okSky!22, cell] (task) at (\dx,0) {\textbf{tasks}: a model\\proposes, code\\admits (16 kinds)};
\node[proc=okOrange!22, cell] (roll) at (2*\dx,0) {\textbf{attempts}: a Sven\\agent in a confined\\copy of the tree};
\node[verif, cell] (ver) at (3*\dx,0) {\textbf{verifiers} and\\exposure audit\\(13 verifier kinds)};
\node[data, cell] (data) at (3*\dx,-31mm) {\textbf{dataset}: views of\\verified experience,\\held-out groups};
\node[proc=okPurple!22, cell] (train) at (2*\dx,-31mm) {\textbf{training} in Brain:\\LoRA, DPO or full,\\from the champion};
\node[verif, cell] (gate) at (\dx,-31mm) {\textbf{release gate}:\\improvement,\\retention, anchor,\\serve};
\node[data, cell] (rel) at (0,-31mm) {\textbf{release}:\\written once; an\\alias moves; rollback};
\draw[flow] (src) -- (task);
\draw[flow] (task) -- (roll);
\draw[flow] (roll) -- node[lab, right, pos=0.5] {} (ver);
\draw[flow] (ver) -- node[lab, right] {verified\\experience} (data);
\draw[flow] (data) -- (train);
\draw[flow] (train) -- node[lab, above] {} (gate);
\draw[flow] (gate) -- (rel);
\draw[ctrl] (rel.north) -- node[lab, right, align=left] {next turn runs on\\the released policy} ++(0,9mm) -| ($(roll.south)+(-6mm,0)$);
\draw[refuse] (src.north) -- ++(0,5mm) node[reflabel, above, align=center] {over 1 MiB, not text,\\dirty tree, no user step};
\draw[refuse] (task.north) -- ++(0,5mm) node[reflabel, above, align=center] {21 reasons: ungrounded,\\duplicate, contradiction};
\draw[refuse] (roll.north) -- ++(0,5mm) node[reflabel, above, align=center] {outside the tree,\\network: kernel-refused};
\draw[refuse] (ver.north) -- ++(0,5mm) node[reflabel, above, align=center] {failed, exposed,\\undecided: withheld};
\draw[refuse] (data.south) -- ++(0,-5mm) node[reflabel, below, align=center] {compacted, over the\\block, near-duplicate};
\draw[refuse] (gate.south) -- ++(0,-5mm) node[reflabel, below, align=center] {any check failed\\or not measured};
\draw[refuse] (rel.south) -- ++(0,-5mm) node[reflabel, below, align=center] {exists already,\\terms forbid it};
\node[proc=okSky!22, text width=14mm, minimum height=5mm, inner sep=1.5pt] (k1) at (0.6*\dx,-59mm) {Splinter};
\node[proc=okOrange!22, text width=14mm, minimum height=5mm, inner sep=1.5pt, right=4mm of k1] (k2) {Sven};
\node[proc=okPurple!22, text width=14mm, minimum height=5mm, inner sep=1.5pt, right=4mm of k2] (k3) {Brain};
\node[verif, text width=14mm, minimum height=5mm, inner sep=1.5pt, right=4mm of k3] (k4) {verifier};
\node[data, text width=14mm, minimum height=5mm, inner sep=1.5pt, right=4mm of k4] (k5) {record};
\end{tikzpicture}
\caption{The self-learning loop, read clockwise from the source. Every stage keeps what passes and
refuses the rest with a named reason (dotted arrows; \cref{sec:intake,sec:admission,sec:verify,sec:dataset,sec:gate}
give the full lists). Splinter owns tasks, verification, datasets, the gate and releases; Sven runs
the attempts; Brain trains and serves. The dashed arrow is the next turn: attempts run on the policy
the release alias points at.}\label{fig:arch}
\end{figure}

Three projects divide the work, and the division is enforced by a layering file and a gate rather than
by convention: \textbf{Sven} runs agents and keeps the record of what an agent did, \textbf{Brain}
computes (inference, training, checkpoints, adapters, serving), and \textbf{Splinter} decides what is
learned, which experience qualifies as training data, what to train and whether to release
it.\cl{S05,S01,S02} \Cref{fig:arch} shows one turn. A source is captured with its provenance; a model
proposes tasks and code admits or refuses each for a named reason; an agent attempts them in a confined
copy of the tree; verifiers that the policy cannot reach keep what they confirm and an exposure audit
withholds attempts that touched hidden material; verified experience is projected into training records
with whole held-out groups; Brain trains an adapter or every weight from the current champion; and a
release gate admits the candidate only if four checks on material it was not trained on pass
(improvement by a paired sign test, retention of earlier releases, an anchor of general tasks, and
agreement of the served model with the trained one). Releases are immutable, an alias moves, and
rollback moves it back.\cl{A03,V02,L08,G01,X02,X06} Every stage is specified, with its ledger, in
\cref{app:system}; this section is what the experiments below need.

In the experiments of \cref{sec:rsi} the loop is reduced to one task kind whose answers an interpreter
fixes, the verifier is the interpreter, and the release gate is replaced by the learner's own dev gate
and, for the registered decision, by a separate examiner program. In \cref{sec:coding-summary} the full
loop runs on repository tasks.

%% file: sections/03-architecture.tex

\section{Division of labour}\label{sec:architecture}\label{app:system}

The flywheel is a composition of three projects with one rule of ownership.
\textbf{Sven} runs agents (tools, sessions, delegation) and keeps the authoritative
record of what an agent did. \textbf{Brain} computes: inference, training algorithms,
checkpoints, adapters and serving. \textbf{Splinter} decides what is learned, which
experience qualifies as training data, what to train and whether to release the
result.\cl{S05} Splinter's repository states that neither Sven nor Brain depends on
it, so each can demonstrate its own capabilities without it and the released artifact
is an ordinary Brain adapter.\cl{S05,X05}

The boundary is enforced by a machine-readable layering file rather than convention.
Splinter's crates are placed in ten tiers from a dependency-free foundation to
standalone samples; a crate may depend only on a strictly lower tier.\cl{S01} Only the
model adapter crate may depend on Brain, only the agent adapter and the model adapter
may depend on Sven, and the command line and every sample depend on the SDK and
nothing else of Splinter's.\cl{S02,S03,S04} A model is never named by a stage: stages
ask for one of seven roles (policy, teacher, generator, planner, judge, critic,
router) and configuration decides which checkpoint plays it.\cl{S06} Experience is
immutable; verdicts, rewards, step labels and dataset membership are facts derived
about it, never written into it.\cl{S07}

\section{Source intake}\label{sec:intake}

A source is content addressed (a blake3 digest over its canonical origin and parts) and
has one of four origins: a document, a repository, a command, or a recorded
session.\cl{I01} A document is one UTF-8 text file of at most 1~MiB; larger or
non-text files are refused.\cl{I02} A directory is captured file by file, ignoring
\code{.git}, \code{target} and \code{node_modules}, and every file left out is recorded
with a reason.\cl{I03} A command is run without a shell, with a 30~s timeout, a 1~MiB
output cap and an environment limited to \code{PATH}, \code{HOME} and \code{LANG}, so
its real output becomes a source with provenance.\cl{I04} A repository revision is
archived as a snapshot of its tracked files and a dirty tree is refused.\cl{I05} A
recorded agent session is an ATIF trajectory of at most 64~MiB; unknown ATIF keys and
sessions without a user step are refused; secrets are removed before storage and only
the fact of the removal is kept.\cl{I06} On the command line, \code{source add} takes a
file, a directory, or \code{cmd:} followed by a command.\cl{I09}

Text is cut into sections that are byte ranges: Markdown at headings outside code
fences, other text at blank lines. Evidence points at a section by content digest and
byte offsets, so a task's grounding can be resolved back to exact source bytes.\cl{I07}
A generator model is shown three sections per request.\cl{I08}

\section{Tasks and admission}\label{sec:admission}

\paragraph{Who decides.} Models propose; code admits.\cl{A03} The built-in catalogue
holds model-writable kinds (recall, taught, arithmetic, format, advise, converse,
explain, predict, construct, debug, counterexample, transform, classify, retrieve,
multi-turn, combine); repository-change tasks are made by code and never proposed by a
model.\cl{A01} One kind, \code{denoise}, is built by code from the source itself: a
window of 12 words with one word dropped and two neighbours swapped, graded by exact
match, so it needs no generator model.\cl{A02}

\paragraph{The funnel and its rejection reasons.} A model-written task is refused for
one of twenty-one reasons: malformed, no\_reply, over\_count, unknown\_section,
quote\_not\_found, too\_few\_sections, unresolved, missing\_material,
material\_not\_shown, not\_self\_contained, wrong\_form, ungrounded, not\_quoted,
checks\_failed, test\_not\_admitted, duplicate, near\_duplicate, reference\_changed,
no\_subject, contradiction, invalid.\cl{A04} Three deserve explanation. \emph{Grounding}:
a text answer is admitted only if at least 80\% of the reference's content words and
every number in it occur in the cited evidence.\cl{A05} \emph{Subject}: a task must name
what it is about, so that someone who never saw the source gets exactly one answer, and
is self-contained only if it does not quote a dropped privileged item of eight or more
words.\cl{A06} \emph{Contradiction}: two tasks on one subject whose grounded references
disagree are both refused.\cl{A07} The batch driver gives up on a kind after 24
proposals if fewer than a tenth were admitted, and sends the generator a correction after
three refusals for one reason, at most twice.\cl{A08} Repository tasks have their own
vocabulary of refusals (red or flaky baseline, equivalent mutant, tests not separable
from the change, not restorable, already held and others).\cl{A09}

The task report counts, per kind, tasks admitted and rejected, rejections by reason,
every rejection with its detail, and each model call with its pre-admission candidate
count, so the funnel is read from the report rather than reconstructed.\cl{A10}

\section{Verification and leakage control}\label{sec:verify}

\paragraph{Rollouts.} Sven runs the attempt. A closed-book solve builds an engine with
no tools;\cl{R01} a code task gives the solver exactly one tool, \code{run\_code};\cl{R02}
a repository episode uses Sven's coding toolset with automatic approval inside a fresh
copy of the snapshot.\cl{R03} Every request runs under Splinter's single system
prompt,\cl{R04} and a solve refuses to run in an environment other than the one its task
records.\cl{R05}

\paragraph{Experience and verdicts.} An experience is content addressed by the digest of
its canonical JSON.\cl{R06} Verdicts, step labels and relations are annotations
appended beside it.\cl{R07} Verdicts have a strength ordered judged, consistency,
formal, executable, constraint.\cl{R09} The decision rule ignores abstentions, uses only
verdicts of the strongest rank present, and returns no decision if they conflict;\cl{V08}
the reward is derived from the decision and never stored.\cl{R08} Tasks are graded by one
of thirteen verifier kinds.\cl{V01} A verifier sees the output and privileged material but
never the trajectory.\cl{V02} Examples: a stated reference may be matched by an answer at
most four times its length plus 60 characters;\cl{V05} an executable check passes only if
a completion line derived from its digest is seen, so a solution that exits early cannot
pass;\cl{V04} a generated test oracle must kill every one of at most 32 mutants;\cl{V06}
a repository verifier overlays the hidden checks on a fresh copy and fails a patch that
touches a protected path.\cl{V07} A judge model may grade only when calibrated to a
precision of at least 0.9 and never grades its own model's output.\cl{V03}

\paragraph{Leakage.} Four mechanisms separate what is trained on from what is measured.
(i)~Held-out records are whole groups, a tenth of the samples grown to at least eight
units and at most a quarter;\cl{L01} groups are formed by merging texts that share at
least eight distinct runs of eight words, ignoring runs held by more than six
texts.\cl{L02} (ii)~Before grading, every gate task that matches a prompt the candidate
trained on, including replay and rehearsal, is excluded and counted as
\emph{leaked}.\cl{L04} (iii)~For recorded sessions, sealed probes refuse any training
record that contains a probe's question or shares an eight-word run with it.\cl{L05}
(iv)~For repositories, exam families are reserved by stable hash before any task is
generated, and a family needs at least 120 words;\cl{L07} the exposure audit fails an
attempt whose tool calls name a hidden path or a forbidden string, or whose outputs
repeat a 12-token shingle of hidden content;\cl{L06} exposed attempts and harness
failures are withheld and counted in no denominator.\cl{L08}

\section{Datasets and corpus expansion}\label{sec:dataset}

\paragraph{Projection.} Datasets are projected from verified experience by named views
and six objectives (SFT, classification, DPO, contrastive, reward, continued
pre-training).\cl{D01} SFT is written as Brain's \code{generic-messages-v2} and DPO as
\code{generic-preference-v1}.\cl{D02} A manifest stored beside the file names the views,
experiences, tasks and sources every record came from.\cl{D03} The student's turn is the
instruction alone: privileged teacher context is stripped by default;\cl{D04} only the
assistant action is supervised, system and user turns carry no loss;\cl{D05} compacted or
pruned trajectories are excluded rather than repaired.\cl{D06} The training set is capped
per concept (0.25), task kind (0.5) and verification strength (0.75), after
deduplication by the same near-duplicate rule as admission,\cl{D07,C03,C04} which compares 3-word shingles and refuses a Jaccard overlap of at least 0.8.\cl{L03} Replay sampling is deterministic in the content digest and the seed.\cl{D09} A monitoring
split of whole families (at least three, default share 0.1) supports early
stopping.\cl{D08}

\paragraph{Packing.} The step view can also write a trajectory as one record instead of one record
per action: the conversation up to its last admitted action, with every admitted action supervised
and every other message as context. It supervises exactly what the per-action records supervise
together, with the context rendered once instead of once per action, and within a step each supervised
token weighs the same in both forms.\cl{D10} What differs is the unit a step counts: a step of eight
packed records spans eight whole trajectories rather than eight actions, so the same step count moves
the adapter on far more tokens. The choice of form is reported with every run that uses it
(\cref{sec:coding-results}).

\paragraph{Expansion beyond the first pass.} The corpus grows by five mechanisms, each
with its own gate. (i)~\emph{Variants}: a model-written task may be given up to three
differently worded questions that keep its reference, evidence and verifiers; a variant
that changes the reference is rejected, and variants are only measured, never trained
on.\cl{C01,C02} (ii)~\emph{Teacher solves}: a task the policy never solves is solved
once more by a teacher shown the source sections it is grounded in, and the answer is
kept only if the task's verifiers pass it.\cl{C08} (iii)~\emph{Retention queue}: when a
retention check fails, the forgotten concepts are queued and their sections receive new
tasks on the next run.\cl{C05} (iv)~\emph{Rehearsal}: half code-built form tasks and
half general requests the base writes, eight per request, answered greedily by the base;
requests matching the anchor are excluded and repeats refused.\cl{C07} (v)~\emph{Claims
from sessions}: per claim, a model writes twelve paraphrases of which four are held out
for stopping, plus four statement, two reverse and two implication forms;\cl{C10} a claim
of kind procedure is stored and never trained on.\cl{C11} Mastery of a concept is its
pass rate over the last 32 decided attempts and the five weakest are listed.\cl{C06}
Exam and development families are reserved before generation (50 each by default, 8
tasks per exam family, 4 per development family).\cl{C09} We found no mechanism that
generates new tasks from failed attempts other than the retention queue and the teacher
solve.

\paragraph{Claims intake.} A claim is a correction, a fact or a procedure;\cl{K01}
extraction keeps only claims that each of two passes produced.\cl{K01} A claim is admitted
by code only when every quote is a verbatim substring of the person's own message at the
cited step,\cl{K02} and every number, quoted term and name in its statement occurs in the
cited words;\cl{K03} eleven refusal codes are defined;\cl{K04} a ruling is admitted
(superseding earlier claims), reinforced, forgotten or refused.\cl{K05}

%% file: sections/04-training-release.tex

\section{Training}\label{sec:training}

Splinter chooses the objective and the data; Brain runs the optimisation. LoRA
supervised fine-tuning is Brain's \code{ChatFineTune} and DPO is its
\code{PreferenceFineTune}; both are called from the model adapter crate.\cl{T01} The
defaults are a learning rate of $1\times10^{-4}$, rank 8, alpha twice the rank and no
weight decay;\cl{T02} the step budget is three passes over the records clamped to between
40 and 2000 steps.\cl{T03} A candidate \emph{continues the champion}: its training starts
from the parent release's adapter,\cl{T06} mixes in a replay of each earlier release's
records (a quarter of each earlier release's records, entering a quarter of training draws),\cl{T04} and mixes in a rehearsal of the base's own
answers (a quarter of draws, half of it code-graded form tasks).\cl{T05} A monitored run
keeps the best evaluation's adapter.\cl{T07}

\paragraph{Consolidation.} A policy that keeps learning would grow its adapter, or must
retrain from scratch on every claim. Splinter therefore consolidates a generation into
the weights in one of two modes: a full fine-tune of every weight of the generation's
base, or merging the adapter into the base.\cl{T08} The full mode defaults to a learning
rate of $1\times10^{-5}$ for six epochs with a retention replay share of a quarter;\cl{T09} a
generation is consolidated automatically after more than 2000 trained records.\cl{T10}
Both are carried out by Brain (\code{ChatFullFineTune}, \code{merge\_adapter}).\cl{T11}

\section{Release gate, releases, rollback and lineage}\label{sec:gate}

\paragraph{The gate.} A candidate is released only if four checks all pass, each over
material it was not trained on.\cl{G01} A check that was not measured fails with the
reason \emph{not measured}.\cl{G02}
\emph{Improvement}: a one-sided exact paired sign test at $\alpha=0.05$ over discordant
clusters, a family of items counting once; the candidate must also have solved at least as
many of the suite's assigned tasks as its champion, an attempt withheld as exposed or as a
harness failure counting as unsolved for the arm that produced it.\cl{G03}
\emph{Retention}: accuracy on every earlier release's held-out suite may not drop by more
than 0.05.\cl{G04}
\emph{Anchor}: accuracy on the frozen suite of general tasks may not drop by more than
0.02; the report carries a 95\% bootstrap interval over 4000 resamples.\cl{G05}
\emph{Serve}: plain \code{brain serve} must report the candidate's adapter digest and
agree with the in-process answers on at least 75\% of up to eight re-answered held-out
tasks.\cl{G06} Repository exams use the same four checks, with the exam families as the
improvement suite and agreement judged on the graded verdicts.\cl{G07} The bootstrap seed
is drawn when not given and recorded;\cl{G08} a failed retention check queues the
forgotten concepts;\cl{G09} and only a candidate trained from the current champion can
replace it.\cl{G10}

\paragraph{Releases.} A release manifest (format \code{splinter-release-v4}) records the
artifact, parent, candidate, datasets, replay, rehearsal, training settings, the whole
gate report, metrics, terms, provenance and generation.\cl{X01} A release is written
once and an existing one is refused;\cl{X02} unrestricted distribution is refused unless
the terms of everything it was made from permit it.\cl{X03} The default policy is the alias
\code{policy:default}; moving an alias is a compare-and-set from the release the gate
measured against.\cl{X04} After every move the serving manifest
\code{<state>/serving/<alias>.json} is rewritten, so \code{brain serve --adapter-manifest}
follows the release without a restart.\cl{X05} \code{splinter rollback} moves the alias to
the release the current one was trained from and refuses when there is none.\cl{X06}
\code{splinter lineage} walks any artifact up to its source bytes and down to everything
derived from it, computed from the stores on every call.\cl{X07}

%% file: sections/05-substrate.tex

\section{Training a 27B model on one card}\label{sec:substrate}

Training a 27-billion-parameter model on one 95~GiB card needs the frozen base in a
reduced-precision tier, activation recomputation, a loss that never materialises the
full logits, an attention kernel that supports the model's head dimension, and a loader
that never holds the checkpoint in host memory. This section states what Brain's main
tree contains for each, what its qualification runs measured for bf16 LoRA and NF4
QLoRA, and what a later throughput measurement of the same architecture shows. What a
trained 27B adapter changes in what the model does is measured in
\cref{sec:coding-results}.

\subsection{What the tree contains}

The Splinter trainer's seam is the Qwen3-family trainer.\cl{B01} In that trainer the
frozen base can be held in bf16,\cl{B02} each layer is recomputed in the backward pass
from its saved residual,\cl{B03} and the logits are sized in row chunks with a chunked
weighted cross-entropy.\cl{B04} Brain has a causal grouped-query flash-attention forward
and backward supporting head dimensions up to 256, with finite-difference gradient
checks at head dimensions 40, 128 and 256.\cl{B05,B10} For the 27B model's crate it has
a streaming FP8 source that dequantises in 128-row bands and a base-dtype enumeration
with F32, bf16 and NF4 members,\cl{B06} and an NF4 weight type with a 16-entry lookup
table.\cl{B07} The numerical gates are \code{make test/gradcheck} (finite-difference
checks of the Qwen3.5 architecture including LoRA) and \code{make test/parity}, which
runs the checks on the CPU backend.\cl{B08,B09}

\subsection{Qualification runs}\label{sec:qual}

Brain's qualification at commit \mEFourCommit{} ran the 27B architecture on one 95~GiB
card; \cref{tab:qual} lists the runs. Probes use a randomly initialised model of the real
width and depth (64 layers), recomputation, rank-16 LoRA and one step, so their loss is not
a quality number. With an NF4 base, 8192 tokens fit in
\mEFourProbeNfFourEightOneNineTwoPeakMib{}~MiB at
\mEFourProbeNfFourEightOneNineTwoTokensPerS{} tokens/s;\cl{M11} with a bf16 base,
8192 tokens took \mEFourProbeBfOneSixEightOneNineTwoPeakMib{}~MiB and 16384 tokens
\mEFourProbeBfOneSixOneSixThreeEightFourPeakMib{}~MiB, both at
\mEFourProbeBfOneSixOneSixThreeEightFourTokensPerS{} tokens/s.\cl{M12,M13}
Through the command line the real FP8 checkpoint trained one step with an NF4 base and
with a bf16 base with a host peak of \mEFourCliNfFourOnestepHostRssMib{} and
\mEFourCliBfOneSixOnestepHostRssMib{}~MiB respectively\cl{M14,M15} and, with an NF4 base,
40 steps drove the loss on eight training examples from
\mEFourCliNfFourFourZerostepLossFirst{} to \mEFourCliNfFourFourZerostepLossLast{}.\cl{M16}
That is a statement about fitting the training examples, not about generalisation: the
validation examples were two of the training examples. The command line's default
batch ran the device out of memory at 1024-token rows.\cl{M17}

\begin{table}[tbp]
\centering
\small
\input{generated/e4_table}
\caption{Qualification runs of the 27B architecture on one card (Brain's qualification
logs, copied unmodified apart from path names; their values, digests and commands are in \code{data/measurements.json}).}\label{tab:qual}
\end{table}

\subsection{Throughput as measured on a later Brain revision}\label{sec:throughput}

The qualification rates above are those of the commit it ran at. A later probe of the
same architecture (random weights, 64 layers, rank 16, NF4 base, 8192 tokens, recomputation)
on the idle card measured \mThrNfFourEightOneNineTwoTokensPerS{} tokens/s at steps two and
three, a step of \mThrNfFourEightOneNineTwoStepS{}~s and a peak of
\mThrNfFourEightOneNineTwoPeakMib{}~MiB.\cl{M38} Brain's own record of the real checkpoint
at 8192 tokens states 635 tokens/s with an NF4 base, and 640 tokens/s with a bf16 base at
79.8~GiB;\cl{B13,B14} we did not rerun those two, and the same record states that
1000 tokens/s was not reached.\cl{B15} At the measured probe rate one million trained tokens
take \mThrHoursPerMillionTokens{}~h of card time, computed as $10^6/(622\cdot3600)$.\cl{C12}
We state the measurement and not a target; a rate is only valid for the revision, the
card and the sequence length it was taken at.

\subsection{What this does not show}

That a trained 27B adapter changes what the model generates. The qualification logs predate
a generation path for such an adapter; Brain has since added one (the \code{sample-adapter}
verb of its \code{qwen35} command)\cl{B12} and keeps its own record of a run, which we did
not rerun or verify here. No run beyond 16384 tokens and no NF4 run at 16384 tokens on the
qualification commit exists.

%% file: generated/e4_table.tex
\begin{tabular}{llrrrr}
\toprule
Run & Base & Tokens & Peak MiB & Seconds & Outcome \\
\midrule
probe, 64 layers, random init & NF4 & 8192 & 35947 & 404.74 & step done \\
probe, 64 layers, random init & bf16 & 8192 & 67011 & 435.47 & step done \\
probe, 64 layers, random init & bf16 & 16384 & 77879 & 873.64 & step done \\
27B, 1 step & NF4 & 1024 & 27989 & 460.4 & finished \\
27B, 1 step & bf16 & 1024 & 59053 & 321.1 & finished \\
27B, 40 steps & NF4 & 256 & 24559 & 936.8 & finished \\
27B, default batch & NF4 & 32768 & 92571 & 322.8 & out of memory \\
\bottomrule
\end{tabular}

%% file: sections/06-requirements.tex

\section{What the loop needs in order to work}\label{sec:requirements}

A turn of the loop adds a capability only if its stages cannot be satisfied by an artefact of
the setup. This section states the requirements we found necessary, the mechanism that
implements each, and which are not yet implemented. Each names its commit or file in the
ledger, which checks that the commit is in the history of the pinned revision of its
repository (\cref{sec:repro}).

\paragraph{Held-out units must be independent.} Records carry the group of the source text they
are grounded in, and a held-out measurement needs whole groups. A dataset that is one source
text cannot be divided, so the split refuses it with a typed error naming the cause instead of
dividing the group and failing the leakage check afterwards; a free record follows the fixed
side of its group, monitoring with no whole group to set aside is skipped with the reason
recorded, and a source that is one independent text is refused at intake before any rollout is
paid for.\cl{FX01,FX05} A source of independent units is therefore one file per unit.

\paragraph{Admission must see through markup.} The subject of a task is compared with the
instruction, the source and the variants through one function that ignores code quotes,
quotation marks, emphasis and a call's empty parentheses, so that ``the \code{q} key'' contains
the subject ``q key''.\cl{FX02} A number written with a thousands separator is the same number
in the stated-reference check.\cl{FX06}

\paragraph{The serve check must start the real server.} The gate's serve check runs
\code{brain serve --adapter}. A candidate's card id must follow Brain's adapter grammar
(base, owner, name, tag), and the check must pass the bank option that covers every linear
Splinter's LoRA corrects; the test double of the server refuses an adapter offered without it,
because a double that accepts any flags hides both.\cl{FX03,FX04}

\paragraph{The anchor bound must count net lost answers on a large enough suite.} A rate bound of
0.02 on a 40-question anchor suite allows no lost answer, since one net loss is 0.025. The gate
can decide the anchor on answers lost net of answers gained (an option; the default rate bound
is unchanged), and a 200-question anchor suite built by code and a reviewed list
exists.\cl{FX07}

\paragraph{Retention must test acquired capabilities.} The retention suite of each earlier
release is built from that release's held-out split, and the exams of workspace families are
added to it.\cl{OP01} For a domain of facts the held-out split holds facts the release never
taught, so a passed retention check is not evidence that taught facts were kept. A retention
exam drawn from what each release demonstrably acquired is not implemented; the
registered coding-task retention check (\cref{sec:protocol}) uses the exam families of every
earlier round for this reason.

\paragraph{The verifier must prove that the tests ran.} The workspace verifier passes a patch
only if every fail-to-pass and pass-to-pass command exits with status 0.\cl{OP02} That does not
prove that the intended tests ran, so a suite that exits zero without running them would pass.
At the current head a command that carries a test expectation passes only when the runner's own
report shows every expected test run and passed, with readers for \code{python3 -m unittest} and
\code{cargo test}; a command without an expectation is still judged by its exit status.\cl{OP05} The
coding-task experiments register conformance measurements for the verifier before they count
(\cref{sec:protocol}).

\paragraph{The gate must carry canaries.} A gate with an empty canary set skips that check. At the
revision the paper's architecture claims are pinned to, the release gate called from \code{practice}
was built without canary tasks;\cl{OP03} at the current head it is handed up to 20 canaries made from
the exam the run reserved, so a release from \code{practice} carries that evidence.\cl{OP04}

\subsection{Trajectory and training requirements of repository tasks}\label{sec:audit}

Training an agent on its own repository trajectories adds requirements to the stack. Each is
located in source; the ledger names the file, commit or test.

\begin{enumerate}
\item \textbf{Loss masking of tool conversations} (Brain; implemented). Under the Qwen3 template
an empty reasoning block is added to the last assistant turn only, so non-final assistant
turns of a multi-step conversation failed the trainer's boundary check. Tool-calling
conversations now train per assistant turn in both thinking modes.\cl{AU01}
\item \textbf{Tool definitions lost between recording and training} (implemented). The agent
records the tool definitions of the run, but the trajectory projections dropped them; an
ATIF trajectory's tool definitions now become the chat sample's tools, and Splinter's
dataset check covers tools.\cl{AU02}
\item \textbf{Trajectory fidelity in the agent SDK} (Sven; implemented). The exported trajectory
marked no context compaction and kept the uncompacted history, listed every registered tool
rather than those offered, and turned unparseable tool-call arguments into an empty object.
Compactions are now marked and the compacted history kept, the offered tools are exported,
and the raw text of non-JSON arguments is kept.\cl{AU03}
\item \textbf{The record is the whole conversation} (Sven; implemented). After a context
compaction the export was rebuilt from the compacted history, so a run that compacted lost the
steps the compaction replaced and the model requests recorded on them. In the withdrawn 27B run
the five episodes stopped at the 60-call cap counted 61 or 62 tool calls while their trajectories
held four to 42. The agent now keeps an append-only record beside the history the model reads, a
child run's store keeps a transcript the same way, and a compaction is marked where it fired. The
projection into training conversations still refuses a trajectory that holds a compaction marker,
so such episodes can be analysed and are not trained on.\cl{TJ03,TJ04}
\item \textbf{Sampling identity on every request} (Sven and Brain; implemented). A run's
sampling (temperature, top-$p$, top-$k$, seed) and its request for the sampled tokens reach every
model request of the run and of the children it starts. Brain's chat endpoint returns the prompt
token ids, the sampled token ids, the log-probability of each sampled token under the
distribution it was decoded from, and the seed it used, the adapter, the base-checkpoint digest, the
backend and the engine build; a step of the trajectory records them, and records each request's
own prompt and tokens when a step covers several. A seed the server drew above $2^{63}$ was once
dropped from the reply and refused when sent back.\cl{TJ05,TJ06,TJ07}
\item \textbf{Crash-robust capture} (Sven; implemented). The trajectory can be written one
record at a time as the run happens, with a flush policy and size bounds; a journal that was never
closed reads back marked incomplete, a test kills a real second process inside a model request
and reads the journal back, and Splinter's projection refuses a trajectory marked incomplete.\cl{TJ08,TJ09}
\item \textbf{Children in the parent's record} (Sven; implemented). A harness run's trajectory
embeds every finished child as a trajectory with its tool actions and the account of each of
its model requests, sampling identity included.\cl{TJ10}
\item \textbf{Release probes were closed-book} (implemented). A coding agent is now gated on the
exam its run reserved, with report types for workspace suites.\cl{AU04}
\item \textbf{Contamination paths} (implemented). The semantic-memory tool is withheld from the
worker, and a context tool that could open any readable path is jailed.\cl{AU05}
\item \textbf{Silent waits} (Sven; implemented). A model server that goes quiet is bounded and
the wait is reported, with a stalling mock provider to test it.\cl{AU06}
\item \textbf{Tool-definition mismatches} (Sven; implemented). Definitions disagreed with the
implementation and were tightened.\cl{AU07}
\item \textbf{Serving several adapters} (Brain; implemented in the qwen3 server). An adapter bank
lets one batch serve the base and many LoRA adapters.\cl{AU08}
\item \textbf{Containment} (implemented for commands run on the host; a container remains
the stronger option). A solver's forbidden operations are prevented by the kernel, not
detected afterwards by the exposure audit. A command of the solver (its shell, and the
checks, and a code call of the \code{run_code} tool\cl{CN13}) runs under Landlock with read access to the system and write access to its own tree
and a private directory only, so the withheld tests,\cl{CN01} a path traversal or a symlink
out of the tree,\cl{CN02} the host's credential files,\cl{CN03} and a sibling candidate's
tree\cl{CN06} are refused with a permission error, also when reached through \code{sh -c}, an
interpreter or a compiled helper the candidate brought along.\cl{CN07} The command's
environment is built from an allowlist from which names that look like secrets are dropped even
when the allowlist names them;\cl{CN04} Landlock's ptrace check already refuses the files of another process's
\code{/proc} directory that need ptrace access (its environment among them),\cl{CN05} and the command
runs in a private user and pid namespace with a \code{/proc} of its own, in which no other process
of the host exists at all;\cl{CN14} the network is denied (Landlock's
TCP rules and a socket filter that also covers UDP), so a command cannot reach a model server on the
loopback;\cl{CN08} and a process budget stops a fork bomb when the operator sets one.\cl{CN09} A kernel
without Landlock ABI~4 refuses to run the command rather than run it open. The agent's own file
and search tools are confined to the attempt's tree by the agent framework's project root,
which its delegated children share.\cl{CN10} Only an experience that an independent verdict
passed at executable strength is training evidence.\cl{CN11} The step-level view of the
training data also admits an action a labeller marked good from an experience nobody verified;
that is the default, \code{--control outcome-verified} turns it off, and the dataset manifest
records which was used.\cl{CN15} A path outside the tree that a command names is reported by what came of it (the sandbox refused it, it did not exist, or something came back), so the audit separates a harmless mention from an access; all three still fail it.\cl{CN12} Not covered: the process budget and network opt-ins change the
environment record, so they are off or unset by default to keep the identity of tasks generated
earlier, and a kernel that cannot create a user namespace refuses to run a command unless the
operator allows the host's \code{/proc}.
\end{enumerate}

\subsection{After the first generation: teaching, acquiring, choosing, retaining}\label{sec:after}

Four further requirements follow from asking the loop to run past one repository and one
round. Each is a property of the stack, shown by a specification that runs on scripted agents
and a test double of the trainer; none is a measurement of a trained model. The experiments
that would be are registered in \code{prereg/acquisition.md} (ACQ1, INT1, GEN1)\cl{PL03} and were
not run (\cref{sec:limitations}).

\paragraph{A task nobody solves needs a teacher that is not trusted.} A repository task has no
passage to show a teacher, so \code{practice} used to drop every task the policy never solved:
the frontier keeps only what is solved sometimes. A teacher is now named for those tasks. It is
either the task's own reference change (the inverse of the seeded fault, which the task keeps
and the student is never shown) replayed as the actions of an agent through sven's tools, or a
stronger model solving the task as an agent. Either produces an experience that is graded by the
same workspace verifier and the same exposure audit as the policy's attempts: a demonstration
that does not make the hidden checks pass is not kept, whoever proposed it.\cl{TC01,TC02,TC03} A
replayed demonstration names the file without searching for it, so it teaches the edit once the
place is known and not the localisation; that is a stated limit.\cl{TC04}

\paragraph{The probes must exist before the training and decide nothing in it.} A run that
turns documentation into skills measures each skill on probes written from the skill's own
sections and checked to exercise it. The probes are split: the dev half measures the baseline,
each round's gain and what a round cost the skills measured before, and the final half is
touched only by the closing assessment, which compares the policy the run began with and the one
it ends with. No training task repeats a probe, and no teacher answers one.\cl{AQ01,AQ02} A
candidate that learns nothing is rejected, counted against its skill, and leaves the policy as it
was.\cl{AQ03} What a planner model proposes is kept only as far as the text bears it out: a quote
that is not in a section the skill cites, a symbol that is not in them, a prerequisite that is
no skill or closes a cycle are refused with the reason.\cl{AQ04}

\paragraph{An intervention is chosen by what it was measured to do, net of what it costs.} The
choice among more samples, another episode protocol, a trained candidate and no change rests on
the lower confidence bound of the paired gain over families, priced against what a success is
worth, and a mechanism that is not available is reported as not measured, never as a gain of
zero.\cl{IV02} On four synthetic clusters whose outcomes are drawn from a known world, the choice
differs by cluster and its realised value exceeds every strategy that applies one intervention to
all four.\cl{IV01}

\paragraph{Retention is measured on every generation, not inferred.} A generation of the coding
policy measures the champion on every protected task before it trains and the candidate and the
champion on the same tasks at the same settings; the full matrix has the base as generation 0 and
every generation on every protected task, so backward and forward transfer have measured
cells.\cl{GN01,GN02} A candidate that loses a protected family beyond a limit, or does not gain
on the new exam, is rejected before the gate is asked, the alias stays, and the rejection is
recorded with the numbers it rested on.\cl{GN03} Replay is a policy rather than a constant: none,
uniform, balanced over capabilities, weighted by the champion's failures, or a rebuild from the
base on everything accumulated.\cl{GN04}

\paragraph{A language's version is part of the environment.} The Lua families and the \code{lua}
runtime ran on whatever \code{lua} the host's alternatives named. The runtime now refuses an
interpreter whose version line is not Lua 5.1, and the interpreter is built from its checksummed
release tarball.\cl{LU01}

%% file: sections/04b-coding-summary.tex

\section{Repository tasks with a stronger teacher: the confirmatory query}\label{sec:coding-summary}

\paragraph{Question and design.} Can verified engineering-agent experience improve completion of
unseen repository tasks?\cl{PR09} The registered test (\cref{sec:protocol}, registered before any run)
trains an 8B student on the verified trajectories a 27B teacher produced on development families and
compares it with the untrained 8B on a frozen qualification set of 46 task families, each task solved
at three registered decoding seeds, by a one-sided exact sign test over families at
$\alpha=0.05$.\cl{PR15,PR04}

\paragraph{Outcome: not shown.} The candidate registered before the query, the decisive-action adapter
of training seed 13, solved 23.0 of the 46 tasks on average over the three seeds against 20.0 for its
base, winning 18 families, losing 12 and tying 16 ($p=0.18$): the candidate is not demonstrated better
than its base.\cl{QUAL1} It had been chosen on a six-seed development read in which it solved 24.5
against the base's 19.67 ($p=0.040$), so part of that margin was selection.\cl{PROTO6} A secondary
contrast did separate: against its all-action control, trained on the same corpus with the same recipe,
seed and binary but supervising every action instead of only file edits and writes, it won 26 families
and lost 11 ($p=0.010$). The control itself solved 16.33, below the base, so the contrast shows that
supervising every action does harm rather than that supervising decisive actions helps.\cl{QUAL1}

\paragraph{What else ran.} Before the query, pilots of the headline and of a 27B model trained on its
own trajectories showed no held-out improvement on a 23-task subset, and the base solved twice at two
protocol seeds differed from itself by as much as either adapter did from it.\cl{M47,M52,M53,M56,M57}
\Cref{app:coding} holds the protocol, the pilots, the development probes and the run-to-run variation
they exposed.

%% file: sections/07-coding-protocol.tex

\section{The coding-task experiment: protocol}\label{sec:protocol}\label{app:coding}

The question that matters to a team, once the loop of
\cref{sec:architecture,sec:admission,sec:verify,sec:dataset} exists, is:
\emph{given a git repository, can a local agentic model be tuned to work competently in
it, repeatedly and with further repositories, and does it get better at what it has met
rather than only at the newest?} This section states the protocol that asks it. It was
registered before any run (\code{prereg/coding-tasks.md}; the changes made since are listed in \cref{app:deviations}); everything
in it is a registered rule or a statement about the code; the coding-task measurements are
reported in \cref{sec:coding-results}.\cl{PR00}

\begin{figure}[tbp]
\centering
\begin{tikzpicture}[fig]
\tikzset{cell/.style={text width=20.5mm}}
\def\dx{26.5mm}
\node[verif, cell] (a) at (0,0) {baseline\\passes twice};
\node[proc=okSky!22, cell] (b) at (\dx,0) {apply the\\seeded fault};
\node[verif, cell] (c) at (2*\dx,0) {a selected\\test fails};
\node[verif, cell] (d) at (3*\dx,0) {the failure\\reproduces};
\node[verif, cell] (e) at (4*\dx,0) {undoing it\\restores green};
\node[data, cell] (f) at (5*\dx,0) {admitted:\\hidden files\\sealed};
\foreach \x/\y in {a/b,b/c,c/d,d/e,e/f} \draw[flow] (\x) -- (\y);
\draw[refuse] (a.south) -- ++(0,-6mm) node[reflabel, below, align=center] {red or flaky:\\the run stops};
\draw[refuse] (c.south) -- ++(0,-6mm) node[reflabel, below, align=center] {no failing\\test: refused};
\draw[refuse] (d.south) -- ++(0,-6mm) node[reflabel, below, align=center] {flaky failure:\\refused};
\draw[refuse] (e.south) -- ++(0,-6mm) node[reflabel, below, align=center] {not restorable:\\refused};
\end{tikzpicture}
\caption{Admission of a mutation-seeded repository task, left to right. Every check is run on a fresh
copy of the snapshot, every refusal is counted by its reason, and an admitted task's failing tests are
moved out of the tree the solver receives.}\label{fig:valid}
\end{figure}
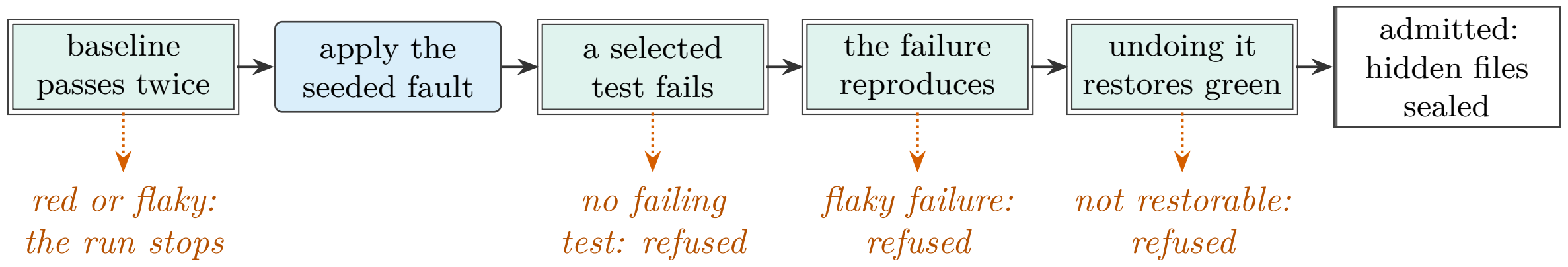

\paragraph{Tasks.} A task is a snapshot of a repository tree at a revision, an
instruction stating a symptom, and privileged material the solver never receives: the
hidden checks (fail-to-pass and pass-to-pass commands) and the paths that carry them.\cl{PR01}
A candidate task is admitted only if the snapshot's selected tests pass twice, a change makes
a previously passing test fail reproducibly, and the reverse change restores green
(\cref{fig:valid}); a red or flaky baseline stops the run.\cl{PR01,PS01} Splinter's repository
tasks are made by code and never proposed by a model, and have their own refusal
vocabulary.\cl{A01,A09} The failing tests are moved out of the tree the solver receives, and a
repair passes when the hidden files are overlaid fresh on the patched tree and the
fail-to-pass and pass-to-pass commands pass; a patch that touches a protected path
fails.\cl{V07}

\paragraph{Solving, recording and trajectory data.} The solver is a Sven coding worker with
file, search and shell tools, a tool-call budget and a repetition stop, started in a fresh
copy of the snapshot. One harness profile is pinned for every solve of every arm and for data
collection alike: at most two follow-up turns, at most sixty tool
calls and a repetition stop at three; a follow-up is given when the visible checks stay red
and also when a turn ends with the tree unchanged, asking the worker to make the change by
calling the edit tools.\cl{R03,PS01} The
recorded unit is the full trajectory; the verdict is produced outside the agent. Only
attempts whose patch passes and whose exposure audit is clean are eligible for training, and
for the headline experiment the attempts are those of a stronger teacher model, not of the
student (\cref{sec:coding-results});\cl{PR15}\cl{L06,L08} each assistant action becomes a training record conditioned on the prior
conversation and on the tool definitions the agent was offered, with system and user turns
carrying no loss, and compacted trajectories are excluded rather than repaired.\cl{D04,D05,D06,AU02}

\paragraph{Validity and leakage.} Four registered rules keep the exam honest. (i)~Families are
assigned to partitions by a stable hash before any rollout, and a family of the qualification
partition is never solved for data purposes.\cl{PR02,L07} (ii)~Hidden files are absent from the
solver's tree and are materialised only in the verifier's.\cl{V07} (iii)~An exposure audit fails
an attempt whose tool calls name a hidden path or whose outputs repeat a shingle of hidden
content, and exposed attempts and harness failures are withheld and counted in no
denominator of the paired test; in the release gate's improvement check they count as tasks
the arm did not solve, so a candidate cannot remove a loss from the test.\cl{L06,L08,PR06} (iv)~The isolation level of every run is recorded; with
\code{none} a host command is still confined by the kernel (\cref{sec:audit}) and the audit
is a second line, and a task whose soundness needs a container is not admitted.\cl{PR06,AU09,CN01} Whether a repository is in the base model's
pretraining data is not known; the base arm on the same exam bounds that effect without
removing it.

\paragraph{Units, arms and the paired exam.} The unit of analysis is the family, the file a
fault was seeded in for the authored repositories and the subsystem for real
ones.\cl{PR02} Tasks within a family are not independent, so no test treats them as such. The
base arm is the unmodified checkpoint in a pinned harness profile; the adapter arm is the same
checkpoint with the adapter attached; system prompt, tool schemas, budgets, decoding (the model family's recommended sampling, each task solved under three registered seeds and pass@1 averaged over them, thinking off as the primary variant and a 1024-token thinking budget as the secondary one) and environment image are identical.\cl{PR03} A family is won by the adapter
when its resolved fraction is higher, lost when lower, and dropped when tied. The primary
decision is a one-sided exact sign test over the non-tied qualification families at
$\alpha=\alphalevel$, on the adapter served through Brain and driven by Sven; a result that is
not significant is \emph{not shown}, never equivalence.\cl{PR04,G03}

\begin{table}[tbp]
\centering\small
\input{generated/design_table}
\caption{The exact one-sided sign test at $\alpha=\alphalevel$: wins required and power if
each non-tied family is a win with the stated probability (a design calculation computed by
\code{scripts/build_data.py}, not a measurement).}\label{tab:design}
\end{table}

\begin{figure}[tbp]
\centering
\begin{tikzpicture}
\begin{axis}[paper, width=0.75\linewidth,height=5.2cm,xlabel={non-tied qualification families $n$},
  ylabel={power of the sign test},ymin=0,ymax=1.02,xmin=2,xmax=40,legend pos=south east,
  every axis plot/.append style={thick}]
\addplot[okBlue,mark=none] table[col sep=comma,x=n,y=q70] {generated/power_curve.csv};
\addplot[okVermillion,mark=none,dashed] table[col sep=comma,x=n,y=q80] {generated/power_curve.csv};
\addplot[okGreen,mark=none,dotted] table[col sep=comma,x=n,y=q90] {generated/power_curve.csv};
\legend{$P(\text{win})=0.7$,$P(\text{win})=0.8$,$P(\text{win})=0.9$}
\end{axis}
\end{tikzpicture}
\caption{Power of the exact one-sided sign test ($\alpha=\alphalevel$) against the number of
non-tied families. The sawtooth is the discreteness of the test.}\label{fig:power}
\end{figure}
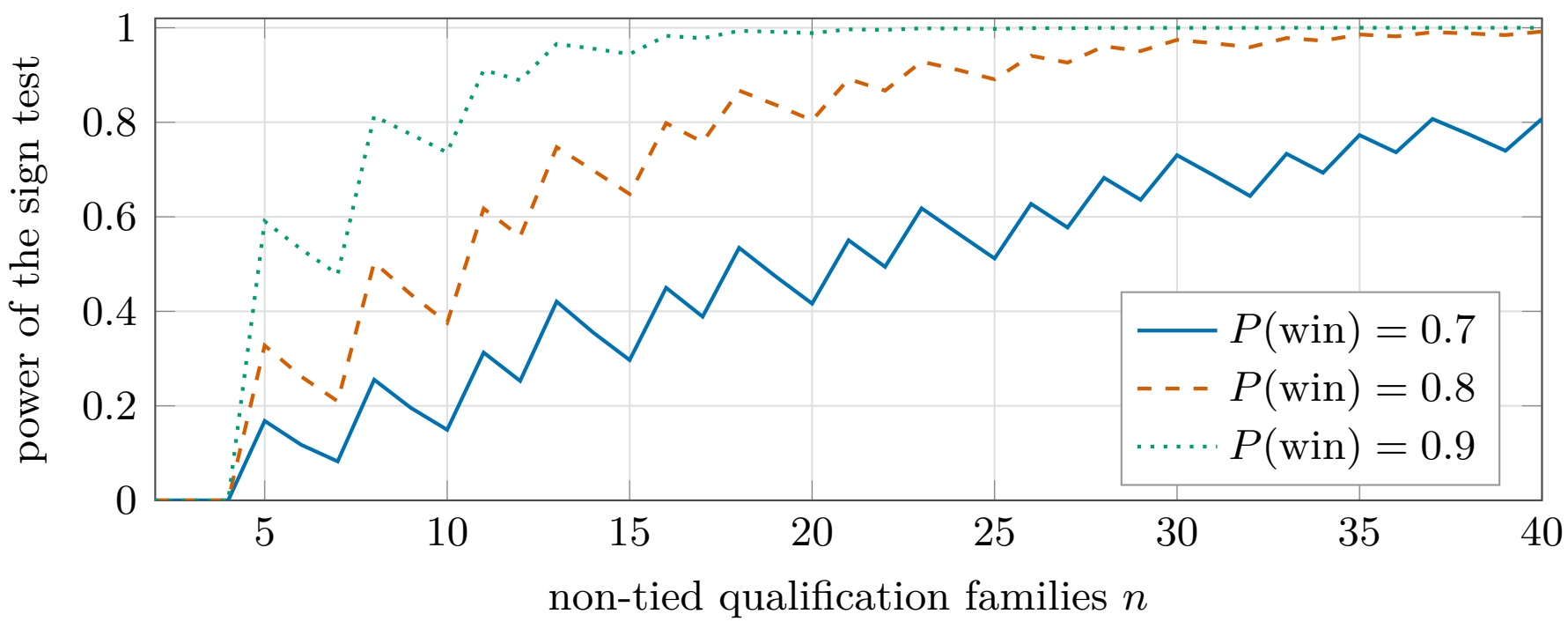

\paragraph{What the test can decide.} \Cref{tab:design} and \cref{fig:power} give the wins
needed and the power at assumed win probabilities: with fewer than five non-tied families no
decision is possible, \mDesignNOneFiveWins{} wins are needed at fifteen non-tied families
and \mDesignNTwoFiveWins{} at twenty-five, where the power at a win share of 0.8 is
\mDesignNTwoFivePowerEightZero{}.\cl{CP01,CP02} The registration therefore asks for about
twenty-five non-tied qualification families, registered as 46 families (28 Python, 18 Lua) before any run, with the base-versus-teacher
discordance on development families reported as the sizing input, and labels a smaller comparison a pilot without changing the
rule.\cl{PR04}

\paragraph{Retention across rounds.} After each round $k$, the exam families of every earlier
round are solved again by the round-$k$ adapter and by its predecessor. Retention holds only if
the one-sided 95\% lower bound of the change in resolve rate (family-cluster bootstrap) exceeds
$-0.05$ absolute; a failed check refuses the candidate.\cl{PR05} The gate of
\cref{sec:gate} already holds each earlier release to the exam of its own run for repository
runs;\cl{G07} the gate's suite does not yet test acquired capabilities in every domain
(\cref{sec:requirements}), which is why the registered check is stricter
than the gate's.

\paragraph{Controls and stop rules.} A gain must survive three controls, each a separate
training run on the same budget: training on trajectories regardless of verdict (does
verification matter), an adapter trained on a different repository (did it learn the tool
protocol rather than the repository), and no replay of earlier rounds; for teacher-distilled data these are teacher trajectories
regardless of verdict and verified teacher trajectories of a different repository.\cl{PR07,PR15} Registered
stop rules include dropping a task kind whose verifier false-pass rate exceeds 5\% and reporting
a development gain without a qualification gain as memorisation.\cl{PR08}

\paragraph{The code that exists.} \code{splinter practice} implements the stages of a
single-repository round from a revision to a gated release.\cl{PS01} Eight authored repository
families ship with the repository (\cref{tab:families}); a family's tests must pass five times
identically and its faults be caught before it is catalogued.\cl{PS02} Two verbs apply the partitions and the paired comparison of this section to the base and the
served adapter.
\code{splinter tasks pick SET... --family-share F --seed S} divides the union of task sets
into a picked set and the rest by a stable hash of the seed and the family, whole families at a
time (\code{--per-family N} caps each family instead, for the one-task-per-family verdict
sample). \code{splinter experiences compare CANDIDATE BASELINE} pairs the two arms' experiences
by task, counts a family once, and reports the exact one-sided sign test the release gate uses,
the per-family wins, losses and ties, quality, cost and autonomy with family-bootstrap
intervals, and the share of tasks on which the arms' verdicts agree; exposed and harness-failed
attempts are left out and counted.\cl{PS09} \Cref{sec:replicate} gives the command for each step of a replication.\cl{PL01}

%% file: generated/design_table.tex
\begin{tabular}{rrrrrr}
\toprule
Non-tied & Wins needed & Smallest $p$ & \multicolumn{3}{c}{Power if $P(\text{win})=$} \\
families $n$ & for $p\le 0.05$ & (all wins) & 0.7 & 0.8 & 0.9 \\
\midrule
4 & none & 0.062 & 0.00 & 0.00 & 0.00 \\
5 & 5 & 0.031 & 0.17 & 0.33 & 0.59 \\
6 & 6 & 0.016 & 0.12 & 0.26 & 0.53 \\
8 & 7 & 0.0039 & 0.26 & 0.50 & 0.81 \\
10 & 9 & 0.00098 & 0.15 & 0.38 & 0.74 \\
15 & 12 & 3.1e-05 & 0.30 & 0.65 & 0.94 \\
20 & 15 & 9.5e-07 & 0.42 & 0.80 & 0.99 \\
25 & 18 & 3e-08 & 0.51 & 0.89 & 1.00 \\
30 & 20 & 9.3e-10 & 0.73 & 0.97 & 1.00 \\
\bottomrule
\end{tabular}

%% file: sections/08-pending.tex

\section{The coding-task experiment: results}\label{sec:coding-results}

\paragraph{The question.} Can verified engineering-agent experience produce a sustained
improvement in completion of previously unseen repository tasks, under a controlled compute budget
and without unacceptable loss of earlier capability?\cl{PR09} The registered confirmatory query
did not show it: on the 46 qualification families the candidate registered before the query
solved 23.0 of the 46 tasks on average over three seeds against 20.0 for its base, winning 18 families, losing 12 and tying
16 (exact one-sided sign test $p=0.18$, not significant at the registered alpha). The same-seed
supervision contrast did separate: supervising only the decisive actions (file edits and writes)
against supervising every action, with corpus, recipe, seed and binary fixed, won 26 families and
lost 11 with 9 tied ($p=0.010$), and the six-decoding-seed development comparison on D46 had
given $p=0.040$ for the same candidate against its base.\cl{QUAL1,PROTO6} The sections below report
what was run, in order, with the pilots that shaped it.

\paragraph{What ran.} The experiments are registered in \code{prereg/coding-tasks.md}, which holds
the protocol of \cref{sec:protocol}. The headline's student is trained on verified trajectories
that the 27B teacher produced on development-partition families and is tested against the untrained
8B;\cl{PR15} the 27B model trained on its own verified trajectories (CE8) is a separate comparison
outside the headline.\cl{PR16} Run, in this order: the teacher's first pass over the eight
development sets (CE1), a pilot of the headline (CE4), two CE8 training runs, CE8-T with its
shuffled control and the G8 agreement check, development probes on D46, and finally the registered
confirmatory query. Registered experiments that were not run are named in \cref{sec:limitations}.
\Cref{tab:results} lists the paired outcomes that have a row in \code{data/results.csv}; every other
number is read from \code{data/measurements.json} through its claim.

\paragraph{Two findings of the first runs.} In a pilot under an earlier,
unbounded harness the untrained 8B model almost never edited a file, which is why the headline's
student trains on verified trajectories of the 27B teacher and why one bounded harness profile is
pinned for every arm. The first teacher run was withdrawn because its shell had no filesystem
confinement; of its 61 episodes, 29 were stopped by the repeated-answer bound and 5 by the tool-call
cap (the 34 conclusions labelled cancelled), and the solve report now names the bound.\cl{TJ01,TJ02}
Every workspace command has since been confined by the kernel and a call naming a path outside the
tree marks the episode exposed (\cref{sec:audit}).\cl{CN01}

\begin{table}[tbp]
\centering\small
\input{generated/results_table}
\caption{Paired outcomes over qualification-partition families (wins: the adapter resolves more
of the family than the base; losses: fewer), with the exact one-sided sign test. Every cell is
read from a finished run; none is an estimate.}\label{tab:results}
\end{table}

\paragraph{CE1: the teacher's first pass, finished.} The 27B teacher has solved the eight development sets once, greedily, confined and under the registered bounds; \cref{tab:ce1} lists every set, each row read from \code{data/measurements.json} and each backed by a ledger entry that quotes its counts.\cl{M39,M40,M41,M42,M43,M44,M45,M46} ``Stopped'' counts the runs that ended at the harness's own bounds (the cap on tool calls or the repetition stop, \cref{sec:protocol}), not infrastructure failures. The exposure audit is applied as registered: an attempt whose trajectory names a path outside its tree is excluded from training evidence and counted, whatever its patch did; the columns ``Hidden pass'' and ``Passed clean'' show how much of the evidence the rule removes in each set.

\begin{table}[tbp]
\centering\small
\input{generated/ce1_table}
\caption{CE1, teacher pass 1 on development sets that have finished. Success and cancelled count solve conclusions, the other columns the verified outcome; exposed attempts are counted among the failed.}\label{tab:ce1}
\end{table}

\paragraph{CE4, the first confirmatory pilot: not shown.} The registered query could be run only on a subset (\cref{app:deviations}), under the protocol in force before the solve protocol of \cref{sec:protocol} (greedy decoding, one attempt per task): one task of each of the 23 families of the first qualification half, solved once by each arm with both served by Brain on the same build and tier. The adapter was trained on the verified, exposure-clean trajectories of the teacher's first pass, projected one tool call per turn. Of the 23 tasks, 21 are paired; the untrained 8B resolves one and the adapter none, so the sign test has a single discordant family against the adapter and does not support the claim; the comparison is a pilot by the registered rule and is reported as \emph{not shown}.\cl{M47} It is a weak test: with the untrained model near zero there is almost no room for a gain, and the adapter's attempts mostly end at the harness's repetition stop. The step count (47) is not the registered rule's (29 for the 114 training records the trainer reports), and the controls, the second qualification half, retention and the serve-agreement sample were not run.

\paragraph{CE8: trained, stopped by its guard, not evaluated.} The registered CE8 training ran
and stopped early at step 51: its training loss fell from 0.1476 at the first
step to 0.0221 at the last, and its best monitoring loss, 0.0839 at step 27 against 0.1101 for the
base, rose to 0.0867 by the last evaluation, so the registered guard calls it memorised.\cl{M48}
The one further run the registration allowed, on a one-pass budget, behaves the same way at a
smaller scale: best monitoring loss 0.0829 at step 30, 0.0832 at its last evaluation, training loss
0.1476 to 0.0282, memorised by the same rule.\cl{M49} Neither adapter is evaluated, and no claim
about CE8 follows from either. The guard reads a single step's training loss on two records and any
rise at the last evaluation, so it fires on noise of this size; the registration keeps it as
written, and the lower monitoring loss than the base's is a training-side quantity, not evidence of
a gain on any task.

\paragraph{CE8-T and its control: trained and evaluated on the pilot subset, no improvement.} CE8-T trained on
one pass over the 113 records the trainer reports (57 steps); its kept evaluation, step 48, has a monitoring
loss of 0.1738 against 0.1799 for the base and a training loss of 0.1331 there, and the training record's
verdict is not memorised, so CE8-T is a candidate for evaluation.\cl{M50} CE8-T-S, trained the same way on the
same 113 records and 57 steps, keeps step 57 at 0.2315 against 0.2635 for the base on its own shuffled
monitoring records, also not memorised.\cl{M51} Both are training-side quantities: the control's monitoring
records are shuffled too, so the two losses are not comparable, and neither is evidence of a gain on any task.

\paragraph{G8: the served adapter agrees with the trained one.} The registration puts G8 before any arm. On the
held-out records of the CE8-T training (33 records in 5 whole families, 354 examples, 68407 supervised
positions) every recorded assistant position is scored twice: by the in-process trainer's forward (bf16 base
with the adapter, no step) and by the evaluation server (Q8\_0 base with the adapter in its bank), and the same
pair is taken for the base. The mean absolute served-versus-in-process difference is 5.005e-02 for the base and
4.694e-02 with the adapter, 0.938 of the base's, so condition (i) holds; top-1 agreement is 97.37\% for the
base and 97.76\% with the adapter, a drop of -0.39 points, so (ii) holds; and the served adapter's mean
log-probability, -0.162778, is above the served base's, -0.181931, so (iii) holds and the verdict is
PASS.\cl{M54} The served adapter is therefore the adapter that was trained, and the evaluation that follows is
of the trained artifact.

\paragraph{The pilot evaluation: no held-out improvement.} Both candidates were then solved on the pilot subset P23 (23 tasks, one per family of the first
qualification half) under the solve protocol of \cref{sec:protocol} with thinking off, one attempt per task,
on one frozen build and one server process that offered the base and both adapters. CE8-T resolves 7 of the
23 tasks and the base 9; of the 13 paired families the adapter wins 3, loses 4 and ties 6, and the one-sided
sign test gives $p=0.773$.\cl{M52} The shuffled control resolves 7 and, over 16 paired families, wins 2,
loses 6 and ties 8 ($p=0.965$).\cl{M53} Neither comparison reaches the registered alpha, and the rule
that labels an underpowered comparison a pilot applies (13 and 16 non-tied families against the 25 the rule
requires): the outcome of both arms is reported as \emph{no held-out improvement}, and no confirmatory query
follows for CE8-T. The adapter does change behaviour on many of the same tasks, but not in one direction, and
the trained adapter is not distinguishable from its shuffled-instruction control in size or direction.

\paragraph{CE8-T-u1b: decision-level supervision, the same outcome.} The pilot's result motivated a third
candidate that changes one variable of the data rather than the recipe: CE8-T-u1b trains on the unpacked
\code{sft-step} view of the same experiences, one record per admitted action with its own context, instead of
one packed record per trajectory. The trainer reports 2490 records, and the run takes 1200 steps at a held-out
monitoring share of 0.05; its kept evaluation, step 960, has a monitoring loss of 0.1676 against 0.1835 for
the base and a training loss of 0.1359 there, and the guard calls it not memorised.\cl{M55} Solved on the same
pilot subset under the same protocol it resolves 6 of the 23 tasks where the base resolves 9; of the 13 paired
families it wins 1, loses 4 and ties 8, and the one-sided sign test gives $p=0.969$, with 6 attempts exposed
against 7 for the base.\cl{M56} Supervising decisions rather than whole trajectories does not rescue the
comparison: the candidate is at or below the base and below its packed predecessor on this subset. Both
candidates are development probes, not registered comparisons.

\paragraph{What these pilots can and cannot show.} Two measurements bracket what the pilot numbers mean.
First, the attempts differ in shape. The median episode is 32 steps for the base, 13 for CE8-T, 12 for its
shuffled-instruction control and 16 for the decision-level candidate; the arms make 782, 474, 392 and 632 tool
calls, and their solve conclusions report 11, 10, 9 and 3 successes, while the longest episodes still run to 58,
60, 61 and 52 steps.\cl{M58} The control matches the trained adapter, so what fine-tuning on these records moves
is when a turn ends, not what a trajectory solves; the decision-level candidate sits between the two, with more
steps than the packed candidates and the most repetition stops. Second, the seed floor: solved twice on the same
23 tasks with two registered protocol seeds, the base itself pairs 13 families with 1 win, 3 losses and 9 ties,
$p=0.9375$, which is statistically indistinguishable from either adapter's comparison.\cl{M57} A pilot of this
size therefore shows no detectable difference, not a demonstrated loss; the registered confirmatory query on 46
families, reported below, is the measurement that could separate them. The training side points the same way: every trajectory
longer than the 16384-token block was removed before training (130 of the 262 successful attempts of the eight
development sets, and 1148 of 3837 for the decision-level view), so the records the adapters were trained on are
the short half of the teacher's successes, and their median record carries about 1650 supervised tokens.\cl{M59}

\paragraph{Development probes on D46: the supervision ablation and recovery.} On 2026-10-25, on two GH200 cards, two series of development probes were run on D46 (46 tasks, 46 families, one task per family, greedy decoding) with one rebuilt training binary (SHA-256 \code{1d366e597bb504bc222304ebbe07ef0165c442f503091340e740004475a120ab}); they are reported as pilots and are \emph{not} the registered confirmatory query, and their outcomes are not used as evidence for it. The first series holds the corpus (769 records, \code{blake3:646692b2}), the recipe (rank 16, alpha 32, $5\times10^{-5}$, 120 steps, packed-bf16 base), the training seed and the binary fixed and changes only the supervision: every admitted action, or \code{--only-actions edit\_file,write\_file} (\code{blake3:92252209}). At seed 13 the decisive-action adapter verifies 30 of 46 tasks with 1 exposed, the all-action adapter 13 with 3 exposed; over 42 paired families the decisive arm wins 16 and loses 1 (exact one-sided sign test $p=0.00014$), and the all-action arm is a regression below its own base (13 against 21, 4W-11L-20T, $p=0.982$).\cl{D46A,D46B} At seed 1 the same contrast keeps its sign but is not significant (23 against 18, both with 4 exposed; 38 paired, 8W-4L-26T, $p=0.194$); the two seeds share the 46 families, so this is one strong same-seed demonstration plus a sign-consistent second seed, not two independent tests.\cl{D46C} The behaviour is seed-stable: the decisive adapters of seeds 1, 7 and 13 edit a file in 35, 36 and 39 of their 46 attempts and end 1, 0 and 2 in the repetition stop, where the all-action adapter edits 22 and aborts 10 and the base edits 24 and aborts 13.\cl{D46D} The same-seed design removes the rebuild confound that earlier comparisons carried: with corpus, recipe, seed and executable fixed, changing only the supervision changes the outcome.

The second series recovers the all-action adapter \code{win1}, damaged to 12/46, on the base's own verified D46 successes (\code{blake3:df000e8f}, 14 records) at $5\times10^{-5}$. Six steps (303 s of training) verify 23 of 46 with 2 exposed against the base's 21 (36 paired, 4W-6L-26T, $p=0.828$); 18 steps verify 20 with 5 exposed (34 paired, 6W-8L-20T, $p=0.788$, measured 2026-10-24); 36 steps (815 s) verify 28 with 7 exposed (31 paired, 10W-5L-16T, $p=0.151$); and the undamaged control, the decisive seed-1 adapter trained for 18 steps on the same corpus, verifies 24 with 10 exposed (30 paired, 7W-2L-21T, $p=0.0898$).\cl{RCVA,RCVB,RCVC,RCVD} The first six steps -- about five minutes -- therefore do most of the restoration, and longer continuation on the benchmark's own families buys a few more tasks while raising the exposed count: the corpus is a labelled rehearsal of the scored families, not a transfer measurement and not free gain. The doses are one seed and share their start, corpus and families, so they are not independent tests, and no comparison reaches the registered alpha.

\paragraph{Six decoding seeds on D46.} The development benchmark was re-solved under the protocol's own sampling at six registered decoding seeds (the three of the earlier pilots and three drawn for this run) by the base, the decisive-action adapter of training seed 13 and its all-action control, on two GPUs. Tasks solved of 46, averaged over the six seeds: base 19.67, decisive 24.5, all-actions 19.83. Per family, decisive wins 22, loses 11 and ties 13 against the base (exact one-sided sign test $p=0.0401$), 22W-11L-13T against the all-action control ($p=0.0401$), and the control 17W-15L-14T against the base ($p=0.430$): the supervision separates from both controls, while the two controls do not separate from each other.\cl{PROTO6} This is a development probe on D46 (46 families, not the qualification set) and no candidate is registered. It is the powered comparison that justifies registering the candidate and running the registered query, not a result of that query.

\paragraph{The registered confirmatory query: not shown.} The candidate registered before the run - the decisive-action adapter of training seed 13 - was solved against its base and its all-action control on the frozen qualification set (46 families) at the protocol's three registered decoding seeds and recommended sampling. Averaged over the three seeds the candidate solved 23.0 of the 46 tasks, its base 20.0 and the control 16.33, with exposed attempts on 10, 35 and 22 of the 138 assigned respectively. Per family the candidate won 18, lost 12 and tied 16 against its base: the exact one-sided sign test reads $p=0.1808$ with a family-bootstrap 95\,\% interval of $[-0.065, +0.188]$ for the mean-fraction difference. At the registered alpha of 0.05 this is \emph{not shown}: the candidate is not demonstrated better than its base on the qualification set, and the paper does not report equivalence. The same-seed supervision contrast does separate: against the all-action control - same corpus, recipe, seed and binary, differing only in the supervision - the candidate wins 26 families and loses 11 ($p=0.0100$, mean fraction +0.145, interval $[+0.007, +0.275]$), while the control against the base is 16W-11L-19T ($p=0.221$). That contrast is a secondary analysis, not the registered decision. The development read on D46 had reached 22W-11L-13T at $p=0.0401$; it did not replicate at the registered level, and the registration's own rule for a development gain with no qualification gain (sign test $p>0.2$, reported as memorisation) is not met either, since $p=0.1808$ - the registered result is a null whose direction matches the development read, next to a causal supervision effect that the control isolates.\cl{QUAL1}

\paragraph{Generation 2: rehearsal protected nothing at this budget.} The decisive seed-1 champion (D46 23/46, P23 16/23) continued on P23's verified successes (25 records) for 120 steps at $5\times10^{-5}$, seed 1; the arms differ only by \code{--rehearsal} of the decisive corpus at 0.25. On D46 the no-rehearsal arm verifies 28 of 46 with 7 exposed (32 paired, 8W-4L-20T, $p=0.194$) and the rehearsal arm 19 with 5 exposed (34 paired, 4W-7L-23T, $p=0.887$); the direct same-start contrast pairs 37 families with 3 rehearsal wins and 13 losses ($p=0.998$ with rehearsal as the candidate, 13W-3L the other way), so rehearsal cost nine tasks relative to the same update without it.\cl{G2A,G2B,G2C} On P23 they verify 12 (5 exposed) and 11 (1 exposed) against the base's 8 (13 paired, 2W-2L, $p=0.6875$; 15 paired, 1W-0L, $p=0.5$), so acquisition over the champion's 16/23 is not demonstrated.\cl{G2D} At this budget the new data did not make the model forget D46 (the no-rehearsal arm added five tasks over its start), so there was nothing for rehearsal to protect and it cost tasks instead. These are one-seed greedy runs on the development benchmark, not the registered query, and the run-to-run variation in the limitations applies to them.\cl{REP1}

\begin{table}[tbp]
\centering\small
\input{generated/pilot_results_table}
\caption{Development probes on D46 (pilot, not the registered confirmatory query): paired families among the 46, wins and losses from the adapter's side, and the exact one-sided sign test computed from them. A row is a probe, not a confirmatory result.}\label{tab:pilots-d46}
\end{table}

\paragraph{How outcomes are read.} The primary decision is the one-sided exact sign test of
\cref{sec:protocol}; a comparison with fewer than 25 non-tied families is labelled a pilot and the
rule is not loosened, and a development gain with no qualification gain (sign test $p>0.2$) is
reported as memorisation.\cl{PR04,PR08}

%% file: generated/results_table.tex
\begin{tabular}{@{}lp{2.9cm}p{4.5cm}rrrrl@{}}
\toprule
Exp. & Repository & Arm & Families & Wins & Losses & $p$ & Status \\
\midrule
CE4 & eight repositories (pilot subset S1) & adapter against untrained 8B & 21 & 0 & 1 & 1.000 & PILOT \\
CE8 & eight repositories (pilot subset S1) & CE8-T against the 27B base & 13 & 3 & 4 & 0.773 & PILOT \\
CE8 & eight repositories (pilot subset S1) & CE8-T-S (shuffled control) against the 27B base & 16 & 2 & 6 & 0.965 & PILOT \\
CE8 & eight repositories (pilot subset S1) & CE8-T-u1b (decision-level) against the 27B base & 13 & 1 & 4 & 0.969 & PILOT \\
\bottomrule
\end{tabular}

%% file: generated/ce1_table.tex
\begin{tabular}{lrrrrrrr}
\toprule
Set & Tasks & Success & Stopped & Hidden pass & Audit removed & Passed clean & Failed \\
\midrule
caldate p1 & 25 & 13 & 12 & 13 & 7 & 6 & 19 \\
exprcalc p1 & 56 & 13 & 43 & 10 & 7 & 3 & 53 \\
graphkit p1 & 51 & 27 & 24 & 24 & 13 & 11 & 40 \\
iniparse p1 & 61 & 21 & 40 & 18 & 7 & 11 & 50 \\
linediff p1 & 50 & 21 & 29 & 18 & 7 & 11 & 39 \\
mdrender p1 & 43 & 19 & 24 & 13 & 4 & 9 & 34 \\
tokenbucket p1 & 38 & 25 & 13 & 22 & 4 & 18 & 20 \\
ttlcache p1 & 46 & 31 & 15 & 34 & 3 & 31 & 14 \\
\bottomrule
\end{tabular}

%% file: generated/pilot_results_table.tex
\begin{tabular}{lp{5.4cm}rrrrl}
\toprule
Probe & Arm & Families & Wins & Losses & $p$ & Status \\
\midrule
D46-SS & decisive actions against all actions (seed 13) & 42 & 16 & 1 & 0.000137 & PILOT \\
D46-SS & decisive actions against all actions (seed 1) & 38 & 8 & 4 & 0.194 & PILOT \\
D46-SS & all actions against the base (seed 13) & 35 & 4 & 11 & 0.982 & PILOT \\
D46-RCV & 6 steps from win1 against the base & 36 & 4 & 6 & 0.828 & PILOT \\
D46-RCV & 18 steps from win1 against the base & 34 & 6 & 8 & 0.788 & PILOT \\
D46-RCV & 36 steps from win1 against the base & 31 & 10 & 5 & 0.151 & PILOT \\
D46-RCV & 18 steps from the undamaged decisive seed-1 adapter (control) & 30 & 7 & 2 & 0.0898 & PILOT \\
D46-GEN2 & no rehearsal against the base & 32 & 8 & 4 & 0.194 & PILOT \\
D46-GEN2 & rehearsal against the base & 34 & 4 & 7 & 0.887 & PILOT \\
D46-GEN2 & rehearsal against no rehearsal & 37 & 3 & 13 & 0.998 & PILOT \\
S1-GEN2 & no rehearsal against the base & 13 & 2 & 2 & 0.688 & PILOT \\
S1-GEN2 & rehearsal against the base & 15 & 1 & 0 & 0.5 & PILOT \\
\bottomrule
\end{tabular}

%% file: sections/08c-generations.tex

\section{Several generations of a 27B model on its own experience}\label{sec:generations}

The repository experiment trains a smaller model on a larger one's trajectories. This section asks
the question of \cref{sec:rsi} on agent workloads at full scale: does
Qwen3.8-27B, trained only on its own verified attempts, become a better agent generation after
generation, on tasks it never trained on? No stronger model takes part, and the base is the only
starting point.

\subsection{Workloads}

Four kinds of task are attempted by the same model in every generation, each with a training
partition the generation practises on and a held-out partition it is examined on; the partitions
are whole families and are fixed before the base attempts anything.
\begin{itemize}
\item \textbf{Repository faults} (\cref{sec:protocol}): one seeded fault per task in the eight
  authored repositories; the exam is the reserved files of six repositories and every file of the
  two sealed ones (mdrender, graphkit), so the exam measures transfer to repositories never trained on.
\item \textbf{Several faults at once}: \code{splinter tasks combine} joins three admitted
  single-fault tasks of one snapshot whose faults lie in different files into one task (the union
  of their checks and reference changes, one merged report); it is admitted only if no member's
  reference alone repairs it, and a combined task with any reserved file is an exam task.
\item \textbf{Embedded firmware}: defects in small firmware services for Cortex-M and nRF5340
  boards, built with the vendor toolchain and graded by running the image in an emulator against
  hidden behavioural checks.
\item \textbf{Mechanical design}: a part drawing to a manufacturable solid, graded by geometry
  against a reference.
\end{itemize}

\subsection{One generation}

Each generation starts from the champion (the base, then the last promoted adapter) and runs four
steps; every step records its result before the next starts.
\begin{enumerate}
\item \textbf{Explore.} The champion attempts every training task $k$ times ($k=4$ for the
  repository workloads) under the model family's recommended sampling; a separate verifier grades
  each attempt and an exposure audit fails any attempt that reached hidden material.
\item \textbf{Learn.} The graded attempts are admitted as experience (held-out families are
  refused by the exam seal) and become on-policy training records, one per model call. Each
  candidate continues from the champion:
  \begin{itemize}
  \item \emph{GRPO}: an attempt's advantage is its reward (1 verified, 0 not) less its task
    group's mean over the group's standard deviation; the clipped surrogate with a $k_3$ KL to the
    base trains the adapter (clip 0.2, $\beta=0.02$).
  \item \emph{GRPO with privileged self-distillation}: the same, and every task whose attempts all
    failed (a group with no reward signal) is distilled instead: the same base model is shown what
    the verifier knows (the failing checks and their output, the reference change where the task
    has one) and each token the student wrote gets the advantage
    $A_t=\mathrm{clip}(\log q(y_t\mid\text{privileged},y_{<t})-\log\pi_{\text{old}}(y_t\mid y_{<t}),\pm 5)$,
    with a cap of 40 on its sum over a sequence, through the same clipped step.
  \end{itemize}
\item \textbf{Examine.} Every candidate and the champion attempt every held-out task under the
  same agent build, serving build, benchmark revision and load per card.
\item \textbf{Decide.} Each candidate is paired task by task with the champion on every workload;
  the candidate with the largest gain over all paired tasks replaces the champion when that gain is
  positive and no workload's mean falls by more than 0.2. The decision, the paired counts and the
  sign test are recorded whether or not the champion changes.
\end{enumerate}

\subsection{What the harness must not decide}\label{sec:generations-harness}

A comparison of agents is only a comparison of models if the software around them fails neither
arm. Two failures of that kind were found while running the loop and are part of the protocol now.
First, a wall-clock limit per trial makes the server's load part of the outcome: with about a
hundred agents on one card, per-stream decoding slowed until correct trajectories were cut off, and
the base's held-out repository pass rate read far below its rate on the same exam at a fifth of the
load. The loop therefore gives a model examined on a share of the cards that share of a workload's trials.
Second, a serving engine that ran out of device memory for one prefill round failed every request
of that round; it now returns the sequences to its queue and narrows admission instead. An exam in
which any trial was ended by the server, the environment or the harness decides nothing.

\subsection{Results}

\paragraph{Generation 0.} The base explored the training tasks of single faults, firmware and
mechanical design, and two candidates were trained on its attempts; the generation was decided on
three workloads with the deviations of \cref{app:deviations}. Against the pooled exams the
self-distilling candidate won more held-out tasks than it lost and replaced the base.\cl{G2701} The GRPO candidate gained on both repository workloads but lost on firmware, so the
rule preferred the self-distilling one even before the firmware loss is weighed.\cl{G2702} Nearly
all of the exam was already solved by the base: the gain is made on the few held-out tasks it
solved only sometimes, and most of what first looked like failures of the base were trials the
serving engine or the per-trial limit ended (\cref{sec:generations-harness}).

\paragraph{Generation 1.} Exploring every training task again teaches little: most single-fault
training tasks were solved by every attempt of generation 0, and a group whose attempts all earn the
same reward has no advantage to learn from. Generation 1 therefore practises a frontier:
the training tasks the base did not always solve and the training tasks with several faults it had
never attempted, with the promoted champion. Its GRPO candidate was promoted on a gain that is
not significant.\cl{G2704}

\paragraph{One measurement of all three.} The generation-0 decision compared the candidates with
a base exam taken on an older serving build, part of which the serving engine had ended. When the
base sat the same exam on generation 1's stack, the generation-0 champion no longer beat
it,\cl{G2703} and after two generations no gain over the base is measured.\cl{G2704} The exams are
near their ceiling for this model: the base already solves about nine held-out tasks in ten,\cl{G2703} so the
few tasks it solves only sometimes carry the whole comparison, and two promotions on small
positive means were not enough to move it. The generations did change the model, in one direction: on the tasks both solved, the
generation-1 champion needed about a tenth fewer completion tokens per solved task than the base on every
workload.\cl{G2705} That was found after the data and over several comparisons, so it is exploratory; it
was registered as a co-primary outcome for later generations of these workloads,\cl{G2706} and no later
run of this paper measured it, because the series that followed moved to a task with headroom for a
small model.
Self-improvement in pass rate across generations is therefore not shown by these runs; what they establish is the loop itself, run end to end without a stronger
model, and the conditions an exam must meet before its decision means anything
(\cref{sec:generations-harness}).

%% file: sections/08b-documentation.tex

\section{A small model learning from raw documentation (exploratory)}\label{sec:documentation}

This section closes the whole loop on a smaller scale: a 4B model is given raw documentation and an
objective, writes its own tasks, has their answers fixed by an interpreter, is trained on every weight
and, once trained, writes the next round's tasks itself; in a third arm a 27B model writes the tasks
instead. These runs were not registered and are exploratory: their design changed between runs, and
their probes were written by the base model and graded under the loop's own protocol. A neutral
re-grading of such probes in S1's arms A and B found most of the gain to be adaptation to that
protocol (\cref{sec:rsi1}),\cl{RS11} so the gains below are gains on the loop's own probes, not
established Lua competence. They motivated the series S1--S4.

\subsection{Setup}

\paragraph{Source and objective.} The source is three chapters of the Lua~5.1 reference
manual (string manipulation and patterns, table manipulation, mathematical functions) as
the HTML the project publishes, in its own Latin-1 encoding; the objective is the
sentence ``Become an exceptional Lua 5.1 programmer''. A local HTML page is read in the
charset it declares and converted to the Markdown the sectioner reads, exactly as a
fetched page is.\cl{SL01} With \code{--structural-graph} the skill graph is read from the
sections, one skill per function or topic, without asking a planner model.\cl{SL02}

\paragraph{Tasks and their answers.} Every task is of the kind \code{predict-lua}: a short
program and the question what it prints. The model writes the program and the question;
admission sets the program below the question in a fenced block, so it is shown once and
exactly as it is run,\cl{SL03} and the answer is what the pinned Lua~5.1.5 interpreter
prints: the program must run to completion, print a bounded, non-empty output and print
the same again, and that output becomes the reference whatever the model predicted.\cl{SL04}
An answer is graded by the formal verifier, which takes an output given in a fenced
block as the answer and treats two different blocks as a hedge.\cl{SL05}

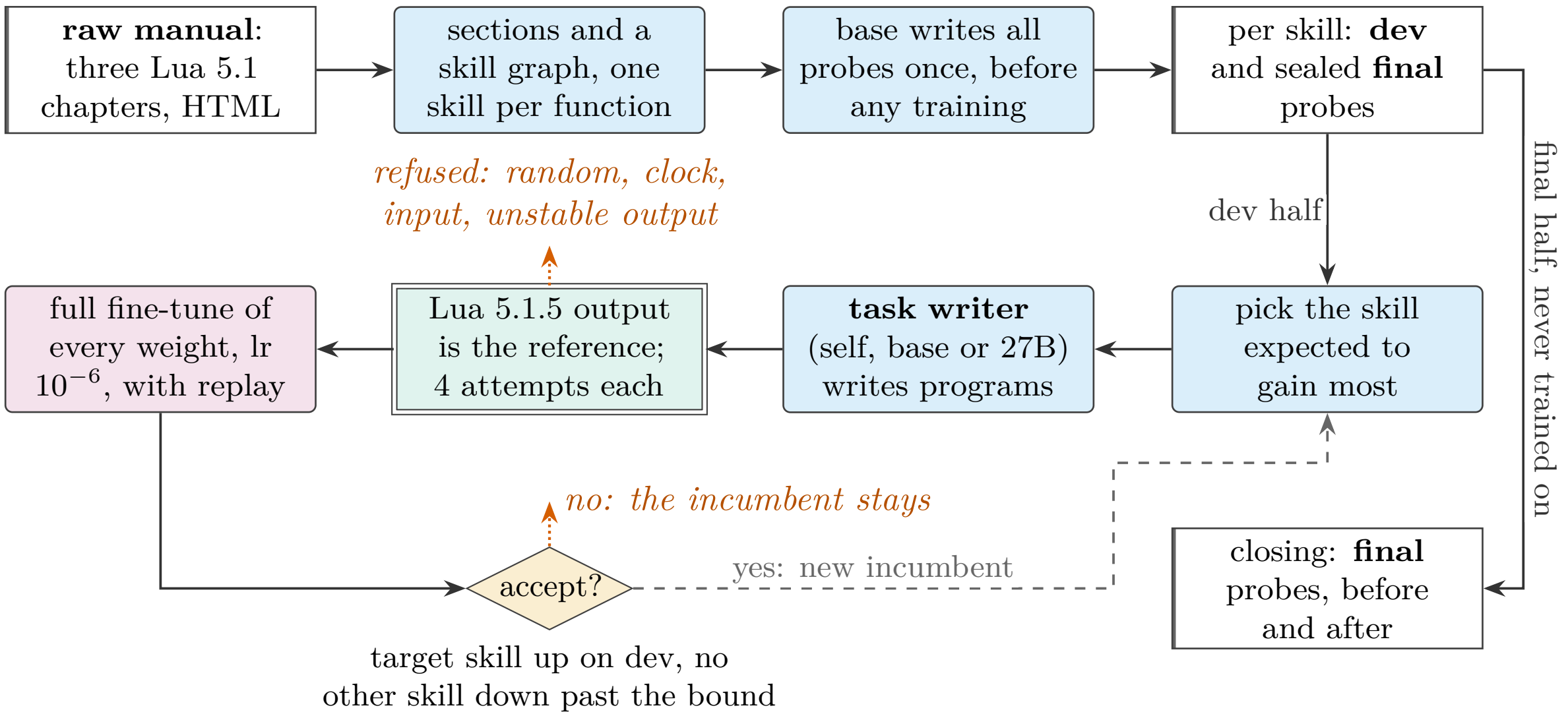
\begin{figure}[tbp]
\centering
\begin{tikzpicture}[fig]
\tikzset{cell/.style={text width=29mm}}
\def\dx{39mm}
\node[data, cell] (html) at (0,0) {\textbf{raw manual}:\\three Lua 5.1\\chapters, HTML};
\node[proc=okSky!22, cell] (graph) at (\dx,0) {sections and a\\skill graph, one\\skill per function};
\node[proc=okSky!22, cell] (probes) at (2*\dx,0) {base writes all\\probes once, before\\any training};
\node[data, cell] (split) at (3*\dx,0) {per skill: \textbf{dev}\\and sealed \textbf{final}\\probes};
\node[proc=okSky!22, cell] (pick) at (3*\dx,-28mm) {pick the skill\\expected to\\gain most};
\node[proc=okSky!22, cell] (write) at (2*\dx,-28mm) {\textbf{task writer}\\(self, base or 27B)\\writes programs};
\node[verif, cell] (lua) at (\dx,-28mm) {Lua 5.1.5 output\\is the reference;\\4 attempts each};
\node[proc=okPurple!22, cell] (ft) at (0,-28mm) {full fine-tune of\\every weight, lr\\$10^{-6}$, with replay};
\node[decide, aspect=2, inner sep=0.5pt, text width=10mm] (acc) at (\dx,-52mm) {accept?};
\node[lab, anchor=north] at ($(acc.south)+(0,-1mm)$) {target skill up on dev, no\\other skill down past the bound};
\node[data, cell] (fin) at (3*\dx,-52mm) {closing: \textbf{final}\\probes, before\\and after};
\draw[flow] (html) -- (graph);
\draw[flow] (graph) -- (probes);
\draw[flow] (probes) -- (split);
\draw[flow] (split) -- node[lab, left] {dev half} (pick);
\draw[flow] (pick) -- (write);
\draw[flow] (write) -- (lua);
\draw[flow] (lua) -- (ft);
\draw[flow] (ft) |- (acc.west);
\draw[ctrl] (acc.east) -- node[lab, above] {yes: new incumbent} (2.45*\dx,-52mm) |- ($(pick.south)+(0,-5mm)$) -- (pick.south);
\draw[refuse] (acc.north) -- ++(0,4.5mm) node[reflabel, right] {no: the incumbent stays};
\draw[refuse] (lua.north) -- ++(0,4mm) node[reflabel, above, align=center] {refused: random, clock,\\input, unstable output};
\draw[flow] (split.east) -- ++(4mm,0) coordinate (a) -- node[lab, sloped, above] {final half, never trained on} (a |- fin.east) -- (fin.east);
\end{tikzpicture}
\caption{The documentation loop of \cref{sec:documentation} (exploratory), read from the raw manual.
Probes are written once by the untrained model and split per skill: the dev half decides every round,
the final half is touched only by the closing assessment. Each round a task writer (the incumbent
itself, the untrained base, or a 27B model in the third arm) writes programs, the interpreter fixes
their answers, and every weight of the 4B incumbent is trained on the attempts it solved sometimes;
dotted arrows are what is refused.}\label{fig:doc-loop}
\end{figure}

\paragraph{The loop (\cref{fig:doc-loop}).} Probes are written once, before any training, by the model as it
is, and split per skill into dev probes (the baseline and every round's decision) and
final probes (the closing assessment only); a skill is asked for probes again while it
has fewer than both halves need.\cl{SL06} Each round picks the skill expected to benefit
most, has the incumbent write training tasks for it and attempt each four times, keeps
the tasks it solved sometimes and the verified answers to them, trains every weight of
the incumbent on them with a replay of the earlier accepted rounds' records, and accepts
the candidate only if the target skill improved on its dev probes and no other skill
lost more than the retention bound.\cl{SL07} With \code{--self-generate} the incumbent writes
every round's tasks: the base in round one, the last accepted checkpoint after.\cl{SL08}
The control arm is the same run without it: the base writes every round's tasks.

The run records this section cites (every round's stage and outcome per arm and replicate, and the
arms comparison) are in the evidence dataset \code{the-frontier-must-grow-evidence}
(\cref{sec:repro}).

\subsection{What full fine-tuning costs on one card}

Brain's full fine-tune holds the weights and gradients of a Qwen3 model in fp32 on the
card and the AdamW moments and master weights in host memory; \code{samples/text/chat-full-finetune}
runs it on any base in the model store.\cl{SL09} \Cref{tab:fullft} gives what eight steps
cost on one GH200 at a 631-token block.\cl{SL10}

\begin{table}[tbp]
  \centering
  \small
  \begin{tabular}{lrrrr}
    \toprule
    Model (optimizer state) & Card (GiB) & Host (GiB) & s / 8 steps & Loss \\
    \midrule
    Qwen3-0.6B (fp32) & 7.8 & 10.9 & 62 & 1.38 $\to$ 0.91 \\
    Qwen3-1.7B (fp32) & 15.9 & 32.6 & 128 & 1.56 $\to$ 0.81 \\
    Qwen3-4B (fp32) & 33.6 & 75.2 & 251 & 1.76 $\to$ 0.91 \\
    Qwen3-8B (fp32) & 65.6 & 152.7 & 488 & 1.12 $\to$ 0.47 \\
    Qwen3-8B (int8 moments) & 65.7 & 107.5 & 459 & 1.30 $\to$ 0.83 \\
    \bottomrule
  \end{tabular}
  \caption{Full fine-tuning on one GH200: the run's peak on the card beyond what it held
  before, the process's peak resident host memory, wall time and training loss before the
  first and after the last of eight steps (80 chat records written from the Lua manual).
  The 8B fits the card; in fp32 its host state is 153~GiB, beyond what a 205~GiB job can
  spare beside a served 27B model, and int8 moments save 45~GiB of it.}
  \label{tab:fullft}
\end{table}

\subsection{What a small model needs from the loop}\label{sec:small-model-needs}

Each of the following stopped the loop on the Lua chapters before a single round could
train; each was fixed in the code the paper describes, and each is a requirement any loop
driven by a model of this size has to meet.

\paragraph{The task must show what is run.} Asked to put its program in the material and
again verbatim in the instruction, the 4B model wrapped it in other code
(\code{material = [[print(\dots)]]}), a program that prints nothing, and admission, which
checked only that the material occurs in the instruction, accepted it. The model now
writes the question alone and admission sets the material below it.\cl{SL03}

\paragraph{The answer must come from the runtime.} With the model's own prediction as
the reference, a task was refused whenever the model mispredicted its program, and most
skills were left with too few probes. The reference is now what the interpreter
prints.\cl{SL04}

\paragraph{The verifier must read the answer the model gives.} The model answers a
predict task with a sentence above a fenced block. The verifier took candidates only
from emphasis and numbers, and most failing baseline probes held the exact output in a
fence: the baseline measured the answer's format, and training would have taught it.\cl{SL05}

\paragraph{One request is not enough.} A 4B generator admits two or three tasks per
request; probes and training tasks are asked for again, shown what was already written,
while a skill is short.\cl{SL06,SL11}

\paragraph{Generation must be compared by what it shows.} Every predict task asks the
same question of a different program; compared by its whole instruction, every pair was a
near duplicate and nearly every training task was refused as repeating a probe.
A task that shows material is now compared by the material.\cl{SL12}

\paragraph{Writing tasks is not executing a procedure.} The generator sampled at the
agent's temperature (0.2), which keeps a small model on task while it acts but has it
write the same programs whenever it is asked again; it now samples at 1.2 with a top-k of
20.\cl{SL13}

\paragraph{The environment can teach.} A task the policy never solves was taught by the
policy shown the task's passage, and a model cannot run a program; with
\code{--reference-teacher} the runtime's output is said as the teacher's reply and graded
like any answer.\cl{SL14}

\paragraph{The exam is not material.} A round's request showed the generator its skill's
probes as tasks not to repeat; the 4B model copied them, and most training tasks of a
round were exact probes. The generator is now never shown the exam.\cl{SL15}

\paragraph{An answer must be one the language defines.} A seeded \code{math.random}
program prints the same on every run and passed every check, but what it prints is the C
library generator's draw. A candidate fully fine-tuned on such answers lifted its target
skill from 0 to 4 of 4 dev probes while the other skills fell from 35 to 20 passed probes
(\code{string.byte(s, 1)} answered 97 for every string); the retention bound rejected
it.\cl{SL18}
Programs that read a generator, the clock, the environment, input or the collector are
now refused.\cl{SL16}

\paragraph{A probe must be answered greedily.} Probes were answered on the model's own
sampling (0.2, a seed per request), so the baseline and every round's decision were draws.
They are now answered greedily, and the same base on the same exam scores the same in both
arms, 34 of 56 dev probes.\cl{SL17,SL19}

\paragraph{The loop learns at a small step.} At a learning rate of $10^{-6}$ (six steps,
one verified answer per task, retention bound 0.5) rounds are accepted: each arm accepted 5
rounds, and on the final probes, which no round trains on, the self-generate arm went from
26 to 35 of 53 passes (7 skills up, 1 down) and the fixed-generator arm from 26 to 40 of 53
(8 up, 0 down).\cl{SL20} The control did better than the arm whose trained model writes the
next tasks, so these runs show gains on the loop's own final probes, not that self-generated tasks
help.

\paragraph{A second replicate.} The same experiment on a fresh exam, written by the base model,
with the full fine-tune of a later brain revision, accepted 3 rounds in each arm. On the final probes the self-generate arm went
from 20 to 30 of 46 in 13 rounds (7 skills up, 0 down) and the fixed-generator arm from 20 to 29 of
46 in 8 rounds (7 up, 2 down).\cl{SL21} Both arms learn in both replicates; which generator does
better changes sign between them, so on this evidence the loop gains on its probes from tasks its own
model writes, and whether the trained model's tasks are better than its base's is not settled. A third
replicate on the second exam, which changes only the runs' random draws, agrees: the self-generate arm
went from 20 to 33 of 46 (2 of 10 rounds accepted, 8 skills up, 0 down) and the fixed-generator arm
from 20 to 32 (3 of 9 accepted, 7 up, 1 down).\cl{SL22} All six arms of the three replicates ended
above their baseline on probes no round trained on; the self-generate arm finished ahead in two
replicates and behind in one.

\paragraph{Who writes the tasks: no difference detected.} If tasks written by a more capable model
taught more, a 4B solver fed tasks from a 27B writer would gain more than one fed tasks from itself.
We ran a third arm in which the 27B, served by brain on the second card, writes every round's tasks
while the 4B still solves, is verified by the interpreter and is fully fine-tuned, and repeated all
three arms on the second exam until each had at least six replicates; runs whose training was refused
for want of device memory are excluded, not counted as zero. The 7 self-generate, 9 fixed-generator
and 6 teacher-generator replicates gained on average 10.86, 10.89 and 11.17 passes of 46 on the final
probes. Exact permutation tests over replicates find no difference (p = 1.0 between self-generate and
fixed-generator, p = 0.89 between self-generate and teacher-generator, p = 0.87 between
fixed-generator and teacher-generator), and per skill the teacher-generator and fixed-generator arms
are each better on 5 skills with 3 tied.\cl{SL23} With 6 to 9 replicates per arm these are weak
tests: the differences of mean gains are $-0.03$ (95\% interval $-1.99$ to 1.92) between self-generate
and fixed-generator, $-0.31$ ($-2.69$ to 2.07) between self-generate and teacher-generator and $-0.28$
($-2.22$ to 1.66) between fixed-generator and teacher-generator, and only differences of about 2.7 to
3.3 passes of 46 would have been detected with power 0.8.\cl{RS44} No effect of who writes the tasks was detected; an
effect smaller than that is not excluded. The tasks here are short programs whose output the
interpreter fixes, so a stronger writer has little room to write a better task.

%% file: sections/09-replication.tex

\section{Replicating the result on another repository}\label{sec:replicate}

A result that only holds for the repositories it was developed on is a demonstration, not a
method. This section is the recipe for running the loop on a repository of one's own, step by
step, each with the command in the pinned revision that does it and whose command line the ledger
checks; there is no separate benchmark command.\cl{PL01} It claims no outcome; the replication of the headline on a
second repository was registered (CE7) and was not run (\cref{sec:limitations}).

\subsection{Requirements on the repository}

\code{splinter practice} learns a codebase from its own tests: a git work tree or bare
repository, a revision git resolves to a commit, and a test suite that runs offline and is
green and stable on a clean copy of that revision. Splinter recognises a Python project with
a \code{tests} directory (run as \code{python3 -m unittest}) and a Cargo project (run with
\code{cargo test --offline}); any other suite is named with \code{--test-command}, and the files
tasks may change with repeated \code{--file}.\cl{PS01,PS04} The suite is run twice on a clean
copy first, and a red or flaky baseline stops the run before any model is loaded.\cl{PS01}
Isolation is recorded as \code{none} unless a container client and a digest-pinned image are
configured, although a host command is confined by the kernel to its tree without a network;
a repository whose tests need the network or a process-level boundary for soundness is not
suitable on a host without a container.\cl{AU09,CN08}

\subsection{Task-family construction}

The protocol's unit is the family (\cref{sec:protocol}). \Cref{tab:families} shows what the
construction yields on the eight authored repositories that ship with Splinter: of several
hundred seeded faults per repository, those that do not build or that no test notices are
dropped, near-duplicates and truncated outputs are refused, and the rest are admitted.\cl{PS02}
For a new repository the steps are, in order:

\begin{enumerate}
\item \textbf{Check the environment}: \code{splinter doctor} confirms the state
root is writable with room to spare and the policy's weights fit the card.\cl{PS05}
\item \textbf{Generate and count the admitted tasks without a model}: run
\code{splinter practice REPO --rev REV --budget 1s --no-release --max-tasks N
--mutants-per-file M --test-command ...}; the \code{[tasks]} line states the admitted count
and the refusals by reason. This is the funnel of \cref{tab:families}, and it needs no model and
no GPU.\cl{PS03,PS06}
\item \textbf{Reserve the exam}: \code{--exam-families} sets families aside by a
stable hash of their names before any attempt, and \code{--exam-seed} is drawn, recorded and
refused if changed on a later run.\cl{PS06}
\item \textbf{Declare partitions and generations}: \code{splinter experiment check} validates a plan whose sealed
families are never trained on; a three-way split with a qualification partition queried once
is the registration's requirement, and \code{splinter tasks pick}
draws the development holdout and the qualification partition from the task sets by seed and
family.\cl{PS07,PL01}
\item \textbf{Measure the untrained agent} (the teacher's pass is reported in \cref{sec:coding-results}): the base attempts the development and
qualification partitions with attempts per task fixed from a throughput probe, and
\code{splinter experiences compare CANDIDATE BASELINE} pairs any two such experience sets
by task and reports the family-clustered sign test and per-family counts.\cl{PL01}
\item \textbf{Train and gate} (one round per run): \code{splinter practice REPO} solves the
non-exam tasks as an agent, keeps the attempts the hidden checks confirm, builds the
action-level SFT dataset, trains a LoRA candidate from the champion with earlier releases
replayed, and runs the release gate over the exam the run reserved; \code{--bf16-base},
\code{--steps}, \code{--rank} and \code{--max-block} bound the training to the card.\cl{PS01,PS06}
\item \textbf{Serve and roll back}: a release moves the alias and rewrites the
serving manifest, \code{brain serve --adapter-manifest} follows it, and \code{splinter rollback}
repoints it.\cl{X05,X06,PS08}
\end{enumerate}

\begin{table}[tbp]
\centering\small
\input{generated/families_table}
\caption{The eight authored repository families: language, tests, and tasks admitted by the
fault-seeding funnel (values read from the family catalogue at the pinned revision and
checked by the ledger).}\label{tab:families}
\end{table}

\subsection{What it costs}

Task generation costs CPU time only: no model is loaded and no GPU is used.\cl{PS03} Training
cost on a 27B model is the token count over the measured rate; at the \mThrNfFourEightOneNineTwoTokensPerS{}~tokens/s of the
probe of \cref{sec:throughput} one million trained tokens take \mThrHoursPerMillionTokens{}~h
of card time.\cl{M38,C12} The remaining terms of the cost of one verified outcome, wall time and
tokens per attempt and attempts per resolved task, depend on the repository and the model and are
not reported here; they are measured by solving a few families with the untrained model before
card time is committed to training.

\subsection{What to report}

For a replication to count it must report, for the new repository, the same quantities as
\cref{sec:coding-results}: the admission funnel by reason, the baseline resolve rate with its
interval, the paired sign-test counts on the qualification partition, the retention check,
the isolation level and the three repository commits. Counts that disagree with ours are as
informative as ones that agree.

%% file: generated/families_table.tex
\begin{tabular}{llrr}
\toprule
Family & Language & Tests & Admitted tasks \\
\midrule
\code{iniparse} & python & 279 & 177 \\
\code{ttlcache} & python & 257 & 159 \\
\code{exprcalc} & python & 256 & 186 \\
\code{linediff} & python & 278 & 152 \\
\code{mdrender} & python & 270 & 148 \\
\code{caldate} & lua & 250 & 132 \\
\code{graphkit} & lua & 262 & 155 \\
\code{tokenbucket} & lua & 296 & 135 \\
\bottomrule
\end{tabular}

%% file: sections/app-related-system.tex

\section{Related work for the system}\label{app:related-system}

\emph{Parameter-efficient training.} LoRA freezes the base and trains low-rank
adapters;~\citep{lora2021}\cl{CT03} QLoRA does so over a base quantised to 4~bits;~\citep{qlora2023}\cl{CT04} DPO optimises
preferences without a separate reward model.~\citep{dpo2023}\cl{CT05} The flywheel uses LoRA and DPO as
objectives it selects, and delegates their computation to Brain (\cref{sec:training}).
\emph{Retrieval-aware fine-tuning.} RAFT trains a model on domain documents with
distractor passages;~\citep{raft2024}\cl{CT06} Splinter has an optional passage-mixing share of the same
family (\cref{sec:dataset}), which we do not evaluate here.
\emph{Forgetting.} Catastrophic forgetting is the loss of earlier capability when a
network is trained on new data;~\citep{ewc2016}\cl{CT07} Splinter's response is empirical: it measures
retention against every earlier release at release time (\cref{sec:gate}).
\emph{Memory and compute for training.} Recomputing activations trades compute for memory
in the backward pass;~\citep{recompute2016}\cl{CT08} fused or chunked cross-entropy avoids materialising the
logits of a large vocabulary;~\citep{cutxent2024}\cl{CT09} FlashAttention computes exact attention without
materialising the score matrix;~\citep{flashattention2022}\cl{CT10} Gated DeltaNet is a gated linear-attention layer based on the delta rule
and is the recurrent layer type of the 27B model Brain implements (three Gated DeltaNet
layers to one attention layer).~\citep{gateddelta2024}\cl{CT11,B11} Section~\ref{sec:substrate}
reports which of these Brain contains.
\emph{Repository tasks and agents.} SWE-bench poses resolving real GitHub issues as the task,
graded by the repository's tests;~\citep{swebench2023}\cl{CT12} SWE-agent shows that the
interface between the model and the repository changes what a model achieves, so the interaction,
not only the final patch, is what is recorded and trained on.~\citep{sweagent2024}\cl{CT13}
SWE-Gym, R2E-Gym and SWE-smith build executable \emph{training} environments, by collecting
instances, by procedural generation from commits and by modifying code so that tests
fail.~\citep{swegym2024,r2egym2025,swesmith2025}\cl{CT14} We take the mutation-and-validate
idea and the commit-based idea as task producers (\cref{sec:protocol}). The difference
in aim is that the target here is one repository's own tasks, reserved before rollouts and
evaluated paired against the base under an identical harness, with retention on earlier
repositories and a replication on a second one registered (and not run, \cref{sec:limitations}); we make no claim about priority.

%% file: sections/app-limits.tex

\section{Limitations of the system and of the other experiments}\label{app:limits}

\paragraph{What the structure claims show.} They show that definitions, constants, tests and messages
exist at the pinned commits; they do not show that the code behaves accordingly. Behaviour is
established only by the measurements of
\cref{sec:substrate,sec:coding-results,sec:documentation,sec:generations,sec:rsi}. The qualification
of the 27B training substrate used random weights for the long sequences and a handful of examples for
the real checkpoint, and no sequence longer than 16384 tokens was run. The requirements of
\cref{sec:requirements} include one that is not implemented: the retention check does not test what
each release demonstrably acquired, so for a domain of facts a passed retention check is not evidence
that taught facts were kept. The workspace verifier's reading of the runner's report and the practice
gate's canaries exist at the current head, not at the revision the architecture claims are pinned
to.\cl{OP04,OP05} The corpus expansion mechanisms (variants, retention queue, rehearsal, claims) are
described from the code and none was evaluated for the value it adds.

\paragraph{Single greedy runs and run-to-run variation.} The development probes of \cref{sec:coding-results} (the supervision ablation, the recovery dose-response and the generation-2 rehearsal ablation) are one run per arm at greedy decoding. The same adapter evaluated three times on the same benchmark decoded 30, 31 and 26 of 46 tasks, and the three reports' verdicts differ pairwise on 9, 12 and 11 of the 46 tasks; a single greedy arm therefore carries about 2.5 tasks of standard deviation, so the paired p-values of single-run comparisons overstate their precision.\cl{REP1} The variation is two different things, and the decode batch is only one of them.\cl{REP2} A single-row decode is reproducible: five sequential identical task-derived requests, sent greedily with the requested seed echoed by the server, returned identical completion token ids and bitwise identical per-token log-probabilities, and a never-seen prompt sent once cold and then twice warm was identical both times. A row's arithmetic does depend on its co-batch: six or eight identical requests sent at once return two distinct sequences, because the int8 projections run a block that is bit-identical below five rows and switch to an int8 tensor-core block from five, while the fused decode attention sizes its split-key count from the batch. But one request in flight does not make a whole episode reproducible either: the same adapter on the same eight tasks, greedy at concurrency one, passed three and then five of eight with two verdicts flipped. The registered multi-seed protocol is therefore the instrument for agent episodes. Training is not bit-reproducible either: two runs of one recipe with the same binary, corpus, start and seed produced two distinct adapters whose losses agree to the fourth digit while their gradient norms and weights do not.\cl{REP3} A same-seed comparison therefore means the same data, recipe and binary, not a bit-identical run. The series' examiner protocol is greedy at one request per item on one build and tier, and its base-against-base null replicate found no discordant item (\cref{sec:rsi-design}).

\paragraph{The coding-task experiment.} The registered confirmatory query does not show the claim
(\cref{sec:coding-results}), and the same-seed supervision contrast that does separate is a secondary
analysis, not the registered decision. Registered experiments that were not run: the ten-stage
training and deployment experiment for the 27B model (CE6), the replication of the unchanged procedure
on a second independent repository (CE7), and the acquisition, choice-of-improvement and generations
experiments of \code{prereg/acquisition.md} (ACQ1, INT1, GEN1). The paper therefore reports no
replication of the headline, no result of the registered retention check or of the three controls of
\cref{sec:protocol}, and no result of ACQ1 (the documentation runs of \cref{sec:documentation} were not
run under that registration). The memorisation guard of CE8 as registered reads one step's training
loss and any rise at the last evaluation, so it can stop a run on noise; it was kept unchanged.\cl{M48,M49}
Both pilots of the adapter are inside the same variation the base shows against itself on a second
protocol seed.\cl{M57} The registration fixes one model size for the first training result (Qwen3-8B),
trained on a stronger teacher's verified trajectories because the untrained 8B produced too few edits
to supply its own, so a gain there would be distillation from the teacher, not
self-improvement;\cl{PR15,PR16} one agent harness; and process isolation \code{none} on the node (a
solver's commands are confined by the kernel to their tree and have no network, but the node is not a
container; \cref{sec:audit}).\cl{AU09,CN01,CN08} Mutation-seeded tasks test whether a model can undo a
synthetic fault, not whether it can debug a historical one. A family is the file a fault was seeded in
for the authored repositories, so clusters are files, not subsystems.\cl{PR02} The base model may have
seen public repositories in pretraining; the paired base arm bounds that effect, it does not remove it.

\paragraph{The 27B generations.} Two generations were run, not the three planned, on three of four
workloads, on exams the base already solved about nine times in ten, so the few tasks it solved only
sometimes carried every comparison.\cl{G2703} The token-efficiency finding was made after the data
over several comparisons and is exploratory.\cl{G2705}

\paragraph{Verification and grounding.} Lexical grounding of a task in its source (\cref{sec:admission})
establishes provenance, not entailment: a task whose reference shares the source's words can still be
wrong about it. The verifier kinds check different properties, and no ranking among them makes a
stronger verdict override a failure of another kind. Aggregate gates (\cref{sec:gate}) can hide the loss
of a rare skill inside an unchanged total. The engineering defects found and fixed along the way
(\cref{sec:requirements,sec:small-model-needs}) are part of the result: each would have produced a
plausible but wrong measurement.

%% file: sections/12-deviations.tex

\section{Deviations from the registration}\label{app:deviations}

The protocol of \cref{sec:protocol} is the registration in \code{prereg/coding-tasks.md}, fixed before
any coding-task run. \Cref{tab:deviations,tab:deviations2,tab:deviations3} list the dated changes made in this paper after the
registration; ``Before'' names the earliest outcome a change preceded. The self-improvement series
(\cref{sec:rsi}) record their own deviations, dated, in their launch records and outcome entries, and
\cref{tab:deviations-rsi} lists them.

\begin{table}[tbp]
\centering\small
\setlength{\tabcolsep}{3pt}
\begin{tabular}{@{}p{0.12\linewidth}p{0.58\linewidth}p{0.25\linewidth}@{}}
\toprule
Date & Change & Before \\
\midrule
10-21 & CE8-T's memorisation guard judges the kept checkpoint: memorised if its monitoring loss is not below the base's, or its windowed training loss is below half its monitoring loss. Defined after the original guard stopped both CE8 runs on a last-evaluation rise of 0.0003 to 0.003; Under this rule both CE8 runs would pass (kept checkpoints at steps 27 and 30, monitoring loss below the base's, training loss above half of it); CE8 stays not evaluated, because the guard in force for it fired. From here on the protocol changes in this paper, not in the registration file. & CE8-T's training \\
10-21 & CE8 being closed, CE8-T gets its own shuffled negative control (CE8-T-S, from the CE8-T experiences) and the one-pass budget the second CE8 run used; its step count follows the trainer's own record count, read from a one-step sizing run. & CE8-T's training \\
10-22 & G8 for the 27B compares served and in-process teacher-forced log-probabilities on held-out records, against the base's own served-vs-in-process gap, instead of agent verdicts: the 27B has no in-process agent path (splinter's local provider runs qwen3 only), found when the served half of the verdict check had run (9 of 23 verified) and before any comparison of the two halves existed. & G8 and any evaluation arm \\
10-22 & CE8-T and CE8-T-S are solved on the pilot subset P23 under amendment 20 (thinking off) before G8 is measured, which the registration places before any arm; both comparisons have fewer than 25 non-tied families, are labelled pilots by the rule, and no confirmatory query follows from them. & any CE8-T evaluation \\
\bottomrule
\end{tabular}
\caption{Changes to the protocol made in this paper, in the order they were made.}\label{tab:deviations}
\end{table}

\begin{table}[tbp]
\centering\small
\setlength{\tabcolsep}{3pt}
\begin{tabular}{@{}p{0.12\linewidth}p{0.58\linewidth}p{0.25\linewidth}@{}}
\toprule
Date & Change & Before \\
\midrule
10-25 & The release gate's improvement check also requires the candidate to have solved at least as many of the suite's assigned tasks as its champion; an attempt withheld as exposed or as a harness failure counts as a task that arm did not solve, so a regression can no longer leave the paired test's denominator. The measured shape that motivated it: a base at 21/46 (9 exposed) against a candidate at 15/46 (19 exposed) read as 5W-3L-15T under the paired test alone. & any release after it \\
10-25 & Development probes on D46 (46 tasks, one per family, greedy) measure the supervision ablation and the recovery dose-response. They are labelled pilots, are not the registered confirmatory query, and their outcomes are not used as evidence for it; the registration, its partitions and its denominators are unchanged.\cl{D46A,D46C,RCVA,RCVC} & the D46 probes \\
10-25 & The evaluation-variance explanation is corrected. The two serving probes that appeared to reproduce run-to-run nondeterminism hashed the whole response payload, which carries a random tool-call id and a random chat id, so their distinct-completion counts were artifacts. Remeasured on the completion's token ids and log-probabilities, a batch-1 greedy decode is bit-reproducible, a cache hit equals a cold prefill, and the co-batch does change a row (the int8 projections switch to a tensor-core block from five rows and the decode attention sizes its split count from the batch) - but one request in flight does not make a whole episode reproducible, so a second, unidentified source remains. The three-run greedy spread this paragraph explains (30, 31 and 26 of 46) is unchanged.\cl{REP1,REP2} & the REP1 paragraph and any reading of a single greedy arm \\
10-25 & The confirmatory candidate is registered before its query: the decisive-action adapter of training seed 13 (corpus \code{blake3:92252209}, rank 16, alpha 32, $5\times10^{-5}$, 120 steps, packed-bf16 base, the rebuilt training binary), chosen after the six-decoding-seed development read on D46 in which it won 22 families, lost 11 and tied 13 against the base and the same against its all-action control (p=0.0401 each) while the two controls did not separate. The query solves the frozen qualification set (\code{blake3:f7fe502f}, 46 families) with the candidate, its base and the all-action control at the three registered decoding seeds and the protocol's sampling; the decision is the registered one-sided exact sign test at $\alpha=0.05$, and a result that is not significant is reported as not shown.\cl{PROTO6} & any candidate query \\
\bottomrule
\end{tabular}
\caption{Changes made in this paper, continued.}\label{tab:deviations2}
\end{table}

\begin{table}[tbp]
\centering\small
\setlength{\tabcolsep}{3pt}
\begin{tabular}{@{}p{0.12\linewidth}p{0.58\linewidth}p{0.25\linewidth}@{}}
\toprule
Date & Change & Before \\
\midrule
10-26 & The registration's amendment history (the dated amendments appended to \code{prereg/coding-tasks.md} before the runs they could affect) is kept in the repository's history of that file and is no longer restated in this paper; the protocol of \cref{sec:protocol} is the registration as it stands, and the changes made in this paper are the rows above. & the paper's final form \\
11-06 & Generation 0 of the 27B loop (\cref{sec:generations}) is decided without the mechanical-design exam (its trials of up to three hours could not finish in the lease at the scaled load) and on the pooled exam trials: a disk-quota failure and one overloaded re-run ended part of the original exam jobs, so each candidate is paired with the champion over every trial of the exam and its re-runs that the agent itself ended, leaving out trials ended by an infrastructure exception, a model-server error or a step timeout (the original candidate exams ran overloaded). & generation 0's decision \\
11-07 & Generation 1 explores a frontier pool, the training tasks generation 0 did not always solve and the never-explored training tasks with several faults, with the promoted champion; its exploration is stopped at a fixed time (04:30) to leave the lease room for a second decision, which gave each pool task 3 or 4 attempts instead of the planned 8, and two of the four planned training variants are trained. & generation 1's training \\
11-07 & No third generation was run (it could not finish in the lease). Instead the base sat the held-out exams on generation 1's stack, the repository exam with several faults completed by a second job on the idle cards, so that the base and both champions are compared on one measurement. & the comparison of the base with both champions \\
\bottomrule
\end{tabular}
\caption{Changes made in this paper, continued (the last three rows concern the 27B generations of \cref{sec:generations}).}\label{tab:deviations3}
\end{table}

\begin{table}[tbp]
\centering\small
\setlength{\tabcolsep}{3pt}
\begin{tabular}{@{}p{0.12\linewidth}p{0.58\linewidth}p{0.25\linewidth}@{}}
\toprule
Series & Change & Before \\
\midrule
S1 HL & The first run of arm HL ended on the project disk's inode quota and was not evaluated; the registered replication was rerun on a cleared disk with the same scripts and binaries.\cl{RS16,RS37} & arm HL's evaluation \\
S2 & The policy-gradient objective scores the behaviour policy as the start of the arm, because the records carry no sampler log-probabilities at the launch build; each arm samples fresh from the policy it starts from. Generation 2 was not run (the shared adapter card id); the criterion was already decided by generation 1.\cl{RS08,RS37} & generation 1 \\
S3 & Lineage r2 was relaunched before its first round with the request cap of its held-out proposals raised to 1400 (its seed reached 415 of 450 programs in 700 requests); r3's generation 3 was resumed from its journals after a transient connection close, with the same build and settings.\cl{RS37} & r2's first round; r3's verdict \\
S4 & The examiner's build was changed so that it identifies a declared base by the checkpoint's content under either digest rule, and it served on another card; exam, protocol, serving build, tier, candidates and criterion unchanged.\cl{RS06,RS37} & any verdict \\
\bottomrule
\end{tabular}
\caption{Dated deviations of the self-improvement series, as their launch records and outcome entries
state them.}\label{tab:deviations-rsi}
\end{table}

%% file: sections/13-glossary.tex

\section{Glossary}\label{app:glossary}

\begin{description}[leftmargin=2.2em, style=nextline, itemsep=2pt]
\item[base] The untrained checkpoint a series or comparison starts from (Qwen3-4B in the series S1--S4).
\item[candidate, champion] A candidate is a newly trained model; the champion is the model the next
  round starts from. A candidate replaces the champion when the learner's gate accepts it.
\item[lineage, generation, step] A lineage is one chain of generations from the base, differing from
  its siblings only in its seed; generation $g$ is trained from material generation $g-1$'s champion
  produced; a step compares generation $g$ with generation $g-1$ (generation 1 with the base).
\item[frontier, frontier union] In S4, the programs a champion predicts wrong in at least one of 4
  sampled attempts; the union is every accepted generation's frontier so far.
\item[held-out exam] The 889 code-written Lua items of the series S1--S4: held by the examiner, never read
  by the learner, items overlapping a trained program excluded. Its file is published, so held-out
  means kept from the learner, not secret.
\item[examiner, hand-off, verdict] The examiner is a separate program that holds the exam and grades
  candidates; the hand-off is the one-way directory through which the learner gives it each candidate
  and its trained programs; a verdict is its signed aggregate result for one step, and the plan
  verdict applies the registered rule over a series.
\item[dev set, retention set, probes] The learner's own measurement sets, built from its generator and
  the base's proposals (S series) or written by the base from the manual (probes, \cref{sec:documentation});
  never exam items.
\item[family] The unit a sign test counts once: in the coding experiments the file a fault was seeded
  in; on the S-series exam the program template an item was generated from (25 templates).
\item[exposed] An agent attempt whose tool calls named a hidden path or whose output repeated hidden
  content; it is withheld from training and from the paired test.
\item[D46, P23] Development benchmarks of the coding experiments: D46 is 46 development tasks, one per
  family; P23 is 23 tasks, one per family of the first qualification half.
\item[CE1, CE4, CE8, CE8-T, CE8-T-S, CE8-T-u1b] Coding experiments: the teacher's first pass (CE1), the
  pilot of the headline (CE4), the 27B model on its own trajectories (CE8), its variant with its own
  guard (CE8-T), the shuffled-instruction control (CE8-T-S) and the decision-level variant (CE8-T-u1b).
\item[pass@1 (coding)] In the coding results, tasks solved out of 46, averaged over decoding seeds.
\end{description}